\documentclass[10pt,a4paper]{article}

\usepackage[english]{babel}

\usepackage[letterpaper,top=2cm,bottom=2cm,left=3cm,right=3cm,marginparwidth=1.75cm]{geometry}
\usepackage[utf8]{inputenc}
\usepackage{graphicx}
\usepackage{subfigure}
\usepackage[colorlinks=true, allcolors=blue]{hyperref}
\usepackage{csquotes}

\usepackage{indentfirst}
\usepackage{algorithmicx}
\usepackage{algorithm}
\usepackage[noend]{algpseudocode}

\usepackage{float}
\usepackage{booktabs}
\usepackage{multirow}

\usepackage{amsmath, amsfonts, amsthm, stackrel}

\usepackage{authblk}
\usepackage[T1]{fontenc}

\title{FedDRL: A Trustworthy Federated Learning Model Fusion Method Based on Staged Reinforcement Learning}
\author[a]{Leiming Chen \thanks{Corresponding author: chenleiming2020@163.com}}
\author[a]{weishan Zhang}
\author[a]{Cihao Dong}
\author[a]{Sibo Qiao}
\author[a]{Ziling Huang}
\author[a]{Yuming Nie}
\author[b]{Zhaoxiang Hou}
\author[c]{Chee Wei Tan \thanks{Corresponding author: cheewei.tan@ntu.edu.sg}}
\affil[a]{China University of Petroleum (East China), China}
\affil[b]{Digital Research Institute, ENN Group, China}
\affil[c]{Nanyang Technological University, Singapore}
\date{} 

\begin{document}
\maketitle

\begin{abstract}
Federated learning facilitates collaborative data analysis among multiple participants while preserving user privacy. However, conventional federated learning approaches, typically employing weighted average techniques for model fusion, confront two significant challenges: (1) The inclusion of malicious models in the fusion process can drastically undermine the accuracy of the aggregated global model. (2) Due to the heterogeneity problem of devices and data, the number of client samples does not determine the weight value of the model. To solve those challenge, we propose a trustworthy model fusion method based on reinforcement learning (FedDRL), which includes two stages. In the first stage, we propose a reliable client selection mechanism to exclude malicious models from the fusion process. In the second stage, we propose an adaptive model fusion method that dynamically assigns weights based on model quality to aggregate the best global models. Finally, We validate our approach against five distinct model fusion scenarios, demonstrating that our algorithm significantly enhances reliability without compromising accuracy.
\end{abstract}

\section{Introduction}

With the advent of deep learning technologies, various industries have been integrating these technologies into their sectors, promoting the development of intelligent transportation, smart logistics, and healthcare systems. These technologies are crucial in reducing production and management costs, enhancing operational efficiency, and accelerating industry digitization. However, supervised learning remains the primary method for training deep learning models, where the volume and diversity of samples are essential for creating high-quality models. Consequently, acquiring extensive and varied data samples has emerged as the initial step in training deep learning models. This approach has led to sample sources expanding from single industries to collaborations across multiple sectors to develop large-scale datasets. To achieve multi-party joint data analysis under the condition of protecting data security and privacy, Google has proposed federated learning technology for the first time. Although federated learning solves the problem of user privacy protection, the traditional federated learning algorithm assumes that all participants are trustworthy. On the contrary, in the actual scenario,  if participants exhibit malicious behavior and intentionally contribute harmful models to the fusion process, it can significantly disrupt the global model's convergence. Thus, creating adaptive defenses for federated learning systems becomes increasingly crucial \cite{AttackRL}. Identifying methods to remove malicious models in federated learning model fusion has become a critical issue. Simultaneously, when a client submits low-quality models for fusion, determining how to adaptively adjust each model's fusion weights based on their quality is also an urgent problem needing resolution. Some studies have applied reinforcement learning techniques to address these weighting issues. For instance, the Favor \cite{Favor} method uses the DDPG to assign weights to participant models. Additional research has applied reinforcement learning to address device selection \cite{zhangPeiYingRL} \cite{digitalTwinRL}, resource optimization \cite{RoF} \cite{DDQNTrust}, and communication optimization in IoT federated learning contexts.

Reinforcement learning (RL) employs a trial-and-error strategy. The essence of this approach is training an intelligent agent that interacts with the external environment through varied actions. The environment then provides feedback in the form of rewards and penalties based on the agent's actions, guiding the agent toward optimal action selection by maximizing reward value. However, employing reinforcement learning presents certain challenges. Firstly, continuous training is required for sample collection through environmental interaction. When the cost of such interactions is prohibitive or unacceptable (for example, in our scenario, where the server must frequently calculate the global model's parameters), the efficiency of sample collection significantly impacts the reinforcement learning training duration. Secondly, when the agent's action space is vast and continuous, it leads to prolonged sampling periods. These issues mean traditional single-agent reinforcement learning training approaches can be exceedingly time-consuming. Applying reinforcement learning in federated learning requires addressing these problems, as increasing participant numbers escalates agent training time. Therefore, optimizing the action space for reinforcement learning to expedite the agent training process is an essential challenge to address.

Why opt for phased reinforcement learning? We take an example to explain this problem. Consider a robot learning to cook through reinforcement learning, with the process divided into washing, chopping, and cooking stages. The robot must master each stage to prepare a successful dish. Traditional reinforcement learning aims to identify the optimal action across all stages simultaneously; however, mastering the initial stage is essential before progressing. By adopting a phased learning approach, the robot sequentially masters each stage, streamlining the learning process and leading to more effective outcomes. Similarly, if malicious models are not initially filtered out, the agent's trial-and-error costs in weight assignment for these models will increase. To resolve these issues, we propose a staged reinforcement learning algorithm (FedDRL).
The contributions of the paper are as follows.
\begin{itemize}
\item We design a federated learning framework that employs reinforcement learning for model fusion, designed to select trustworthy clients and optimally assign model weights.

\item We propose an adaptive client selection strategy based on the A2C algorithm, dynamically identifying and selecting trustworthy clients while excluding malicious ones from the model fusion process based on situational analysis.

\item We propose an adaptive weight assignment method that adaptively adjusts the weights according to the quality of their uploaded models.

\item We propose an adaptive weight assignment method that adaptively adjusts the model fusion weights according to the quality of their uploaded models.

\item We present five types of model fusion scenarios to validate the performance of each algorithm. We also compare the performance of our algorithm with the baseline algorithm on three public datasets.
\end{itemize}

\section{Related Work}
\subsection{Federated Learning}
Research in federated learning primarily aims to address two challenges: enhancing the generalization of the global model on the server side and personalizing the model on the client side. Consequently, federated learning algorithms are bifurcated into server-side and client-side optimization strategies. Google initially introduced the FedAvg algorithm \cite{FedAvg} to address the problem of server-side global model fusion. To improve global model convergence, Karimireddy et al. developed the Scaffold method \cite{Scaffold}, which mitigates client-side drift by integrating a control variable. Similarly, Li et al. introduced FedProx \cite{FedProx}, applying a regularization function to client models to correct deviations. Additionally, Wang et al. unveiled FedNova \cite{FedNova}, addressing global model convergence issues by normalizing parameters on both client and server ends. Furthermore, Li et al. have introduced the MOON \cite{MOON} technique, leveraging model comparison learning to enhance global model convergence. Chen et al. \cite{FedCFB}. also proposed a client identification method based on model parameter features to achieve trustworthy federated learning.

While those approaches enhance the global model's convergence speed, practical federated learning situations reveal variances in the quality of models trained by individual participants. These discrepancies stem from the diversity in computational resources and the calibre of data samples available to each participant. Additionally, variations arise due to the quantity and type of samples possessed by each participant, a phenomenon known as Non-IID (Non-Independent and Identically Distributed). Consequently, these factors complicate the attainment of optimal global model aggregation in the Non-IID environments.

\subsection{Challenges of Non-IID Data Distribution}
The Non-IID data issue significantly impacts federated learning models' convergence. Zhao et al. explored various federated learning methods' performance on non-IID datasets, demonstrating significant accuracy challenges \cite{nonIIDProblem}. Accordingly, several studies have addressed the non-IID dilemma in federated learning. For instance, Zhang et al. proposed the FedPD approach \cite{FedPD}, optimizing models and communication for non-convex objective functions. Moreover, Gong et al. introduced AutoCFL \cite{AutoCFL}, utilizing a weighted voting client clustering strategy to mitigate non-IID and imbalanced data effects. Huang et al. developed FedAMP \cite{FedAMP}, which addresses Non-IID data-induced client-side model personalization issues through personalized model updates. Li et al. devised Fedbn \cite{Fedbn}, incorporating a batch normalization layer into local models to address feature shift challenges due to data heterogeneity. Briggs et al. suggested a hierarchical clustering method (FL+HC) \cite{FLHC}, improving Non-IID dataset model performance by grouping clients for independent model training. Additionally, Gao et al. offered the Feddc approach \cite{Feddc}, bridging client and global model parameter disparities through a control variable. Lastly, Mu et al. introduced Fedproc \cite{Fedproc}, directing client model training by integrating a comparative loss between client and global models. Chen \cite{FedTKD} et al. proposed a federated learning method based on adaptive knowledge distillation to improve the accuracy of heterogeneous model scenarios.

Although these methodologies advance Non-IID issue mitigation in federated learning, they typically assign uniform fusion weights to all clients, failing to exclude malicious or low-quality model contributions. Consequently, dynamically selecting clients for fusion and adaptively calculating each model's weight remains critical for successful global model integration.

\subsection{Federated Reinforcement Learning}
Given the adaptive learning potential of reinforcement learning, its application within federated learning contexts has garnered interest. Some research has concentrated on leveraging reinforcement learning to boost global model performance. For instance, Wang et al. introduced the Favor method \cite{Favor}, which adaptively selects clients for model fusion. Sun et al. developed the PG-FFL framework \cite{PGFFL}, addressing the challenge of client weight computation during model fusion. Additional studies have applied reinforcement learning for device optimization within federated IoT frameworks. For example, Zhang et al. utilized the DDPG algorithm \cite{zhangPeiYingRL} for optimal device selection. Zhang also formulated the FedMarl strategy \cite{FedMarl}, employing multi-agent reinforcement learning for node selection. Similarly, Yang et al. proposed a digital twin architecture (DTEI) \cite{digitalTwinRL}, applying reinforcement learning for device selection issues. Other investigations have addressed resource optimization and scheduling challenges within IoT contexts, such as Zhang et al.'s RoF methodology \cite{RoF}, which leverages multi-intelligent reinforcement learning for optimal resource scheduling. Additionally, Rjoub et al. have developed trusted device selection techniques \cite{trustClient} and the DDQN-Trust method \cite{DDQNTrust}, utilizing Q-learning to assess devices' credit scores for optimal scheduling. To ameliorate federated learning communication issues, Yang et al. introduced a reinforcement learning-based model evaluation method \cite{evaluationRL}, selecting optimal devices for training and fusion. Nevertheless, while these efforts predominantly focus on IoT environment applications—such as device selection, resource optimization, and communication enhancement—they seldom address federated learning's model weight calculation challenges. Therefore, Zhang et al. proposed the $R^2$Fed framework \cite{R2Fed}, employing the DDPG reinforcement learning method for adaptive client weight calculation.
Chen et al. \cite{FedPlatform} designed a task platform for implementing trustworthy federation learning.

Although current research addresses the issue of weight allocation in federated learning, it often neglects the training efficiency of the agents. Therefore, optimizing the training efficiency of agents is a significant challenge that needs attention.

\section{Method}
\subsection{Problem Definition}
In this section, we scrutinize the prevailing challenges of the current federated learning approach and subsequently propose a solution. In federated learning, the objective is to get the global model by amalgamating local models from all clients through server-side aggregation.
We define n clients as involved in model fusion, and the client is denoted as $C_{i}$ 
where $C_{i} \in\left\{C_{1}, C_{2}, C_{3} \ldots C_{n}\right\}$. 
Each client has a network model $M_{i}$, 
where $M_{i} \in\left\{M_{1}, M_{2}, M_{3} \ldots M_{n}\right\}$.
Each client has its private data $D_{i}$,
where $D_{i} \in\left\{D_{1}, D_{2}, D_{3} \ldots D_{n}\right\}$.
The number of samples in each dataset is $S_{i}$, where $S_{i} \in\left\{S_{1}, S_{2}, S_{3} \ldots S_{n}\right\}$. The total number of samples is $\sum_{i=1}^{N} S_{i}$.
We define the $\theta_{i}$ as a model parameter of $M_{i}$.
where $\theta_{i} \in\left\{\theta_{1}, \theta_{2}, \theta_{3} \ldots \theta_{n}\right\}$.

Additionally, the server-side model aggregation process per round is defined as shown in equation \ref{eqn-1}:
\begin{equation}
\label{eqn-1} 
\theta_{\text {global }}=\sum_{i=1}^{N} w_{i} \theta_{i},
\text { where } w_{i}=\frac{S_{i}}{\sum_{i=1}^{N} S_{i}}, \quad  w_{i} \geq 0, \sum_{i=1}^{N} w_{i}=1
\end{equation}
The $ w_{i}$ is the fusion weight of each model parameter.

Traditional Federated Learning typically employs a weighted average approach for computing model fusion weights, with each model's weight determined by its corresponding client's data sample size relative to the total. Thus, clients contributing more data exert a greater influence on the aggregated model. However, this method fails to consider the quality of each client's model and the potential inclusion of malicious models in real-world scenarios. We illustrate the deficiencies of the traditional federated fusion algorithm through two scenarios:

\textbf{Scenario 1}: A client's data represents 20\% of the total, yet its model's accuracy is merely 53\%. Employing the conventional federated fusion algorithm in this case would detrimentally impact the global model's accuracy.

\textbf{Scenario 2}: A client engaged in model fusion launches malicious attacks, intentionally skewing its model's output to reflect a mere 10\% accuracy. If such malicious models are incorporated through the standard fusion process, the accuracy of the global model would be severely compromised.

Addressing these challenges necessitates an adaptive weight calculation strategy capable of nullifying malicious models by assigning them a weight of zero, thus excluding them from the fusion process. Concurrently, this approach should dynamically adjust the weights of each client's model, prioritizing those of higher quality to enhance the global model's overall accuracy.

Adopting a single-agent reinforcement learning strategy to tackle these issues introduces new challenges. As the number of clients increases, so too does the agent's action space, prolonging the training duration. Additionally, a single-agent framework is limited to interacting with just one environment, further extending the sampling period. We propose a bifurcated solution inspired by hierarchical reinforcement learning to mitigate these concerns, thereby streamlining the lengthy reinforcement learning training process. This solution comprises two primary stages: the selection of trustworthy clients and the assignment of optimal weights.

\begin{figure}
	\centering
	\includegraphics[width=0.95\linewidth, height=0.75\textheight]{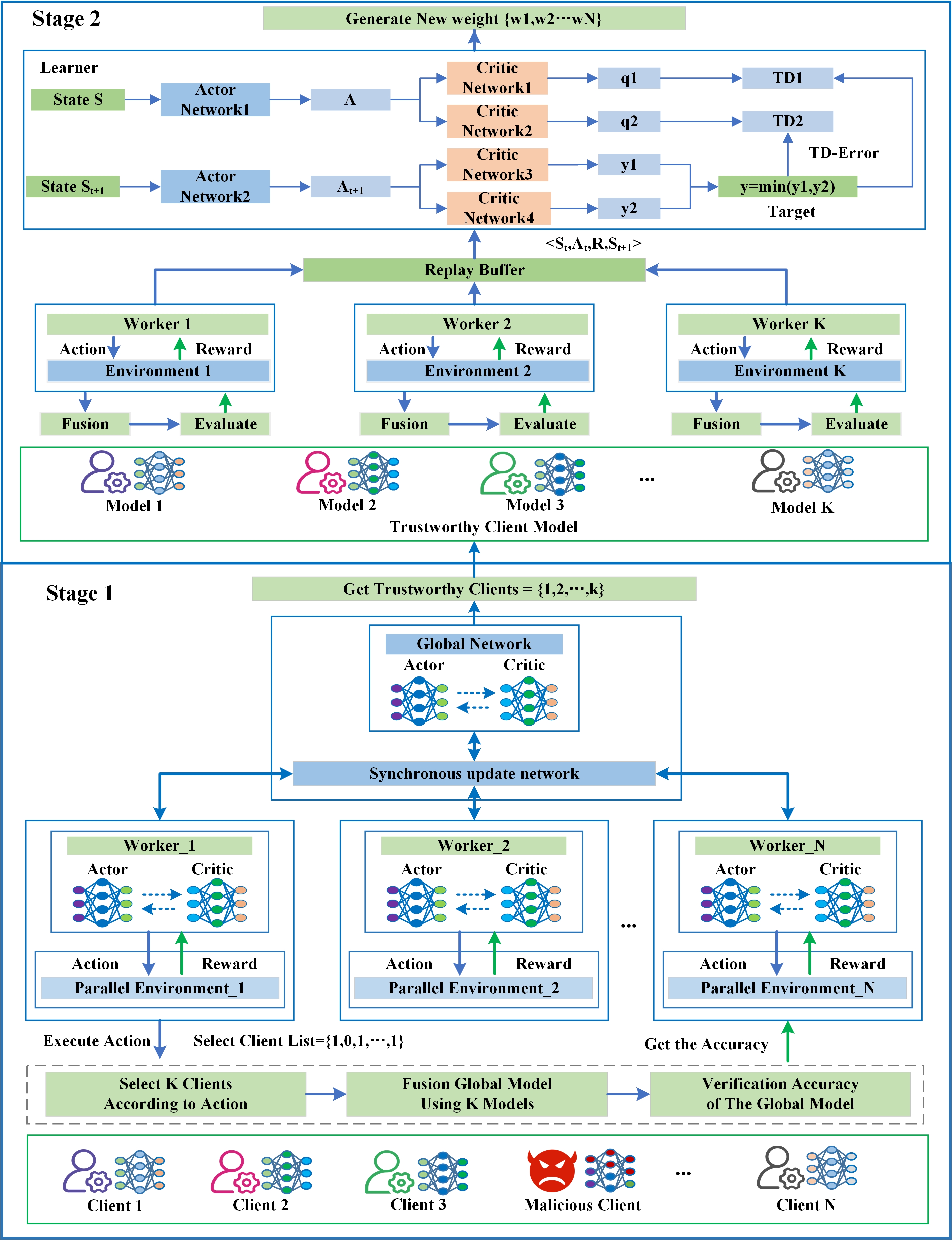}
	\caption{The Process of FedDRL framework}
	\label{fig-FedDRL}
\end{figure}

\textbf{Stage 1}: During this phase, the objective is to identify K trustworthy models from a pool of N for inclusion in the global model fusion. Identifying clients who have uploaded malicious models is challenging. We address this by employing reinforcement learning to dynamically select and autonomously screen client models, as delineated in equation \ref{eqn-2}.
\begin{equation}\label{eqn-2}
\{M_{a}, M_{\mathrm{b}}, \ldots, M_{\mathrm{k}}\} \leftarrow \text{SelectTrustworthyModel}\left(\{M_{1}, M_{2}, \ldots, M_{\mathrm{n}}\}\right)
\end{equation}

\textbf{Stage 2}: Building on the first step, we then allocate optimal weights to the verified models to bolster the global model's accuracy, formalized in equation \ref{eqn-3}.
\begin{equation}\label{eqn-3}
\{W_{1}, W_{2}, \ldots, W_{\mathrm{n}}\} \leftarrow \operatorname{AdaptCalculateWeight}\left(\{M_{a}, M_{b}, \ldots, M_{\mathrm{k}}\}\right)
\end{equation}
Here, $AdaptCalculateWeight(.)$ signifies a method for adaptive weight computation, and $W_{i}$ represents the optimal computational weight assigned to each client's output.

\subsection{A Trustworthy Federated Learning Approach Based on Staged Reinforcement Learning}
To address these challenges, we introduce a trusted federated learning framework anchored in staged reinforcement learning (FedDRL). This framework unfolds across two distinct phases. In the first phase, we propose an adaptive client selection strategy aimed at identifying and selecting trustworthy clients for participation in model fusion. Subsequently, in the second phase, we formulate a model weight assignment algorithm designed to dynamically allocate fusion weight values to models based on the prevailing fusion environment. The process is depicted in Figure \ref{fig-FedDRL}. 

\subsubsection{Adaptive client selection method}  \label{A2Csubsection}
Once we have defined the base elements of reinforcement learning, We use a distributed A2C approach to train the agent; A2C is an improved method-based A3C algorithm \cite{A3C}. Figure \ref{fig-FedDRL} shows the A2C architecture, which consists of a central node and $K$ workers. Each worker contains an Actor and a Critic network, where the actor network generates action, and the Critic network evaluates the action and gives the corresponding reward. Meanwhile, each worker independently interacts with the related environment to achieve sampling and training of the Actor and Critic networks. In addition, the Actor and Critic networks of the central node are used to synchronize the network information of each worker and to achieve the fusion and sharing of network parameters of multiple workers.

Therefore, our main objective is to train Actor and Critic networks. We define the Actor-network parameters as $\pi(\theta)$ and the Critic network parameters as $V(w)$. The process of the worker and the central node is as follows.

\textbf{Step 1}: Each worker initializes the local network by pulling the global network model parameters from the centre node. Then, each worker trains the Actor and Critic networks by interacting with the environment independently. Finally, the two networks are uploaded to the central node.

\textbf{Step 2}: After the central node collects the network parameters uploaded by all workers, it updates the global model by the weighted averaging method. Then, the server sends the two networks to each worker.

Steps 1 and 2 are repeated according to the total number of times to obtain the final global model.

The training process for the step 1 neutralization network is as follows: The gradient of the primary communication algorithm of the policy network is calculated as equation \ref{eqn-8}.

\begin{equation}\label{eqn-8} 
\nabla_{\theta} J(\theta)=\nabla_{\theta} \log \pi\left(a_{t} \mid s_{t} ; \theta\right) A\left(s_{t}, a_{t} ; \mathrm{w}\right)
\end{equation}

Where $A\left(s_{t}, a_{t} ; \theta_{v}\right)$ is the advantage function. The k-step sampling strategy is used in the A2C algorithm to calculate the advantage function, so the definition is expressed as equation \ref{eqn-9}.

\begin{equation}\label{eqn-9} 
A\left(s_{t}, a_{t} ; \theta_{v}\right)=\sum_{i=0}^{k-1} \gamma^{i} r_{t+i}+\gamma^{k} V\left(s_{t+k} ; \mathrm{w}\right)-V\left(s_{t} ; \mathrm{w}\right)
\end{equation}
The Loss function of the actor network is calculated as in equation \ref{eqn-10}, and The Critic network is calculated as in equation \ref{eqn-11}.
\begin{equation}\label{eqn-10} 
\nabla_{\theta}J(\theta)=\nabla_{\theta}\log \pi\left(a_{t} \mid s_{t} ; \theta\right)\left(\sum_{i=0}^{k-1} \gamma^{i} r_{t+i}+\gamma^{k} V\left(s_{t+k} ; \mathrm{w}\right)-V\left(s_{t} ; \mathrm{w}\right)\right)
\end{equation}

\begin{equation}\label{eqn-11} 
\nabla_{w}J(w)=\nabla_{w}\left(\sum_{i=0}^{k-1} \gamma^{i} r_{t+i}+\gamma^{k} V\left(s_{t+k} ; \mathrm{w}\right)-V\left(s_{t} ; w\right)\right)^{2}
\end{equation}
We update the Actor and Critic network parameters using the derivative formula as equation \ref{eqn-12}.
\begin{equation}\label{eqn-12} 
w \leftarrow w+\nabla_{w} J(w), \quad \theta \leftarrow \theta+\nabla_{\theta} J(\theta)
\end{equation}
Finally, each worker uploads the Actor and Critic network to the server. Then, the network parameters of the server are calculated using the weighted average method. The process is equation \ref{eqn-13}.
\begin{equation}\label{eqn-13} 
w_{global}=\frac{1}{n}   {\textstyle \sum_{1}^{n}} w_{i},\theta_{global} =\frac{1}{n}   {\textstyle \sum_{1}^{n}} \theta_{i}, i\in [1,n]
\end{equation}

When the parameters of the Actor and Critic networks in the central stage are updated, the central node sends down these two networks to all workers, and each worker uses the updated networks to continue interacting with the external environment. The process is repeated for the specified number of rounds until the agent at the central node can obtain a stable reward value.
 The process is shown in algorithm \ref{Algorithm 1}.
 
\begin{algorithm}
    \caption{The process of trustworthy client selection}
    \label{Algorithm 1}
    \begin{algorithmic}[1]
        \Require { 
        Client Models $ \left\{m_{1}^{t}, m_{2}^{t}, m_{3}^{t}, \ldots m_{\mathrm{n}}^{t}\right\}$, Round T, Worker Number K, 
        Sampling Step Length S
        }
        \Ensure{ Chosen Credible Client Model List M =  $ \left\{m_{2}^{t}, m_{3}^{t}, \ldots m_{\mathrm{k}}^{t}\right\}$}

        \State /* Each Worker Training Step */
        \State worker $ (\theta, w) \leftarrow  GetGlobalParamter  \left(\theta_{\text {global }}, w_{\text {global }}\right)$
        \State  the Client Upload Current Epoch Model, Turn to State $\mathrm{s}_{0}$, $t_{\text {start }}= t = 1$
        \For{$e$ from 1 to $S$} 
            \State According to Current State $\mathrm{s}_{0}$ Randomly Choose Action $\mathrm{s}_{t}$
            \State ${s_{t}, a_{t}, r, s_{t+1}} \leftarrow \operatorname{Step}\left(\mathrm{a}_{t}\right)$ 
            // Execute Action $\mathrm{a}_{t}$ to Acquire Reward $\mathrm{r}$ and Next State $\mathrm{s}_{t+1}$
            \State $\mathrm{t_{start}}$ = $\mathrm{t_{start}}$ + 1
            \State if $\mathrm{s_{t}}$ != terminal: R $\leftarrow$ V$(\mathrm{s_{t}}; w )$ else: R = 0
        \EndFor
        \For{ $\mathrm{i} \in\left\{t-1, \ldots, \mathrm{t}_{\text {start }}\right\} $}
                \State R $\leftarrow {\mathrm{r_{i}} + \gamma}$R // Compute Target TD
                \State $\nabla_{\theta} J(\theta)=\nabla_{\theta} \log \pi_{\theta}\left(a_{t} \mid s_{t}\right)\left(R-v\left(s_{i} ; w\right)\right)$ // Compute Strategy Gradient
            \State $\nabla_{w} J(w)=\nabla_{w}\left(R-v\left(s_{i} ; w\right)\right)^{2}$ // Compute Critic Network Gradient
            \State Update Actor Network Parameters: $\theta\leftarrow{\theta+\nabla_{\theta}J(\theta)}$
            \State Update Critic Network Parameters:          $w \leftarrow w+\nabla_{w} J(w)$
            
        \EndFor
        \State /* Center Node Process */
        \For{$round$ from 1 to $T$} 
            \For{$\mathrm{worker}_{i}$ from 1 to $K$} 
                \State Receive Each Worker Parameters$(\theta, w)$
                \State $\operatorname{Global}\left(\theta_{\text {global }}, w_{\text {global }}\right) \leftarrow \mathbf{A g g}\left(\left\{\left(\theta_{1}, w_{1}\right),\left(\theta_{2}, w_{2}\right), \ldots\right\}\right)$ // Aggregate Parameters
                \State $\mathrm{worker}_{i}\leftarrow{SendGlobal\left(\theta_{\text {global }}, w_{\text{global }}\right)}$ // Send New Parameters to Worker
            \EndFor
        \EndFor
    \State /* Results Process */
    \State Output Trusted Client Model List M = $\left\{m_1^t, m_2^t \ldots m_k^t\right\}$
    \end{algorithmic}
\end{algorithm}

\subsubsection{Adaptive model weight calculation method} \label{TD3subsection}

In this phase, our main objective is to achieve the optimal weight assignment for each model. For each communication round, we assume that $K$ trustworthy client models were selected. We need to train the agents in each communication round and use the weight output of the agent to achieve the global model fusion. We first describe the process of global model fusion for agent-based actions. We define $\theta_i$ as the $i$-th client model, and the all client models as $\{ \theta_1, \theta_2, \cdots, \theta_k\}$. We also define $s_i$ as the number of samples of $i$-th client. The process is as follows:

(1) In this step, the agent needs to output the weight values for each model. We define the $t$-th time, the action adopted by the agent as equation \ref{eqn-14}. 
\begin{equation}\label{eqn-14} 
\mathrm{W^{t}}=\left\{w_{1}^{t}, w_{2}^{t}, w_{3}^{t} \ldots w_{k}^{t}\right\}
\end{equation}
$w_{i}$ is the $i$-th weight value output by agent for $i$-th model.

(2) We aggregate the global models based on the model weights assigned by the agent, and the process is expressed as equation \ref{eqn-15}.
\begin{equation}\label{eqn-15} 
\theta_{global}^{k}={\textstyle \sum_{t=1}^{T}}  w_{i}^{t} \theta_{i}
\end{equation}

 We aim to train the agent so that it can output the optimal fusion weight values based on the quality of each model. To accomplish this goal, we first describe the basic elements of reinforcement learning as follows:

\textbf{Environment}: The external environment is the server-side global model fusion module, which fuses the global model based on the actions output by the agent and then verifies the accuracy of the global model on the reserved dataset on the server side. Finally, the server side feeds back to the agent the corresponding reward and punishment values based on the accuracy of the global model.

\textbf{State}: We define the agent's state information to include the number of samples corresponding to each client, the accuracy of each client's model, and the accuracy of the global model fused using the weights output by the agent. as shown in \ref{eqn-16}.
\begin{equation}\label{eqn-16} 
\mathrm{S^{t}}=\left\{s_{1}^{t}, s_{2}^{t}, s_{3}^{t} \ldots s_{k}^{t}, acc_{1}^{t}, acc_{2}^{t}, acc_{3}^{t} \ldots acc_{k}^{t}, acc_{global}^{t}\right\}
\end{equation}
The $acc_{global}^{t}$ is the accuracy of the global model fused using the weights output by the agent.

\textbf{Action}: In each stage, the agent needs to assign each model's weights based on the model's quality.
The action space is shown as \ref{eqn-17}. $a_{i}^{t}$ denotes the weight value assigned to the $i$-th client in the state t, while the sum of the corresponding weight values of all clients is 1.
\begin{equation}\label{eqn-17} 
\mathrm{A^{t}}=\left\{a_{1}^{t},a_{2}^{t},...a_{k}^{t}\right\},  
 {\textstyle \sum_{1}^{k}} a_{i}^{t}=1,a_{i}^{t}\in (0,1)
\end{equation}

\textbf{Reward}: We define the model accuracy aggregated using the average method as $Acc_{all}$, where each model weight is $\frac{1}{N} $. At the $m$-th time, we define the weight set output by the agent as $W$. Then, we use the weight set to fusion the global model, and we define the accuracy of the global model as $Acc_{m}$. We calculate the reward value by subtracting the difference of $Acc_{m}$ from $Acc_{all}$. If the calculated result is greater than zero, this indicates that the weights assigned by the agent improve the accuracy of the global model, and we give a positive reward. Conversely, we give a penalty reward. $\varphi$, $ \phi$ denotes the reward and penalty factors, respectively. So, the reward is defined as equation \ref{eqn-18}.

\begin{equation}\label{eqn-18} 
Reward = \begin{cases} 
\varphi \cdot (Acc_{m}-Acc_{all} ) ,  Acc_{m} >Acc_{all} 
\\
\phi \cdot (Acc_{m}-Acc_{all} ), Acc_{m} \le Acc_{all} 
\end{cases}
\end{equation}

When we have finished defining the basic elements, We implement a distributed reinforcement learning approach based on TD3 \cite{TD3} to train the agent. The training process is shown in figure \ref{fig-FedDRL}. This stage includes a central Learner and multiple Worker nodes. Each worker corresponds to a parallel environment. The workflow of each worker is as follows: first, each worker performs global model fusion based on the assigned weights; then verifies the accuracy of the global model by interacting with the parallel environment; and finally receives the reward values from the parallel environment feedback. Finally, each worker stores the corresponding ones in the sampling buffer pool. Multiple workers interact with each environment independently, thus achieving parallel sampling to improve the sampling efficiency. After each worker collects a certain batch of samples, the Learner trains the agent by taking a certain amount of sample data from the experience pool.

The TD3 algorithm consists of six network models, including an Actor network $P(w)$, two Critic networks $Q_{1}\left(\theta_{1}\right)$,$Q_{2}\left(\theta_{2}\right)$, and a target Actor-network $P^{\prime}(w)$, two target Critic networks $Q^{\prime}_{1}\left(\theta_{1}\right), Q^{\prime}_{2}\left(\theta_{2}\right)$. 
Each network is shown in figure \ref{fig-FedDRL}.
The Learner randomly draws $N$ batches of sample data from the buffer pool every certain round to train the model. The training processes are as follows. 

(1) First, select the action  $a_{t+1}$ based on the target Actor-network  $\mathrm{P}^{\prime}\left(\mathrm{~s}_{t+1} \right)$. The state $\mathrm{s}_{t+1}$  and action $a_{t+1}$ are input to the target Critic network$Q_{1}^{\prime}\left(\theta_{1}^{\prime}\right)$ and $Q_{2}^{\prime}\left(\theta_{2}^{\prime}\right)$, respectively. The two target Critic networks will calculate the predicted reward $q_{1}$ and $q_{2}$.

(2) The TD target value is calculated using equation \ref{eqn-19}, where $\operatorname{Min}\left(q_{1}, q_{2}\right)$ takes the minimum value of both.
\begin{equation}\label{eqn-19} 
y_{t} \leftarrow r+\gamma \operatorname{Min}(q 1, q 2)
\end{equation}

(3) Select the action based on the actor network, input the state and action into the critical network separately, and let these two networks output the corresponding prediction reward sum.

(4) Calculate the TD error. The calculation formula is as equation \ref{eqn-20}.
\begin{equation}\label{eqn-20} 
\delta_{1, t}=q_{1, t}-y_{t}, \quad \delta_{2, t}=q_{2, t}-y_{t}
\end{equation}

(5) Update the Critic network as equation \ref{eqn-21}.

\begin{equation}\label{eqn-21} 
\begin{array}{l}\theta_{1} \leftarrow \theta_{1}-\alpha \cdot \delta_{1, t} \cdot \nabla_{w} Q_{1}\left(s_{t}, a_{t} ; \theta_{1}\right) \\\theta_{2} \leftarrow \theta_{2}-\alpha \cdot \delta_{2, t} \cdot \nabla_{w} Q_{2}\left(s_{t}, a_{t} ; \theta_{2}\right)\end{array}
\end{equation}

(6) Update the strategy network every d rounds through the Actor-network output action as equation \ref{eqn-22}.
\begin{equation}\label{eqn-22} 
\mathrm{w} \leftarrow \mathrm{w}+\beta \cdot \nabla_{w} P\left(s_{t} ; \mathrm{w}\right) \cdot \nabla_{w} Q_{1}\left(s_{t}, a_{t} ; \theta_{1}\right)
\end{equation}

(7) Update the target Actor and Critic network parameters every d rounds as equation \ref{eqn-23}.
\begin{equation}\label{eqn-23} 
\begin{array}{l}w^{\prime} \leftarrow \tau w+(1-\tau) w^{\prime} \\\theta_{1}^{\prime} \leftarrow \tau \theta_{1}+(1-\tau) \theta_{1}^{\prime} 
\\\theta_{2}^{\prime} \leftarrow \tau \theta_{2}+(1-\tau) \theta_{2}^{\prime}
\end{array}
\end{equation}

Repeating the above steps for the specified number of rounds, we will get the trained agent. Finally, we output the optimal value of each model through the agent. The process is shown in algorithm \ref{Algorithm 2}
\begin{algorithm}
    \caption{The process of model weight calculation}
    \label{Algorithm 2}
    \begin{algorithmic}[1]
        \Require { 
        Client Models $ \left\{\theta_{1}^{t}, \theta_{2}^{t}, \theta_{3}^{t}, \ldots \theta_{\mathrm{n}}^{t}\right\}$, Round R, Worker Number N, 
        Buffer Memory Pool M
        
        Initialize Learner Parameters: Actor Parameter P$\left(w\right)$, Critic Network $Q_{1}\left(\theta_{1}\right), Q_{2}\left(\theta_{2}\right)$
        
        Target Actor Parameter $P^{\prime}\left(w^{\prime}\right)$, Target Critic Network $Q_{1}^{\prime}\left(\theta_{1}^{\prime}\right), Q_{2}^{\prime}\left(\theta_{2}^{\prime}\right)$

        $w^{\prime}\leftarrow w, \theta_{1}^{\prime}\leftarrow\theta_{1}, \theta_{2}^{\prime}\leftarrow\theta_{2}$
        }
        \Ensure{ Optimized Client Model Weight $W = \left\{w_{1}^{t}, w_{2}^{t}, \ldots w_{\mathrm{k}}^{t}\right\}$}

        \State /* Each Worker Sampling Step */
        \For{$worker$ from 1 to $N$} 
            \State $a_{t}\leftarrow P\left(s_{t}, w\right)$  // Randomly Choose an Act from $P\left(s_{t}, w\right)$
            \State $\left\{w_{1}^t, w_{2}^t, w_{3}^t \ldots w_{k}^t\right\}\leftarrow Step \left(a_{t}\right)$
            \State $\theta_{global}^t\leftarrow Agg \left(\sum_{i=1}^k w_{i}^t\theta_{i}\right)$
            \State $R_{t}\leftarrow CaculateReward \left(ACC_{t}-ACC_{avg}\right)$
            \State $M\leftarrow Store \left(<S_{t}, A_{t}, R_{t}, S_{t+1}>\right)$
        \EndFor
        \State /* Center Learner Training Step */
        \For{$r$ from 1 to R}
            \State Randomly Sampling N Batches of Data from M
            \State $a_{t+1}^{\prime}\leftarrow P^{\prime}\left(s_{t+1}\right)$
            \State $y\leftarrow r+\gamma Min\left(Q_{1}^{\prime}\left(s_{t+1}, a_{t}^{\prime}\right), Q_{2}^{\prime}\left(s_{t+1}, a_{t}^{\prime}\right)\right)$
            \State Update Critic Network $\theta_{1}\leftarrow argmin_{\theta_{1}}\frac{1}{N}\sum \left(y-Q_{\theta_{1}}\left(s, a\right)\right)^2$
            \State Update Critic Network $\theta_{2}\leftarrow argmin_{\theta_{2}}\frac{1}{N}\sum \left(y-Q_{\theta_{2}}\left(s, a\right)\right)^2$
            \State Every d Rounds:
                \State Update Actor-Network: $\nabla_{w}J\left(w\right)=N^{-1}\sum\nabla_{w}Q_{\theta_{1}}\left(s, a\right)\mid_{a=P\left({s}\right)}\nabla_{w}P\left({s}\right)$
                \State Update Target Critic Network: $\theta_1^{\prime}\leftarrow\tau\theta_1+\left(1-\tau\right)\theta_1^{\prime}, \theta_2^{\prime}\leftarrow\tau\theta_2+\left(1-\tau\right)\theta_2^{\prime}$
                \State Update Target Actor-Network: $w^{\prime}\leftarrow\tau w+\left(1-\tau\right)w^{\prime}$
        \EndFor
    \State After R Rounds, Save Trained Model
    \State Output Optimized Model Weight W = $\left\{w_1^t, w_2^t \ldots w_k^t\right\}$
    \end{algorithmic}
\end{algorithm}

\section{SYSTEM DESIGN}
\begin{figure}
	\centering
	\includegraphics[width=0.88\linewidth, height=0.45\textheight]{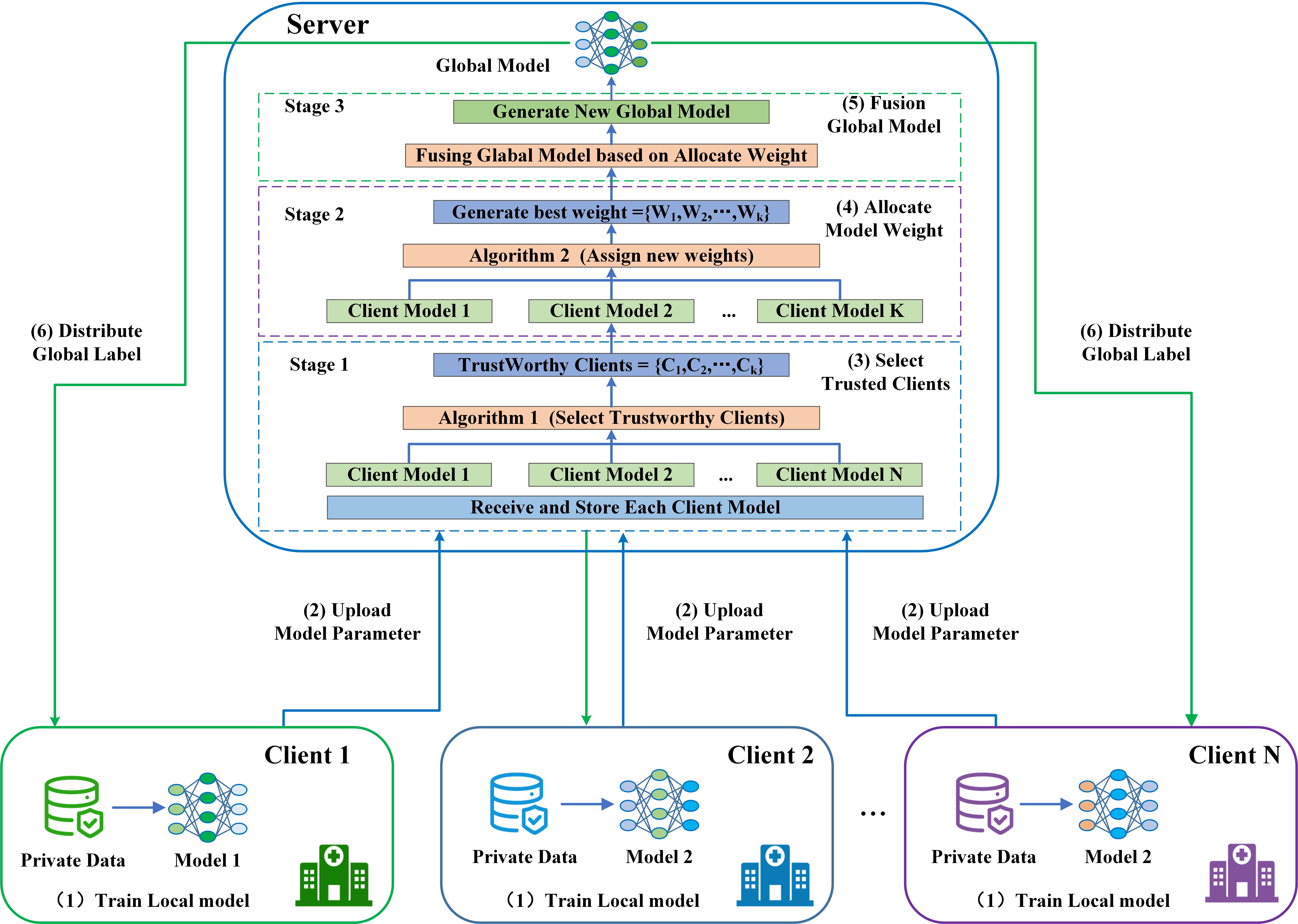}
	\caption{The system architecture of FedDRL}
	\label{fig-System}
\end{figure}

To establish a reliable federated learning process, we developed a framework for trustworthy federated learning (FedDRL). The framework employs a staged reinforcement learning approach to achieve trustworthy federated learning. In the first stage, we train agents to accomplish the selection of trustworthy clients to participate in global model fusion. Then, in the second stage, we also use the trained agent to dynamically adjust the fusion weights of each model and finally realize the optimal global model fusion. The framework workflow consists of six steps, as shown in Figure \ref{fig-System}.

\textbf{Step 1 (Local Model Training)}:
Each client downloads the global model, initializes its parameters accordingly, and conducts model training using local private data.

\textbf{Step 2 (Upload Model)}:
After local model training, each client uploads its model parameters to the server.

\textbf{Step 3 (Select Trustworthy Clients)}:
Upon receiving client model parameters, the server employs the $SelectTrustClient(.)$ algorithm to train an agent. Subsequently, the trained agent selects trustworthy clients.

\textbf{Step 4 (Assigning Model Weights)}:
 The server Utilizes models from trustworthy clients and performs global model fusion. It then employs the $AdaptCalculateWeight(.)$ algorithm to train an agent, which optimizes weight assignments for each client model.

\textbf{Step 5 (Fusing Global Model)}:
The server fuses the global model using the calculated weights from the previous step.

\textbf{Step 6 (Distribute Global Model)}:
The server disseminates the global model to all clients, initiating the subsequent federation task.

The federation task is set to execute a specified number of communication rounds until the final global model is obtained. This process is shown in Algorithm \ref{Algorithm 3}.

\begin{algorithm}
    \caption{ The FedDRL framework}
    \label{Algorithm 3}
    \begin{algorithmic}[1]
        \Require { 
        Private Dataset $ \left\{D_{1}, D_{2}, \ldots D_{\mathrm{n}}\right\}$, communication round E
        }
        \Ensure{ The Global model $ \left\{M_{global}\right\}$}
        \State /* Client Process */
            \For{$C_{i}$ from 1 to $N$}
               \State $ M_{i} \leftarrow \operatorname{GetGlobalModel}(round=i)$
                  // Get the global model and init client model
                \State $ M_{i} \leftarrow \operatorname{TrainLocalModel}(D_{i})$
                  // Train model $ M_{i}$ based Dataset $\left\{D_{i} \right\}$
                \State $\text { Server } \leftarrow \operatorname{Send}(M_{i})$
            \EndFor
         \State /* Server Process */
            \For{$e$ from 1 to $E$} 
                    \State \text { Store }$(\left\{M_{1}, M_{2}, \ldots M_{n}\right\}) \leftarrow   \operatorname{Receive}\left(M_{i}\right)$ // Receive Client Model

                    /* FedDRL Algorithm Process */
                    \State Train the Stage 1 Agent
                        \State Update the $\operatorname{SelectTrustClient}\left(.\right) 
                        $  Algorithm parameters   // According to Algorithm 1
                        \State $\left\{M_{a}, M_{b}, \ldots M_{\mathrm{k}}\right\} 
                        \leftarrow \operatorname{SelectTrustClient}\left(\left\{M_{1}, M_{2}, \ldots M_{n}\right\}\right)$
                    \State Train the Stage 2 Agent
                        \State Update the $\operatorname{AdaptCalculateWeight}\left(.\right)$ 
                         Algorithm parameters  // According to Algorithm 2
                        \State $\left\{W_{1}, W_{2}, \ldots W_{\mathrm{n}}\right\} \leftarrow \operatorname{AdaptCalculateWeight}\left(\left\{M_{a}, M_{b}, \ldots M_{k}\right\}\right)$
                    \State $ M_{global} \leftarrow \operatorname{FusionGlobalModel}\left(\left\{W_{1}, W_{2}, \ldots W_{n}\right\}\right)$
                    \State $ C_{i} \leftarrow \operatorname{SendGlobalModel} \left(M_{global}\right)$
            \EndFor
    \end{algorithmic}
\end{algorithm}

\section{EXPERIMENT}
\subsection{Experiment setup}
\subsubsection{Experiment datasets}
We evaluated the FedDRL framework using three distinct image classification datasets:

\textbf{Fashion-MNIST}: This dataset includes 60,000 training samples and 10,000 test samples, each a 28x28 grayscale image, classified into one of 10 categories.

\textbf{CIFAR-10}: The CIFAR-10 dataset comprises 60,000 32x32 colour images, evenly distributed across ten classes, each containing 6,000 images.

\textbf{CIFAR-100}: Similar in size to CIFAR-10 but with a broader spectrum, CIFAR-100 features 100 classes with 600 images each, totalling 60,000 colour images.

\begin{figure}
    \begin{minipage}[t]{0.33\linewidth}
        \centering
        \includegraphics[width=\textwidth]{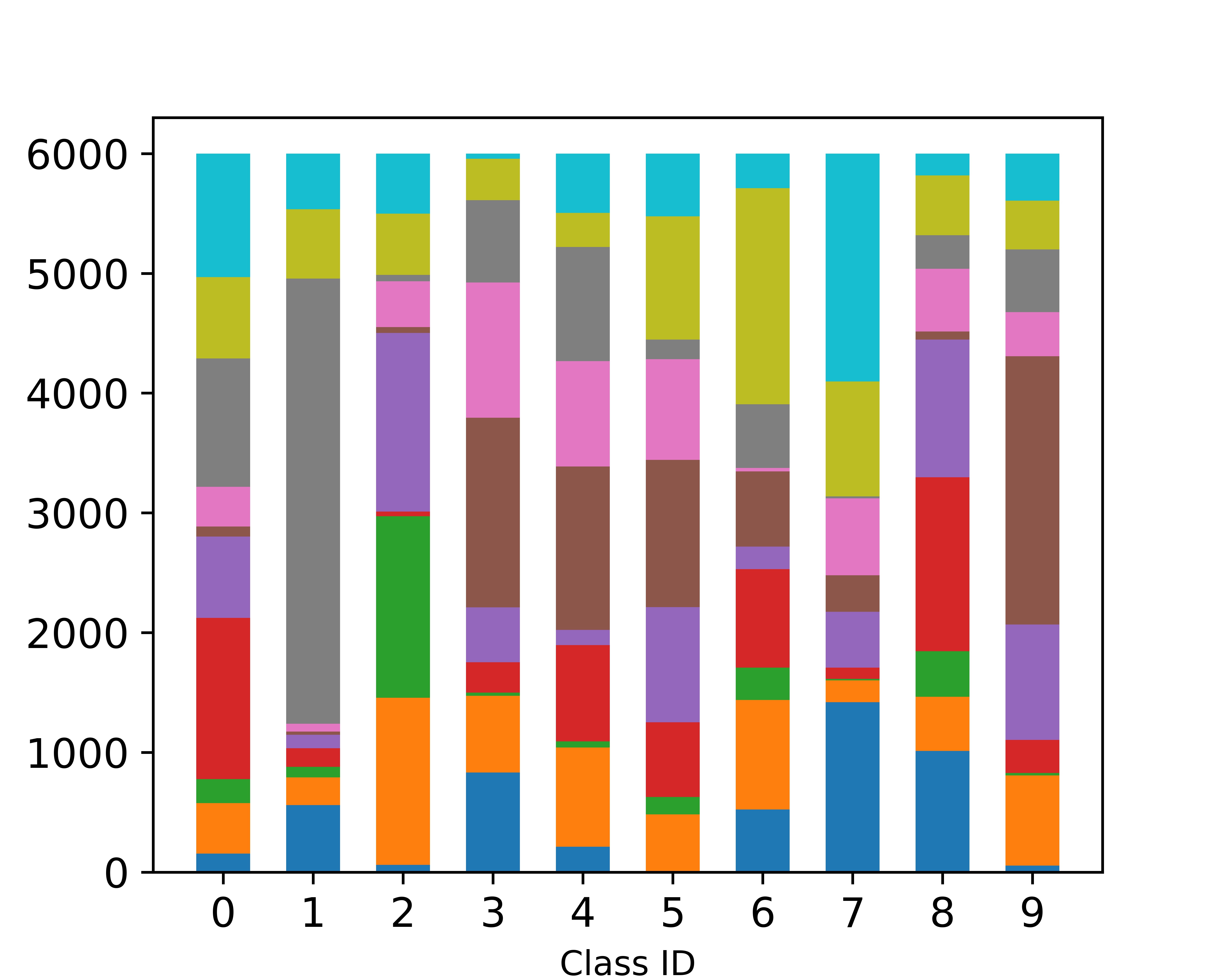}
        \centerline{(a) Fashion-MNIST}
    \end{minipage}%
    \begin{minipage}[t]{0.33\linewidth}
        \centering
        \includegraphics[width=\textwidth]{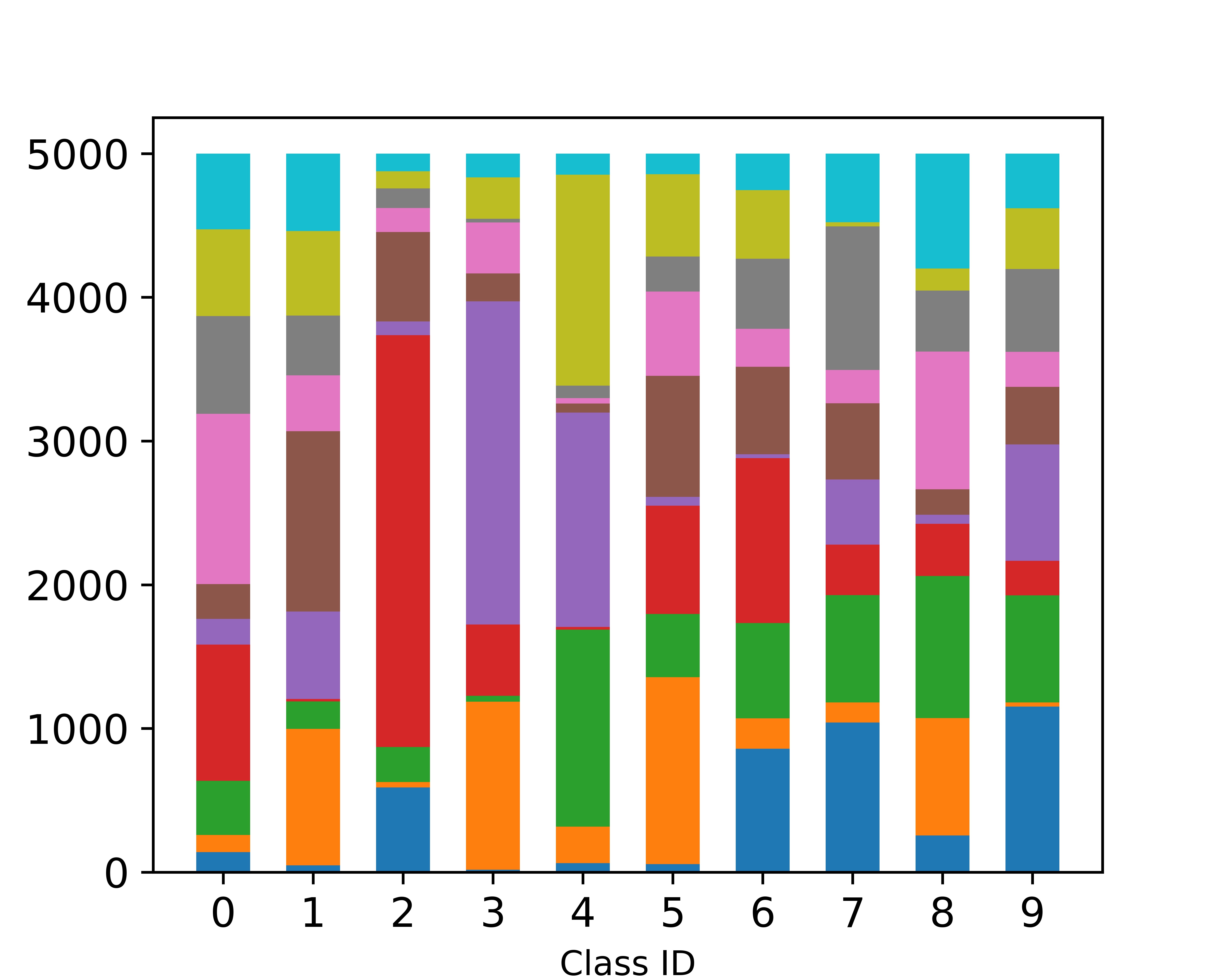}
        \centerline{(b) CIFAR-10}
    \end{minipage}
        \begin{minipage}[t]{0.33\linewidth}
        \centering
        \includegraphics[width=\textwidth]{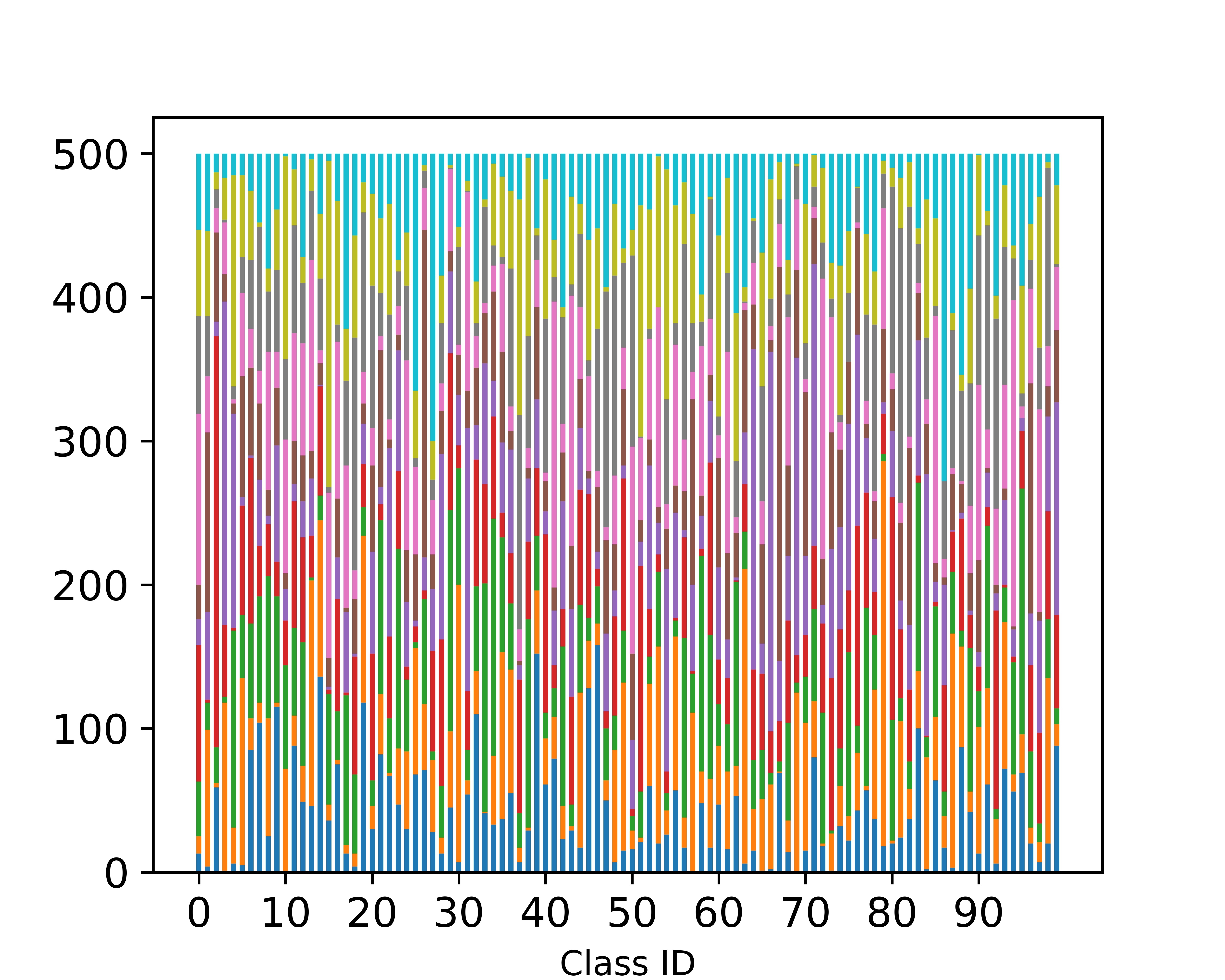}
        \centerline{(c) CIFAR-100}
    \end{minipage}
    \caption{The non-iid distribution of 10 clients(alpha=1).}
\label{fig-Non-iidDistribution}
\end{figure}

\textbf{Data Set Partitioning}: For simulating non-IID data distribution among clients. We utilized the Dirichlet function to segregate data across various clients in the open-source dataset. This method can partition the data for each client by adjusting the alpha parameter. As the alpha parameter approaches zero, clients’ data distributions are skewed towards specific classes within the dataset. Conversely, as alpha increases towards infinity. 
Using the CIFAR-10 dataset as a case study, we set alpha to 1, thereby dividing the three datasets among ten clients. In the figure, Different categories are represented by distinct colours, and the length of each segment within the graphs reflects the sample count within that category. The resulting data distribution is illustrated in Figure \ref{fig-Non-iidDistribution}.

\subsubsection{Comparison of Methods}
We contrasted the FedDRL algorithm with two established federated learning approaches.

\textbf{FedAvg}\cite{FedAvg}: Serving as the foundational benchmark in federated learning, the FedAvg method determines the weight of each client model based on the proportion of samples contributed by the client relative to the aggregate sample size.

\textbf{FedProx}\cite{FedProx}: Enhancing the FedAvg approach, FedProx incorporates a regularization term within the client model, thereby refining federated learning performance.

\subsubsection{Experimental Metrics}
We employed accuracy as the metric to gauge the performance of the global model in multi-classification tasks and across individual clients. Assuming n clients engage in model fusion with m communication rounds, the accuracy of the global model in the t-th round is denoted as $A_{\text{global}}^{(t)}$. The collective global model accuracies across all rounds are represented as follows:
\begin{equation}\label{eqn-24}
A_{\text{global}}=\{A_{\text{global}}^{1}, A_{\text{global}}^{2}, \ldots, A_{\text{global}}^{m}\}
\end{equation}

We denote the accuracy of the $c$-th client's model as $A_c$. Additionally, we document each client model's accuracy per round, compiling these as follows:
\begin{equation}\label{eqn-25}
A_{\text{c}}=\{A_{\text{c}}^{1}, A_{\text{c}}^{2}, \ldots, A_{\text{c}}^{m}\}, c \in [1,2,\ldots,n]
\end{equation}

\subsubsection{Experimental Configuration}
\textbf{Hardware Configuration}: The experiments were conducted on a workstation equipped with an Intel i9-12900k CPU, 64GB RAM, and an NVIDIA RTX3090 GPU.

\textbf{Software Configuration}: We utilized two distinct frameworks for the Federated Learning and Reinforcement Learning experiments. Federated Learning trials were carried out using FedBolt, our custom-built framework, enabling simulation of varied client numbers and data distributions. For reinforcement learning model training, we employed the stablebaseline3 framework, designing two distinct algorithms for trusted client selection and model weight assignment.

\textbf{Network Setup}: We implemented different network architectures tailored to each dataset. For CIFAR-10 and CIFAR-100, a 6-layer CNN was utilized for model training. Conversely, a 4-layer MLP was developed for the Fashion-MNIST dataset.

\textbf{Agent Network Setup}: Implementing a staged reinforcement learning strategy necessitated the training of two distinct agents. The initial phase, adhering to Section \ref{A2Csubsection}, utilizes the A2C algorithm, with each worker and the central node comprising a 6-layer MLP actor and critic network. For the second phase, the TD3 algorithm outlined in Section \ref{TD3subsection} was employed for agent training, where each module within the TD3 setup incorporates a 6-layer MLP, with further details available in Section \ref{TD3subsection}.

\subsection{Experimental Results}

We evaluate the FedDRL framework through four experiments: client attack scenarios, low-quality model fusion, hybrid scenarios, and multi-agent training efficiency. The client attack experiments assess the efficacy of the trustworthy client selection algorithm (stage 1). The low-quality model fusion experiment examines the adaptive weight calculation method (stage 2). The hybrid experiments, combining client attacks and low-quality model elements, validate the comprehensive performance of FedDRL. The final experiment focuses on the training efficiency of multi-agents.

\subsubsection{Malicious Client Attack Experiment}

In this experiment, we define three types of client-side attacks in federated learning to evaluate our FedDRL framework under adversarial conditions. The experiment spans different client numbers and attack types across three datasets, detailed in Table \ref{table-AttackExperimental-Setting}.

\textbf{Type 1}: The client directly uploads the initialized model or makes the model accuracy less than 10\% by modifying the model's hyperparameters.

\textbf{Type 2}:  We use falsified data to perform the attack. We use a certain percentage of forged data to participate in model training (e.g., mix the CIFAR-10 dataset with $80\%$ of CIFAR-100 data and generate these CIFAR-100 data labels as CIFAR-10 corresponding label types). We conduct the attack by faking sample data to train the client's local model, thus reducing the client model's accuracy.

\textbf{Type 3}: We select some clients to simulate the attack and divide the training process of these clients into standard and attack rounds. In the standard round, each client does not perform the attack behavior. Instead, each client deliberately uploads the prepared malicious model in the attack round. We also set that these clients alternately initiate the attack behavior.

\begin{table}

\centering
\caption{Experimental setup for malicious attack scenarios}
\resizebox{\textwidth}{!}{\begin{tabular}{ccccccc} 
\toprule %
\multirow{1}{*}{Number} & 
\multicolumn{1}{c}{Attack Type} &
\multicolumn{1}{c}{Malicious ID} &
\multicolumn{1}{c}{Number of samples}&
\multicolumn{1}{c}{Accuracy of models ($\leq$)} \\
\midrule %

\multirow{3}{*}{Clients=5} 
&Type 1 & Client1 & 7750 & $A \leq 10\%$ \\
&Type 2 & Client1 & 7750 & $10\% \leq A\leq 20\%$\\
&Type 3 & Client1 & 7750 & Attack round $A \leq 10\%$\\
\hline

\multirow{3}{*}{Clients=10} 
&Type 1 & Client1,Client6 & 4222, 4938 & $A\leq 10\%$ 
\\
&Type 2 & Client1,Client6 & 4222, 4938 & $10\% \leq A\leq 20\%$ 
\\
&Type 3 & Client1,Client6 & 4222, 4938 &Attack round $A \leq 10\%$ \\
\hline

\multirow{3}{*}{Clients=15} 
&Type 1 & Client1,Client6,Client11 & 3670, 3314, 4454 & $A\leq 10\%$ 
\\
&Type 2 & Client1,Client6,Client11 & 3670, 3314, 4453  & $10\% \leq A\leq 20\%$ 
\\
&Type 3 & Client1,Client6,Client11 & 3670, 3314, 4453 &Attack round $A \leq 15\%$
\\

\bottomrule %
\end{tabular}}
\label{table-AttackExperimental-Setting}
\end{table}

According to the experimental setup, we compared the FedDRL algorithm with the FedAvg and FedProx. In the attack experiments, we set the total number of communication rounds to 100 rounds, and each client performs local model training with one epoch. To show the attack behavior of each client and the accuracy of different algorithms more detail, we counted the accuracy of each client's local model and the accuracy of the server-side global model in each communication round. The specific experimental results are shown in table \ref{table-CleintAttack-Accuracy}.

\begin{table}

\centering
\caption{Accuracy of each algorithm under different malicious attack scenarios}
\resizebox{\textwidth}{!}{
\begin{tabular}{|l|c|lll|lll|lll|} 
\toprule %
\multicolumn{1}{c|}{\multirow{2}{*}{\textbf{DataSet}}}               
& \multirow{2}{*}{\textbf{Method}} 
& \multicolumn{3}{c|}{Clients=5}
& \multicolumn{3}{c|}{Clients=10}                                     
& \multicolumn{3}{c}{Clients=15}                                                 
\\ \cline{3-11} 

\multicolumn{1}{c|}{}&                                  
& \multicolumn{1}{l}{Type 1} & \multicolumn{1}{l}{Type 2} & Type 3                  
& \multicolumn{1}{l}{Type 1} & \multicolumn{1}{l}{Type 2} & Type 3                  
& \multicolumn{1}{l}{Type 1} & \multicolumn{1}{l}{Type 2} &  \multicolumn{1}{l}{Type 3}                  
\\ 
\hline
\multicolumn{1}{c|}{\multirow{3}{*}{\begin{tabular}[c]{@{}c@{}}Fashion\\ -MNIST\end{tabular}}}
& FedAvg                           
& \multicolumn{1}{c}{0.862}     & \multicolumn{1}{c}{0.875}     & \multicolumn{1}{c|}{0.863}
& \multicolumn{1}{c}{0.792}     & \multicolumn{1}{c}{0.881}     & \multicolumn{1}{c|}{0.821} 
& \multicolumn{1}{c}{0.776}     & \multicolumn{1}{c}{0.878}     & \multicolumn{1}{c}{0.812} 
\\ 

\multicolumn{1}{c|}{}     

& FedProx                           
& \multicolumn{1}{c}{0.873}     & \multicolumn{1}{c}{\pmb{0.884}}    & \multicolumn{1}{c|}{0.864} 
& \multicolumn{1}{c}{0.791}     & \multicolumn{1}{c}{0.882}     & \multicolumn{1}{c|}{0.824} 
& \multicolumn{1}{c}{0.764}     & \multicolumn{1}{c}{0.879}     & \multicolumn{1}{c}{0.811} 
\\ 

\multicolumn{1}{c|}{}     
& Ours                          
& \multicolumn{1}{c}{\pmb{0.885}}     & \multicolumn{1}{c}{0.878}    & \multicolumn{1}{c|}{\pmb{0.883}} 
& \multicolumn{1}{c}{\pmb{0.877}}     & \multicolumn{1}{c}{\pmb{0.886}}     & \multicolumn{1}{c|}{\pmb{0.887}} 
& \multicolumn{1}{c}{\pmb{0.881}}     & \multicolumn{1}{c}{\pmb{0.886}}     & \multicolumn{1}{c}{\pmb{0.882}} 
                      
\\ 
\hline

\multicolumn{1}{c|}{\multirow{3}{*}{Cifar10}}                                               & FedAvg                           
& \multicolumn{1}{c}{0.596}     & \multicolumn{1}{c}{0.691}    & \multicolumn{1}{c|}{0.751} 
& \multicolumn{1}{c}{0.335}     & \multicolumn{1}{c}{0.681}     & \multicolumn{1}{c|}{0.314} 
& \multicolumn{1}{c}{0.139}     & \multicolumn{1}{c}{0.664}     & \multicolumn{1}{c}{0.197} 
\\ 

\multicolumn{1}{c|}{}     

& FedProx                           
& \multicolumn{1}{c}{0.586}     & \multicolumn{1}{c}{\pmb{0.732}}    & \multicolumn{1}{c|}{\pmb{0.775}} 
& \multicolumn{1}{c}{0.331}     & \multicolumn{1}{c}{\pmb{0.716}}     & \multicolumn{1}{c|}{0.363} 
& \multicolumn{1}{c}{0.122}     & \multicolumn{1}{c}{\pmb{0.719}}     & \multicolumn{1}{c}{0.202} 
\\ 

\multicolumn{1}{c|}{}     

& Ours                          
& \multicolumn{1}{c}{\pmb{0.731}}     & \multicolumn{1}{c}{0.701}    & \multicolumn{1}{c|}{0.747} 
& \multicolumn{1}{c}{\pmb{0.694}}     & \multicolumn{1}{c}{0.711}     & \multicolumn{1}{c|}{\pmb{0.727}} 
& \multicolumn{1}{c}{\pmb{0.679}}     & \multicolumn{1}{c}{0.702}     & \multicolumn{1}{c}{\pmb{0.689}} 
\\ \hline

\multicolumn{1}{c|}{\multirow{3}{*}{Cifar100}}                                              & FedAvg                           
& \multicolumn{1}{c}{0.376}     & \multicolumn{1}{c}{0.412}    & \multicolumn{1}{c|}{0.298} 
& \multicolumn{1}{c}{0.273}     & \multicolumn{1}{c}{0.416}     & \multicolumn{1}{c|}{0.287} 
& \multicolumn{1}{c}{0.162}     & \multicolumn{1}{c}{0.398}     & \multicolumn{1}{c}{0.176} 
\\ 

\multicolumn{1}{c|}{}     

& FedProx                           
& \multicolumn{1}{c}{0.398}     & \multicolumn{1}{c}{\pmb{0.442}}    & \multicolumn{1}{c|}{0.321} 
& \multicolumn{1}{c}{0.223}     & \multicolumn{1}{c}{\pmb{0.436}}     & \multicolumn{1}{c|}{0.208} 
& \multicolumn{1}{c}{0.172}     & \multicolumn{1}{c}{\pmb{0.426}}     & \multicolumn{1}{c}{0.183} 
\\ 

\multicolumn{1}{c|}{}     

& Ours                          
& \multicolumn{1}{c}{\pmb{0.412}}     & \multicolumn{1}{c}{0.438}    & \multicolumn{1}{c|}{\pmb{0.431}} 
& \multicolumn{1}{c}{\pmb{0.426}}     & \multicolumn{1}{c}{0.432}     & \multicolumn{1}{c|}{\pmb{0.421}} 
& \multicolumn{1}{c}{\pmb{0.423}}     & \multicolumn{1}{c}{0.412}     & \multicolumn{1}{c}{\pmb{0.422}}                   
\\
\bottomrule %
\end{tabular}}
\label{table-CleintAttack-Accuracy}
\end{table}

To show the effect of the FedDRL algorithm on global model fusion at each communication round, we conducted experiments using the CIFAR10 dataset on 5, 10, and 15 clients. We compared FedDRL with the FedAvg and FedProx algorithms for global model accuracy. 

We analyze the experimental results for different numbers of client models and different client data. In attack type 1,  In malicious data attack type 2, our algorithm outperforms the FedAvg algorithm and slightly underperforms the FedProx algorithm alone. In the attack type 3 scenario, our algorithm outperforms the comparison algorithm in most cases, especially when multiple malicious clients are involved in model fusion.

\begin{figure}

    \begin{minipage}[t]{0.33\linewidth}
        \centering
        \includegraphics[width=\textwidth]{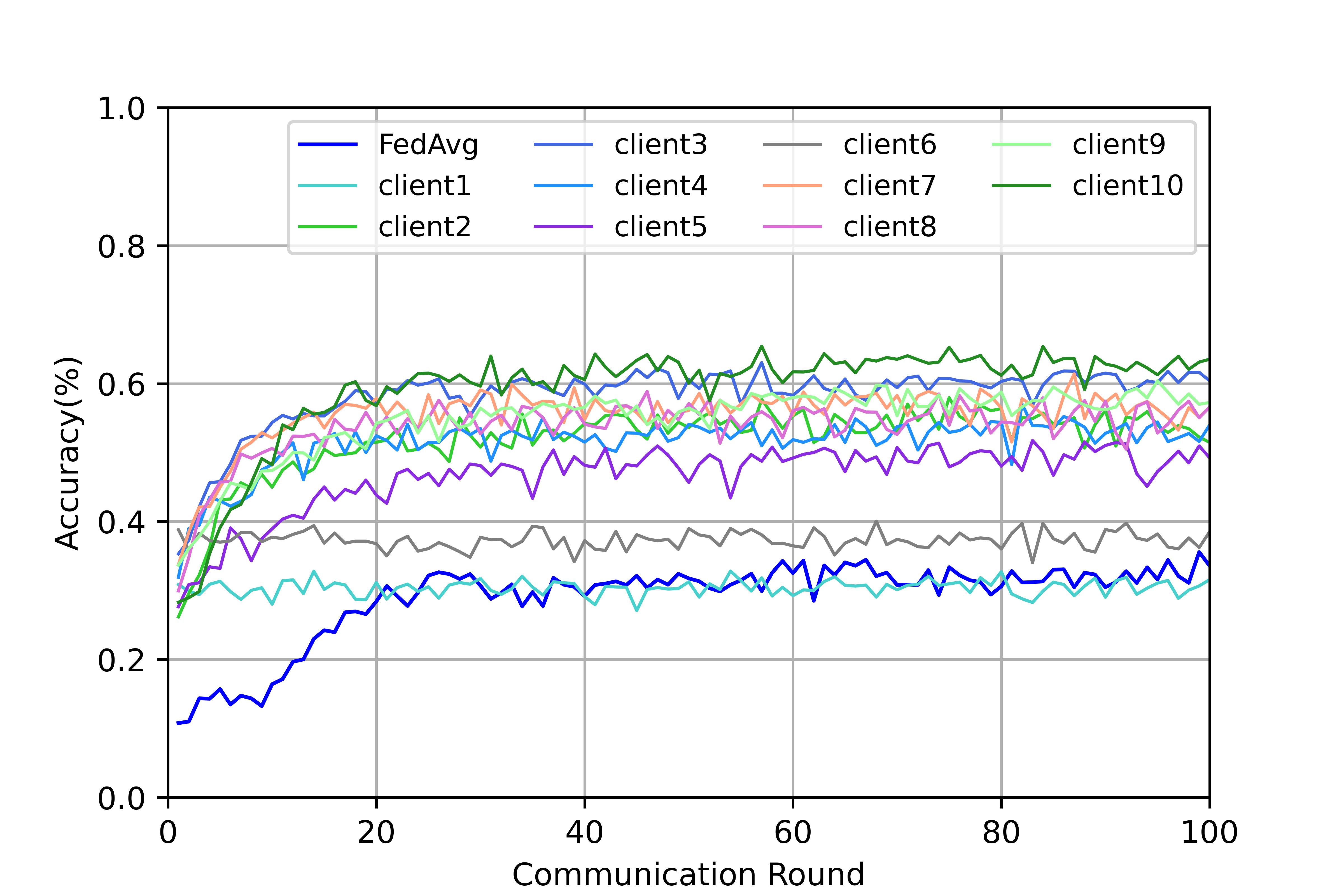}
        \centerline{(a) 10 clients FedAvg}
    \end{minipage}%
    \begin{minipage}[t]{0.33\linewidth}
        \centering
        \includegraphics[width=\textwidth]{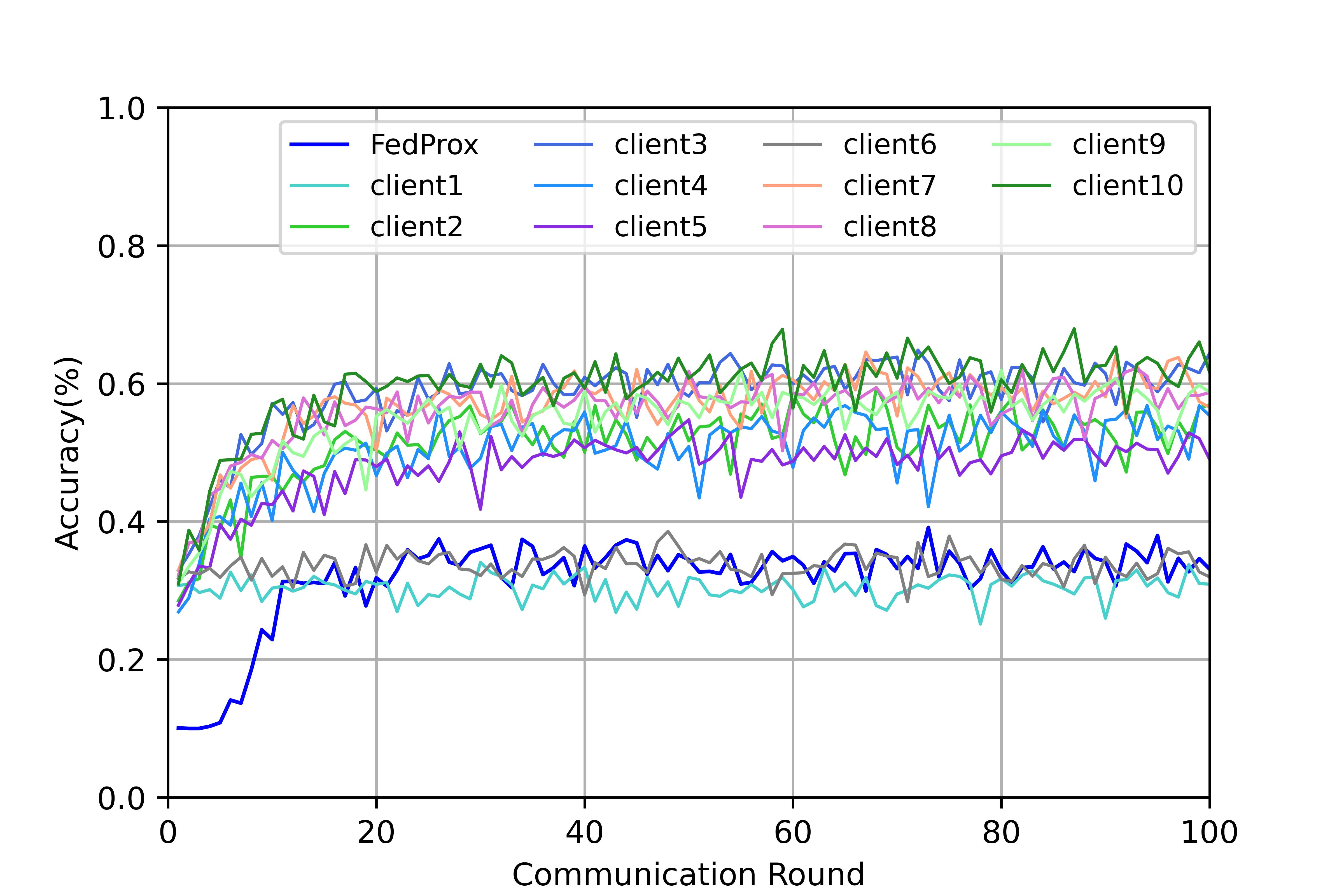}
        \centerline{(b) 10 clients FedProx}
    \end{minipage}
    \begin{minipage}[t]{0.33\linewidth}
        \centering
        \includegraphics[width=\textwidth]{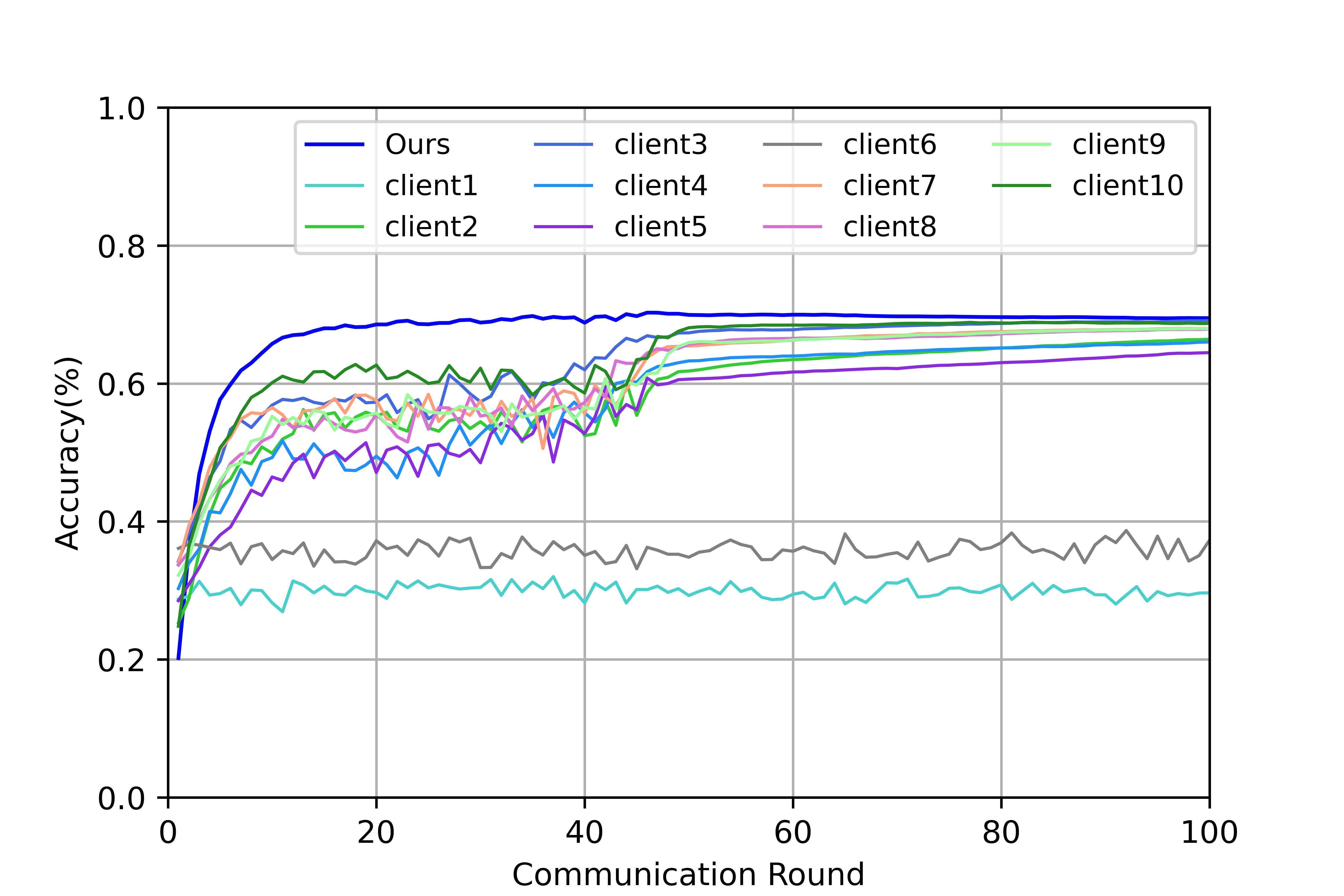}
        \centerline{(c) 10 clients FedDRL}
    \end{minipage}
      \begin{minipage}[t]{0.33\linewidth}
        \centering
        \includegraphics[width=\textwidth]{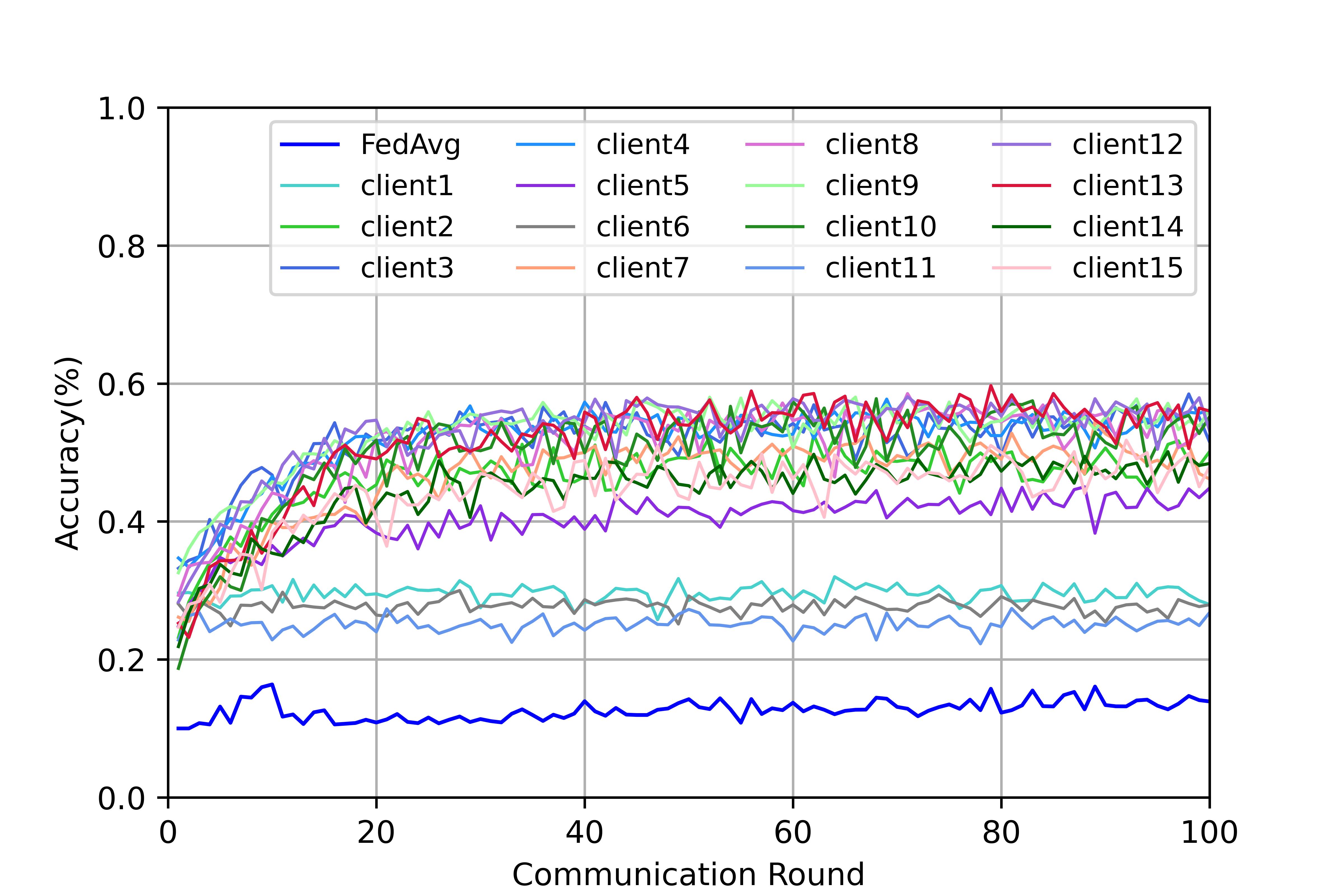}
        \centerline{(d) 15 clients FedAvg }
    \end{minipage}%
    \begin{minipage}[t]{0.33\linewidth}
        \centering
        \includegraphics[width=\textwidth]{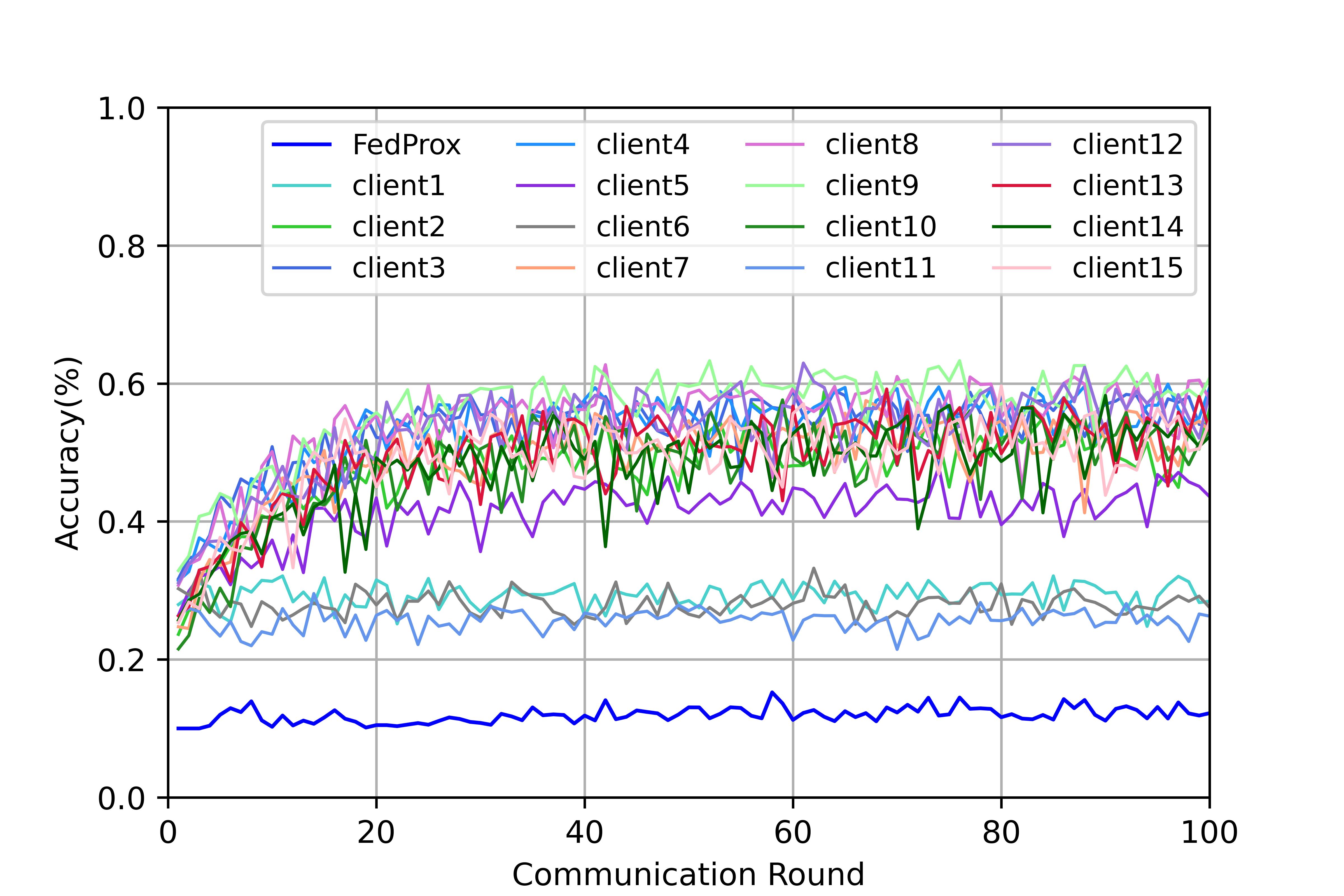}
        \centerline{(e) 15 clients FedProx }
    \end{minipage}
    \begin{minipage}[t]{0.33\linewidth}
        \centering
        \includegraphics[width=\textwidth]{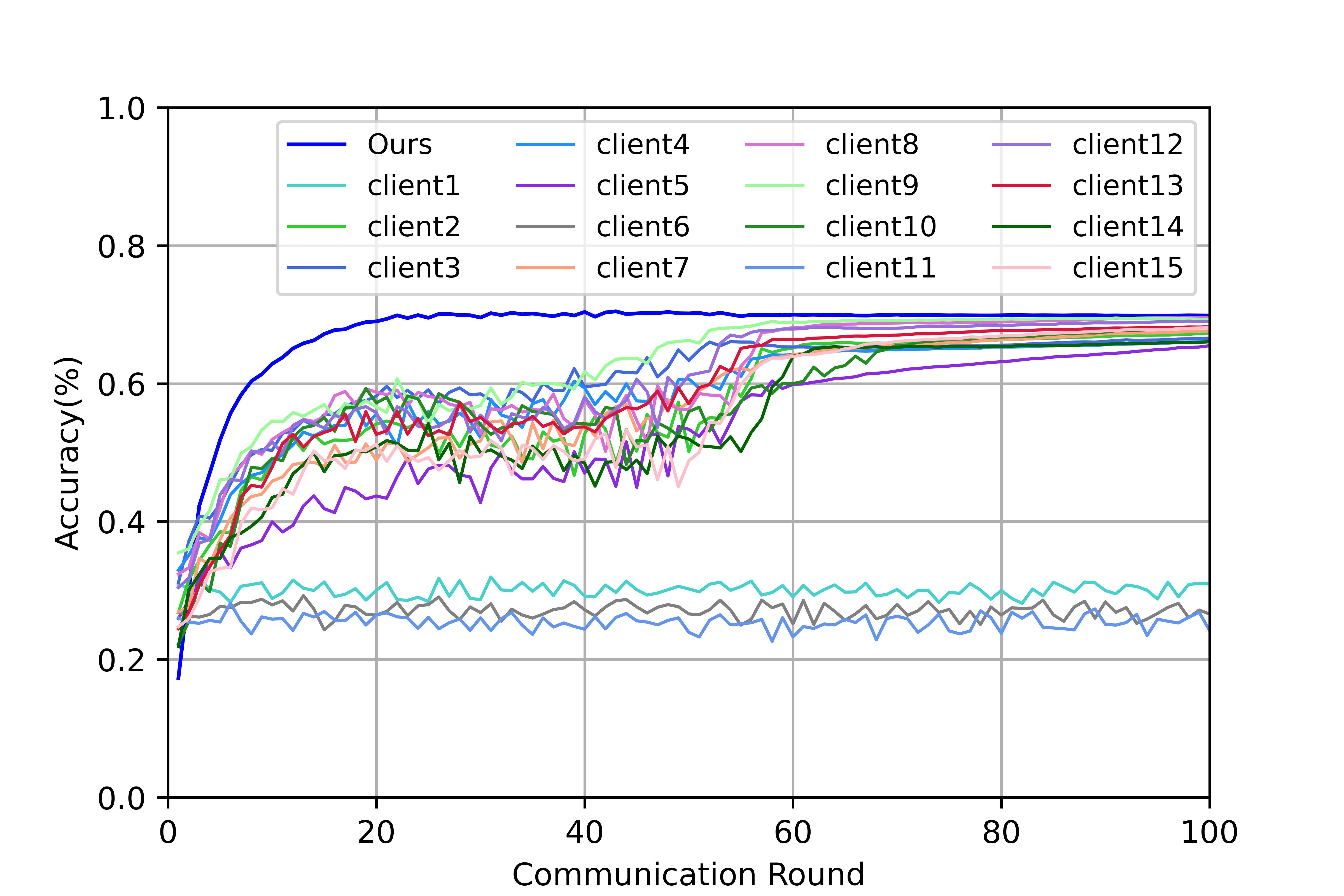}
        \centerline{(f) 15 clients FedDRL }
    \end{minipage}
\caption{The accuracy of the global model for different number of client in attack type 1}
\label{fig-attack1-acc}
\end{figure}

To show the relationship between the global and client model's accuracy in each attack scenario. We conducted more detailed experiments on the Cifar10 dataset. 

In attack type 1, the global model accuracy plummets with increasing malicious clients under FedAvg and FedProx, dropping below 40\% and 20\% in 10 and 15 client setups, respectively. Conversely, FedDRL's dynamic client selection maintains higher reliability. However, Our trained agent can dynamically select trusted clients for model fusion and eliminate malicious models from participating, so our algorithm has higher reliability. The experimental results are shown in Figure \ref{fig-attack1-acc}.

\begin{figure}
    \begin{minipage}[t]{0.33\linewidth}
        \centering
        \includegraphics[width=\textwidth]{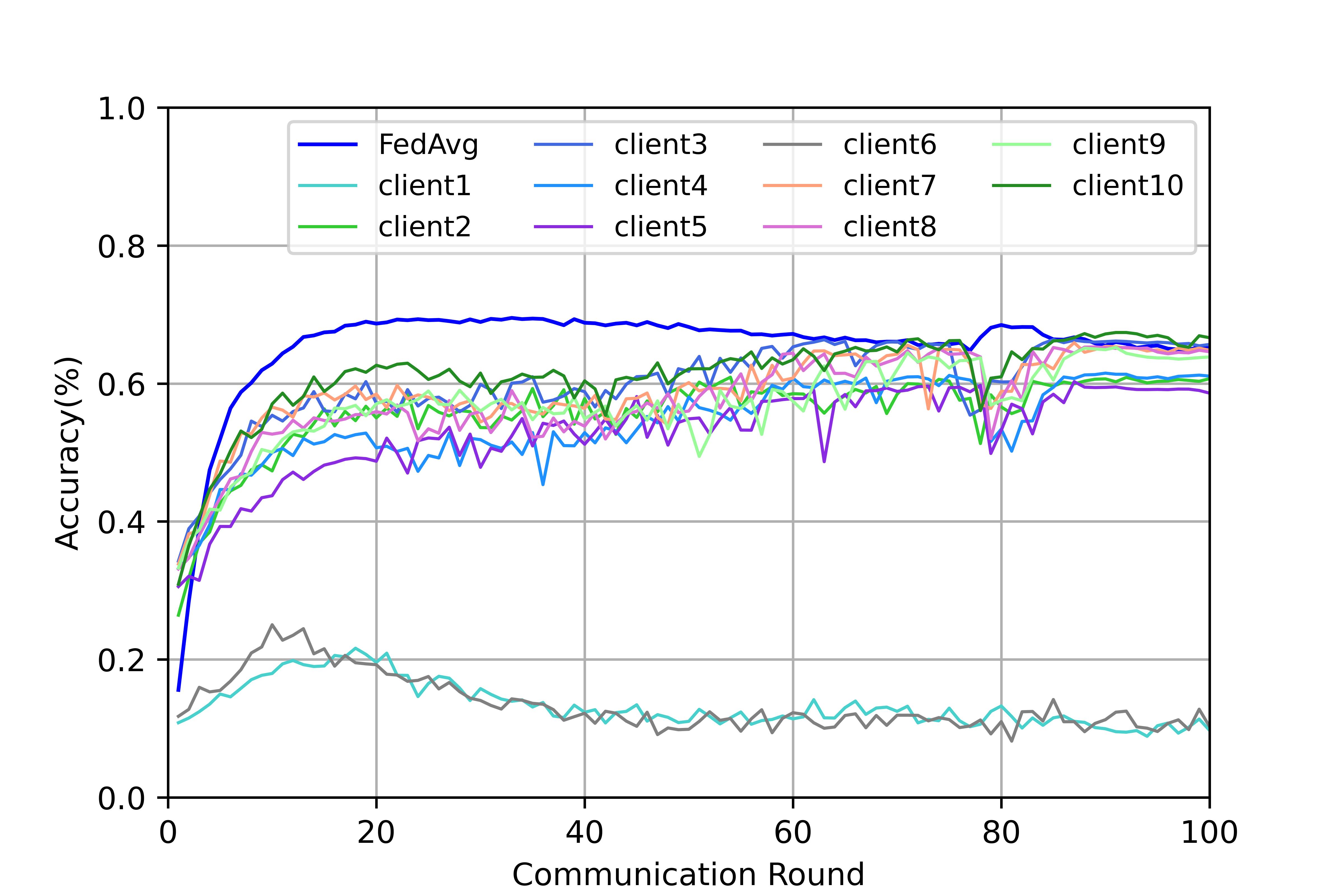}
        \centerline{(a) 10 clients FedAvg }
    \end{minipage}%
    \begin{minipage}[t]{0.33\linewidth}
        \centering
        \includegraphics[width=\textwidth]{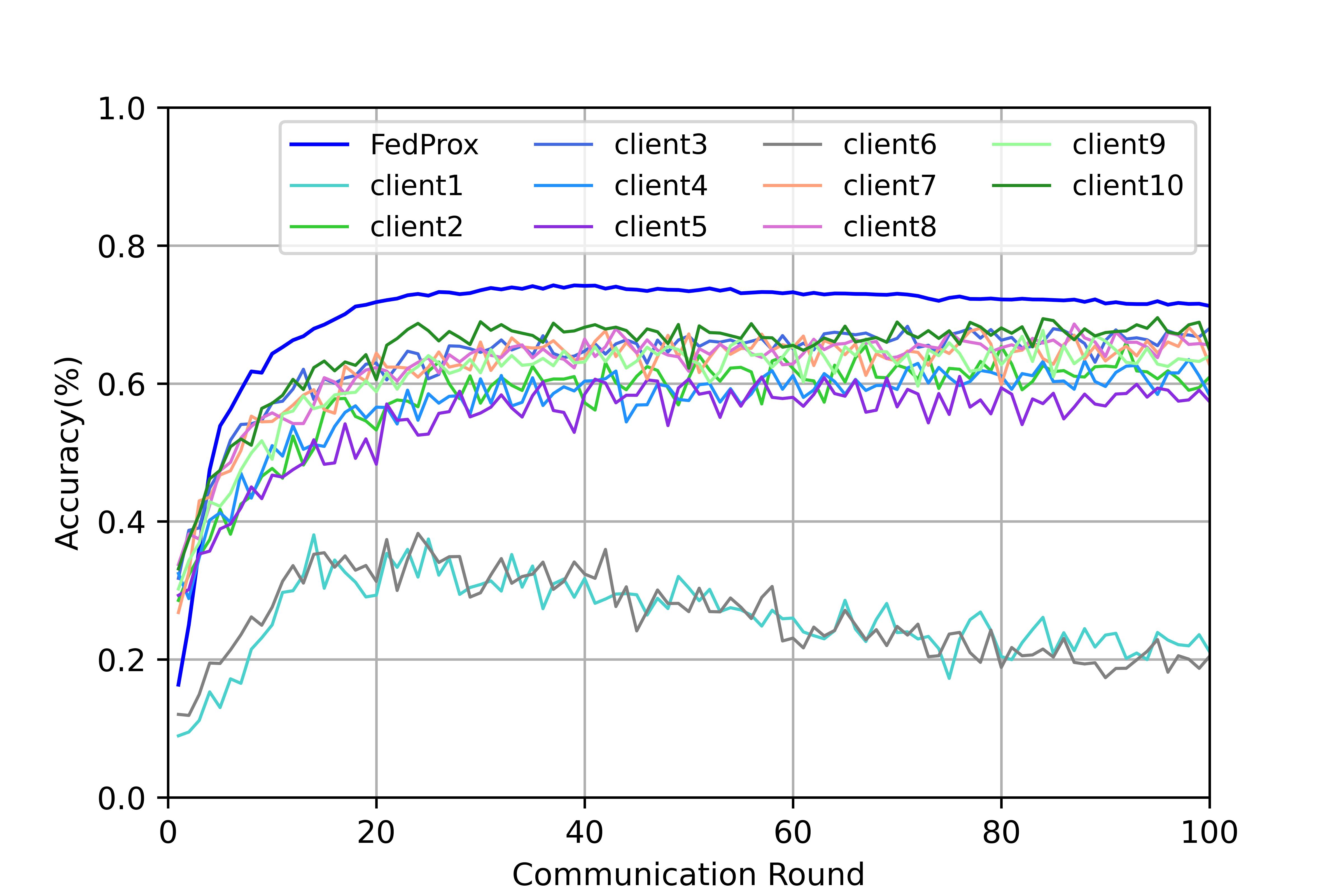}
        \centerline{(b) 10 clients FedProx }
    \end{minipage}
    \begin{minipage}[t]{0.33\linewidth}
        \centering
        \includegraphics[width=\textwidth]{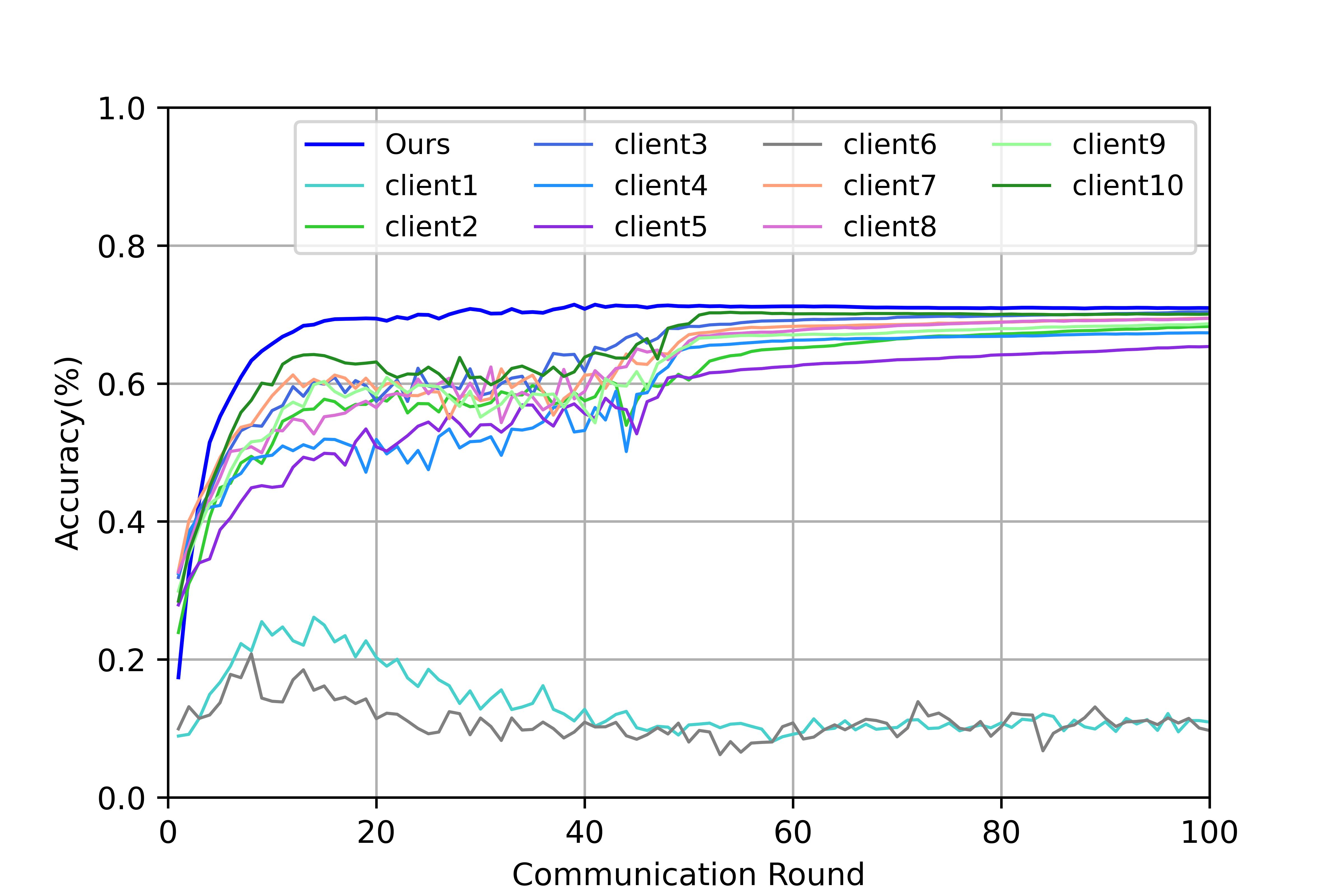}
        \centerline{(c) 10 clients FedDRL }
    \end{minipage}
    \begin{minipage}[t]{0.33\linewidth}
        \centering
        \includegraphics[width=\textwidth]{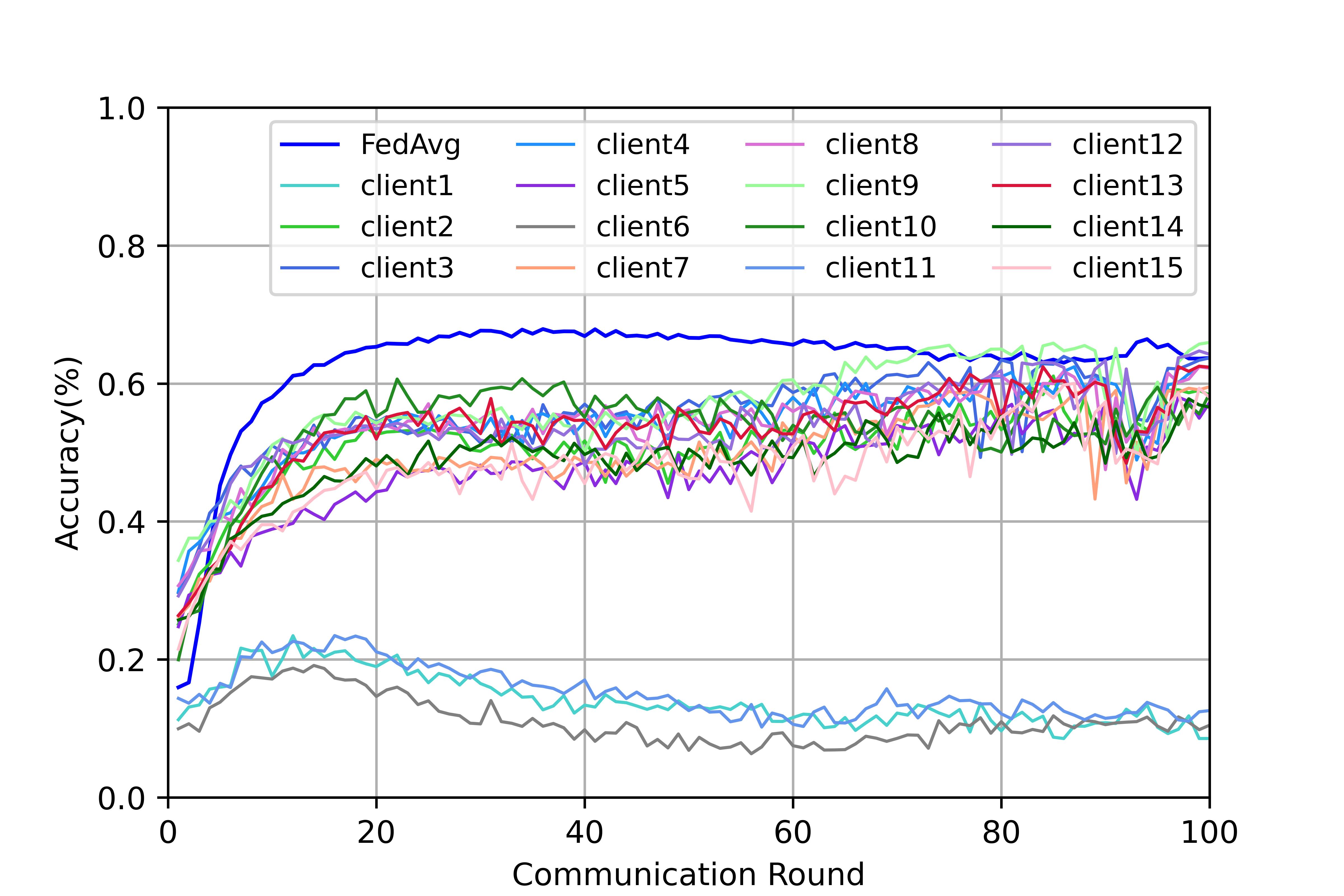}
        \centerline{(d) 15 clients FedAvg }
    \end{minipage}%
    \begin{minipage}[t]{0.33\linewidth}
        \centering
        \includegraphics[width=\textwidth]{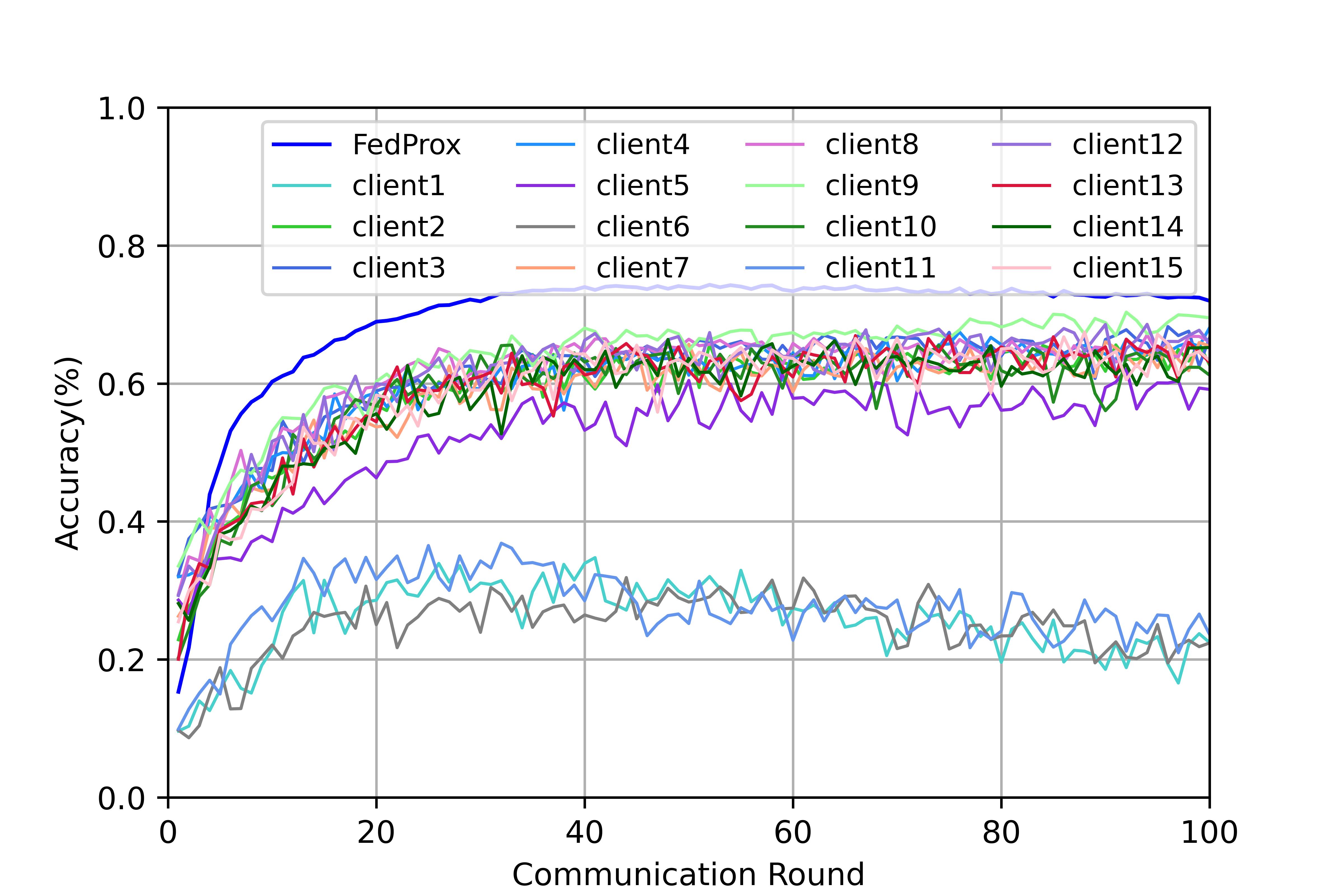}
        \centerline{(e) 15 clients FedProx }
    \end{minipage}
    \begin{minipage}[t]{0.33\linewidth}
        \centering
        \includegraphics[width=\textwidth]{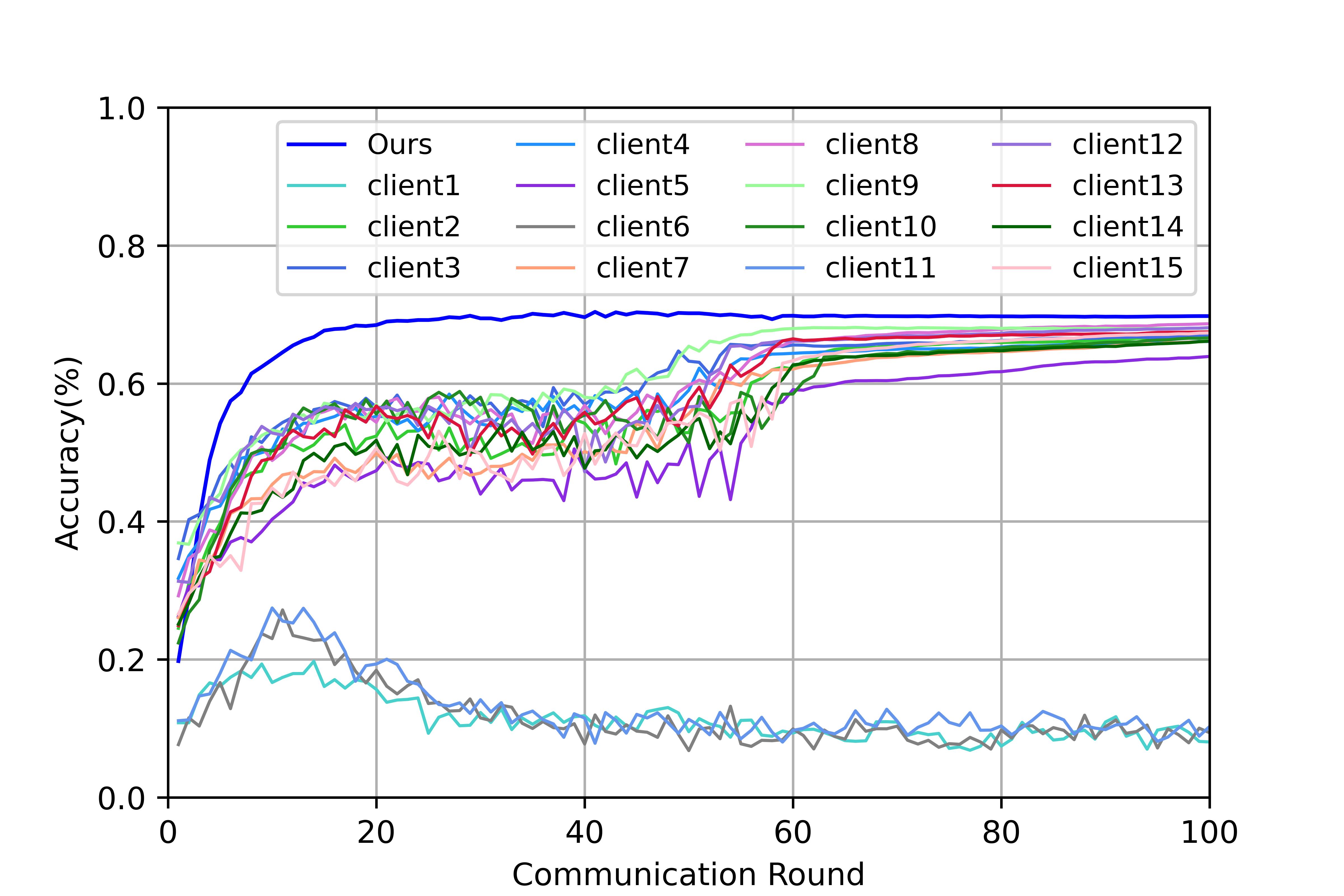}
        \centerline{(f) 15 clients FedDRL}
    \end{minipage}
\caption{The accuracy of the global model for different number of client in attack type 2}
\label{fig-attack2-acc}
\end{figure}

In attack type 2, our algorithm is better than FedAvg but lower than FedProx. The FedProx algorithm uses control parameters to force the models of each client to converge to the global model, which will improve the global model's accuracy by improving the malicious model's accuracy to some extent. Our trained agent will filter out low-accuracy models to participate in the fusion after several communication rounds.
The experimental results are shown in Figure \ref{fig-attack2-acc}.

\begin{figure}
    \begin{minipage}[t]{0.33\linewidth}
        \centering
        \includegraphics[width=\textwidth]{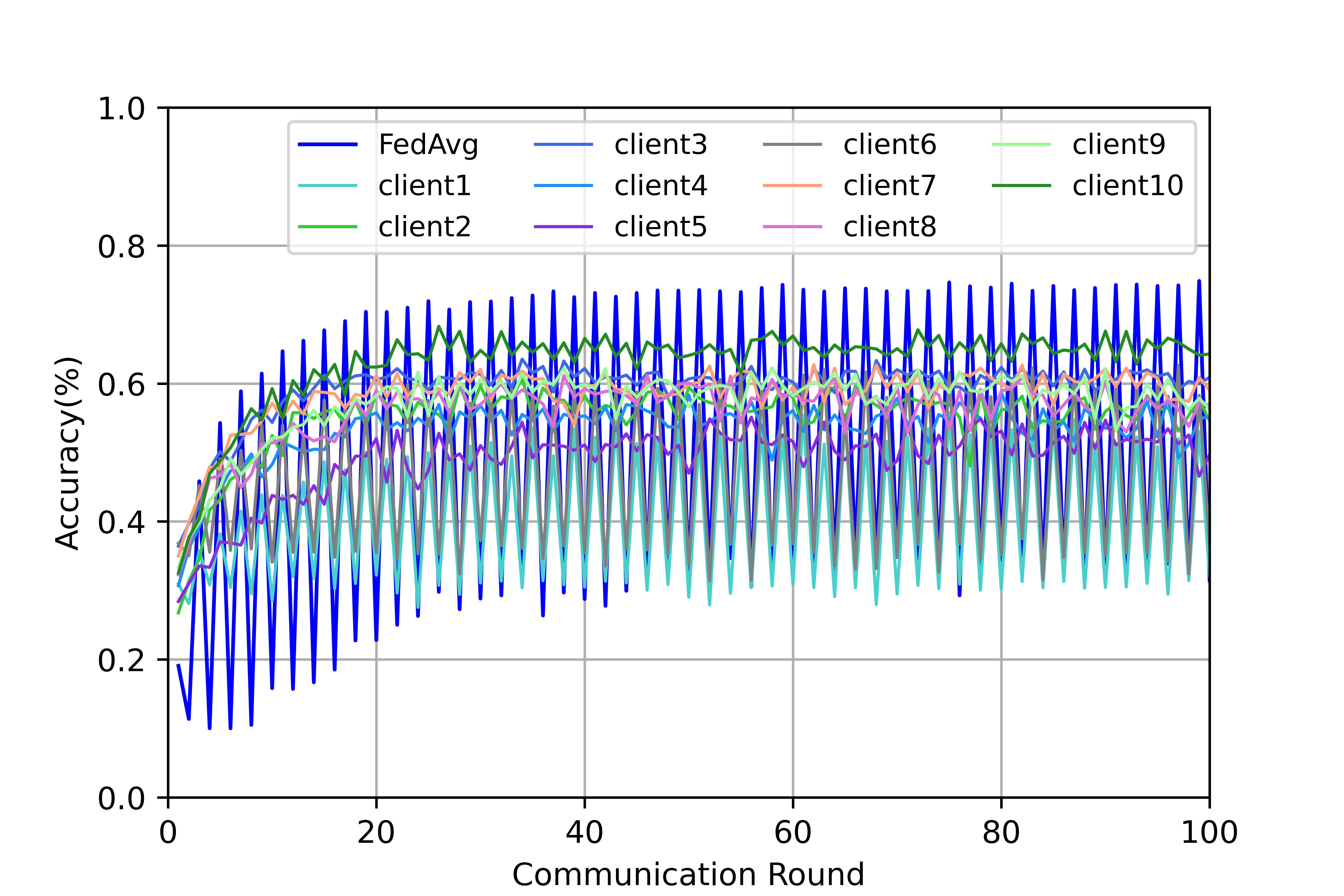}
        \centerline{(a) 10 clients FedAvg }
    \end{minipage}%
    \begin{minipage}[t]{0.33\linewidth}
        \centering
        \includegraphics[width=\textwidth]{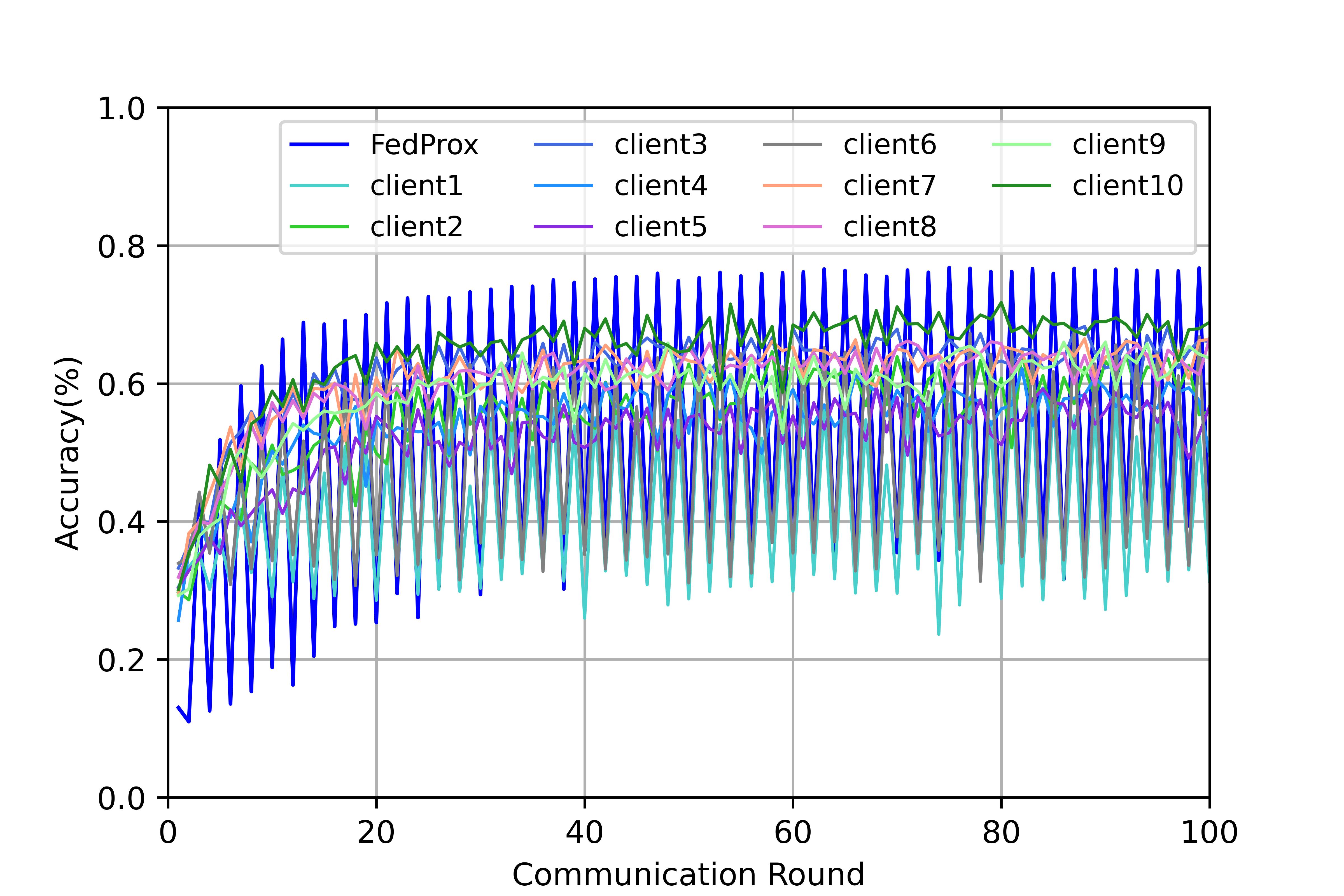}
        \centerline{(b) 10 clients FedProx }
    \end{minipage}
    \begin{minipage}[t]{0.33\linewidth}
        \centering
        \includegraphics[width=\textwidth]{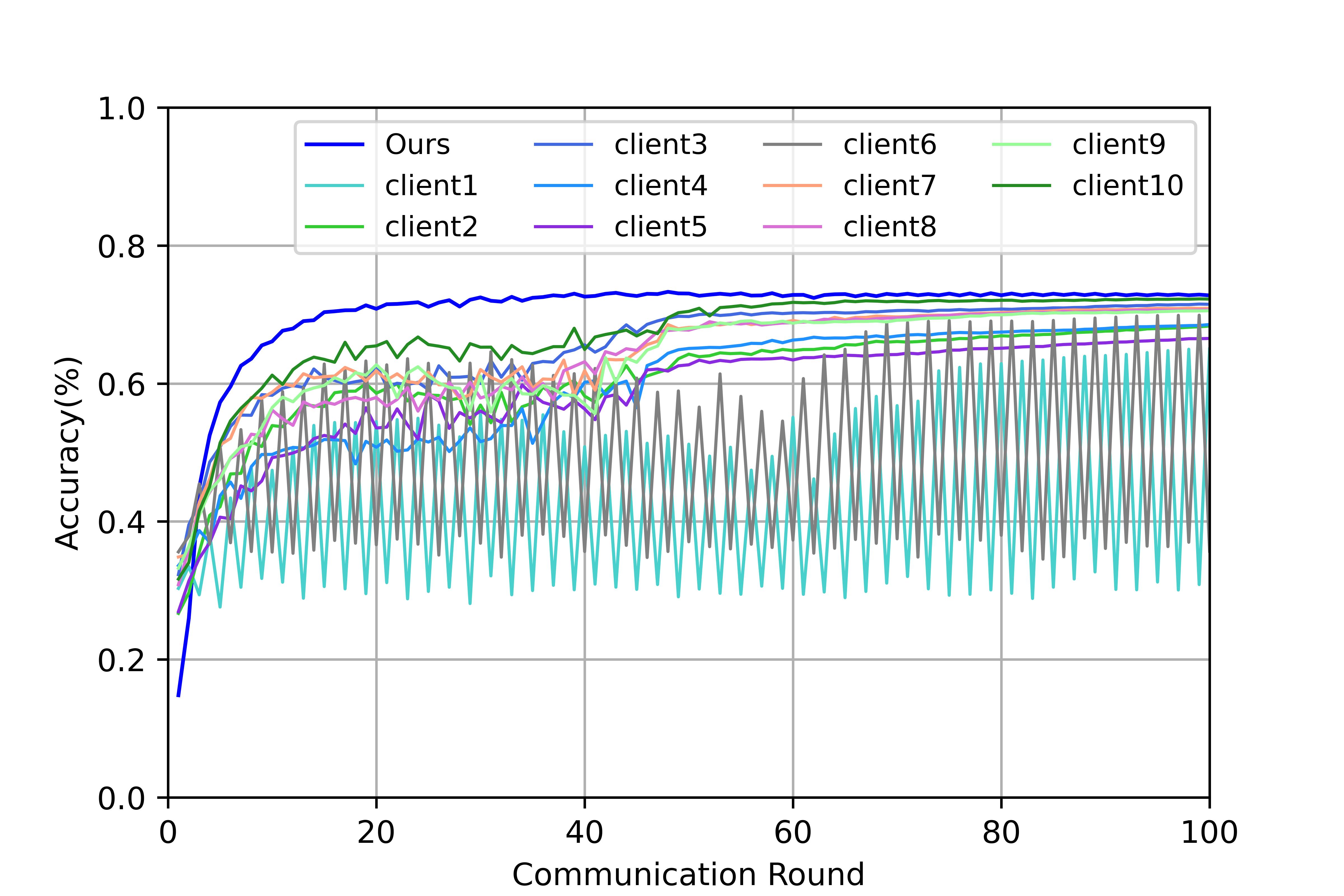}
        \centerline{(c) 10 clients FedDRL }
    \end{minipage}

    \begin{minipage}[t]{0.33\linewidth}
        \centering
        \includegraphics[width=\textwidth]{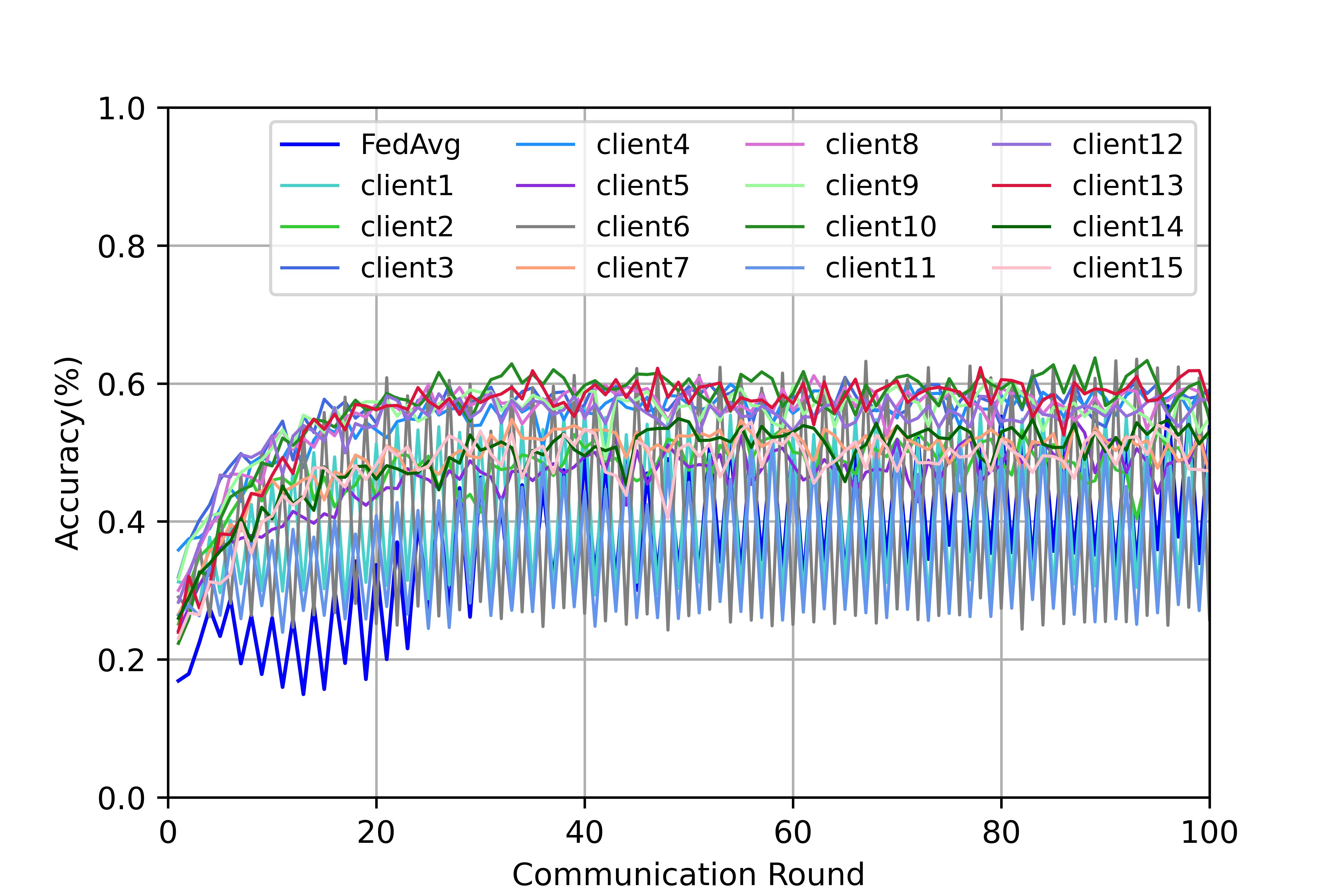}
        \centerline{(d) 15 clients FedAvg }
    \end{minipage}%
    \begin{minipage}[t]{0.33\linewidth}
        \centering
        \includegraphics[width=\textwidth]{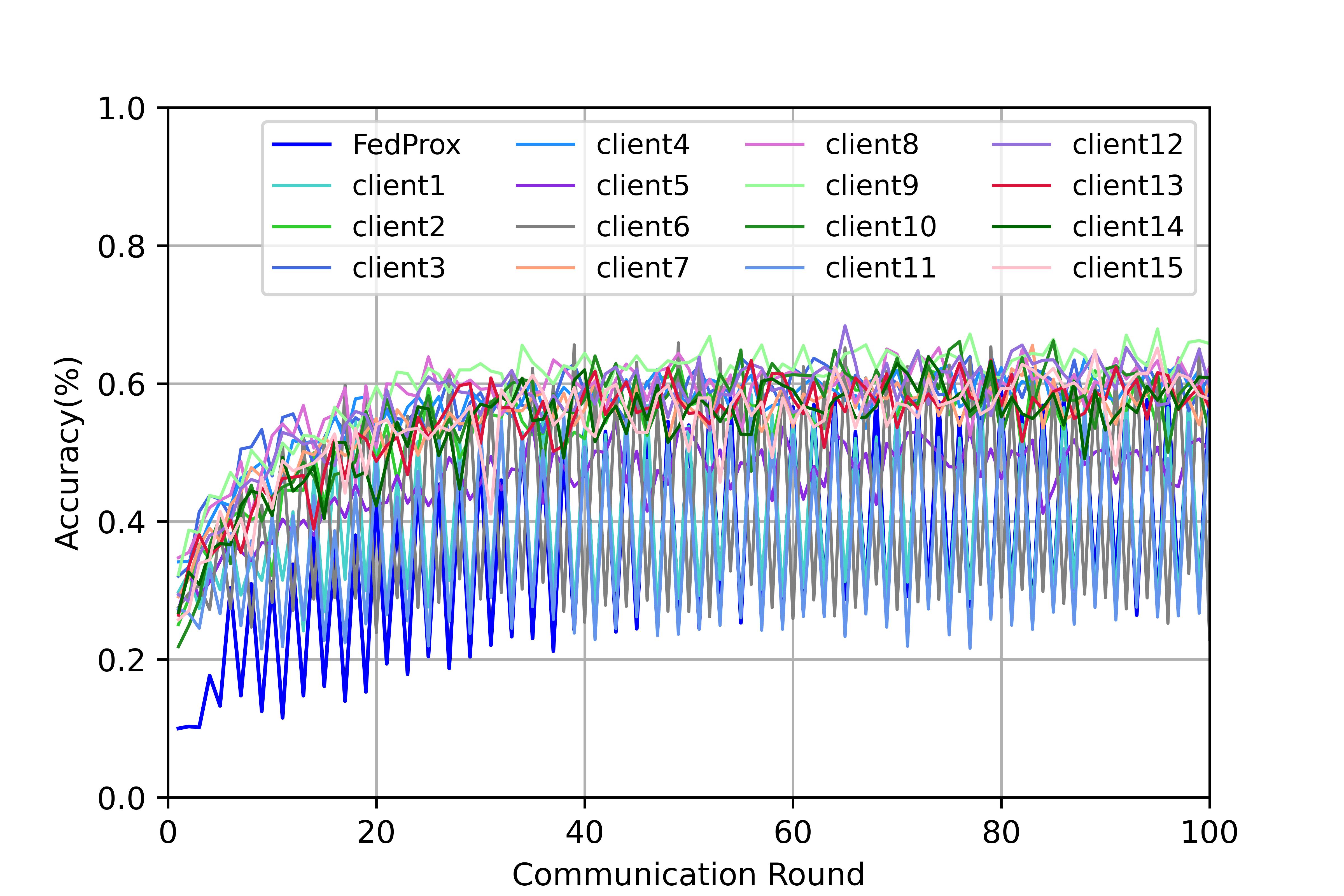}
        \centerline{(e) 15 clients FedProx }
    \end{minipage}
    \begin{minipage}[t]{0.33\linewidth}
        \centering
        \includegraphics[width=\textwidth]{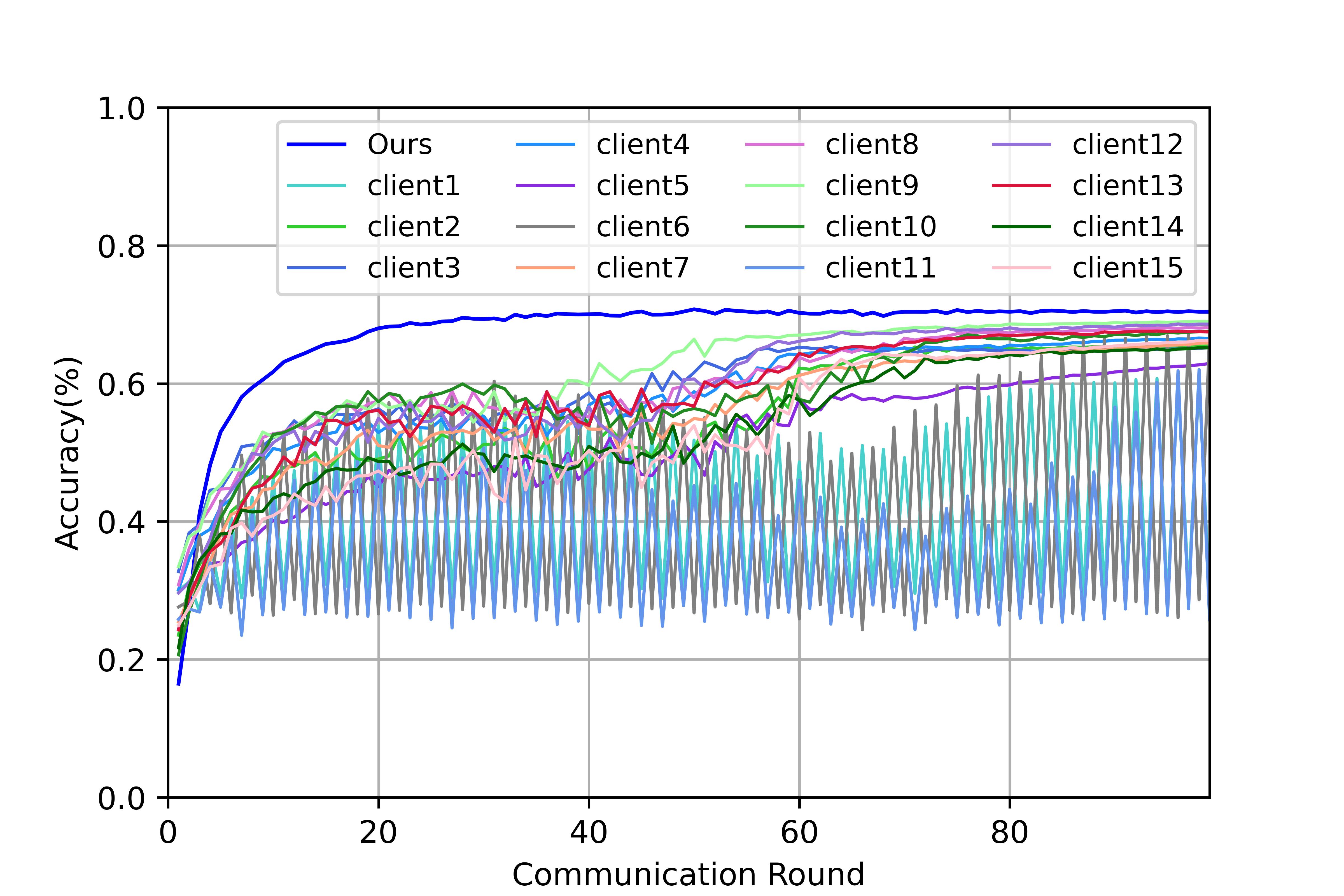}
        \centerline{(f) 15 clients FedDRL}
    \end{minipage}
\caption{The accuracy of the global model for different number of client in attack type 3}
\label{fig-attack3-acc}
\end{figure}

In attack type 3 scenarios, the FedAvg and FedProx algorithms experience significant fluctuations in global model accuracy due to alternating attack behaviors by malicious clients. Conversely, the agent within the FedDRL framework adaptively selects trusted clients, effectively excluding malicious entities from participating in model fusion, thereby enabling the FedDRL algorithm to operate with stability.
The experimental results are shown in Figure \ref{fig-attack3-acc}.

\subsubsection{Low-quality Model Fusion Experiments}

In evaluating our FedDRL framework, we undertook validation using the Fashion-MNIST, CIFAR-10, and CIFAR-100 datasets. Given their open-source nature, these datasets are of high quality, leading to minimal variance in model accuracy among clients utilizing them directly. Thus, to simulate real-world conditions, we incorporated low-quality models into the global fusion process. We established a model accuracy threshold, ensuring that models uploaded by low-quality clients did not exceed this threshold in any communication round.

Experiments were carried out on the three datasets, with client groups of varying sizes—5, 10, and 15—participating in the global model fusion. We applied a Dirichlet distribution with parameter alpha=1 to achieve dataset segmentation among clients.
We set some clients to upload low-quality models; after several communication rounds, we controlled these client models' accuracy in global fusion, ensuring it remained within the 40\% to 55\% range.

Details of these low-quality model experiment configurations are specified in Table \ref{table-LowModel-Setting}. The FedDRL algorithm was compared against the FedAvg and FedProx methods across 100 communication rounds, with each client executing one epoch of local model training. Results are summarized in Table \ref{table-LowModel-Accuracy}.

\begin{table}
\centering
\caption{Experimental settings for low-quality model experiments}
\resizebox{\textwidth}{!}{
\begin{tabular}{ccccccc} %
\toprule %
\multirow{1}{*}{Number} & 
\multicolumn{1}{c}{Dataset} &
\multicolumn{1}{c}{Low-quality Model ID} &
\multicolumn{1}{c}{Number of samples} &
\multicolumn{1}{c}{Accuracy of models ($\leq$)} \\
\midrule %

\multirow{3}{*}{Clients=5} 
&Fashion-MINST & Client1 & 9061 & 53\%\\
&CIFAR-10 & Client1 & 7750 & 52\%\\
&CIFAR-100 & Client1 & 9278  & 22\% \\
\hline

\multirow{3}{*}{Clients=10} 
&Fashion-MINST & Client1,Client5 & 5071, 7245 & 51\%, 52\%  \\
&CIFAR-10 & Client1,Client5 & 4222, 6039 & 50\%, 54\%  \\
&CIFAR-100 & Client1,Client5 &4191, 5491 & 22\% \\
\hline

\multirow{3}{*}{Clients=15} 
&Fashion-MINST & Client1,Client5,Client10 & 4405, 3752, 1809 & 52\%, 51\%, 53\%  \\
&CIFAR-10 & Client1,Client5,Client10 & 3670, 3128, 1509 & 49\%, 52\%, 55\% \\
&CIFAR-100 & Client1,Client5,Client10 &3073, 3494, 2910 & 22\% 19\% 23\% \\
\bottomrule 
\end{tabular}}
\label{table-LowModel-Setting}
\end{table}

The FedDRL algorithm was compared against the FedAvg and FedProx methods across 100 communication rounds, with each client executing one epoch of local model training. Results are summarized in Table \ref{table-LowModel-Accuracy}. Details of these low-quality model experiment configurations are specified in Table \ref{table-LowModel-Setting}. 

\begin{table} 

\centering
\caption{Accuracy of each algorithm for low-quality modeling experiments}
\resizebox{\textwidth}{!}{
\begin{tabular}{cccccccccccc} %
\toprule %
\multirow{2}{*}{Method} &
 & Fashion-MINST & & & CIFAR-10 & & & CIFAR-100  & 
\\
\cline{2-11}
 & C=5 & C=10 & C=15  & C=5 & C=10 & C=15 & C=5 & C=10 & C=15&\\ 
\midrule %

FedAvg & 0.857 & 0.858 & 0.841 & 0.705 & 0.664 & 0.602 & 0.386 & 0.373 & 0.365 \\
\hline

FedProx & 0.865 & 0.861 & 0.829 & 0.714 & 0.652 & 0.607 & 0.402 & 0.391 & 0.386\\
\hline
Ours & \pmb{0.885} & \pmb{0.887} & \pmb{0.884} & \pmb{0.725} & \pmb{0.706} & \pmb{0.698} & \pmb{0.422} & \pmb{0.418} & \pmb{0.407}\\
\bottomrule %
\end{tabular}}
\label{table-LowModel-Accuracy}
\end{table}

\begin{figure}
    \begin{minipage}[t]{0.33\linewidth}
        \centering
        \includegraphics[width=\textwidth]{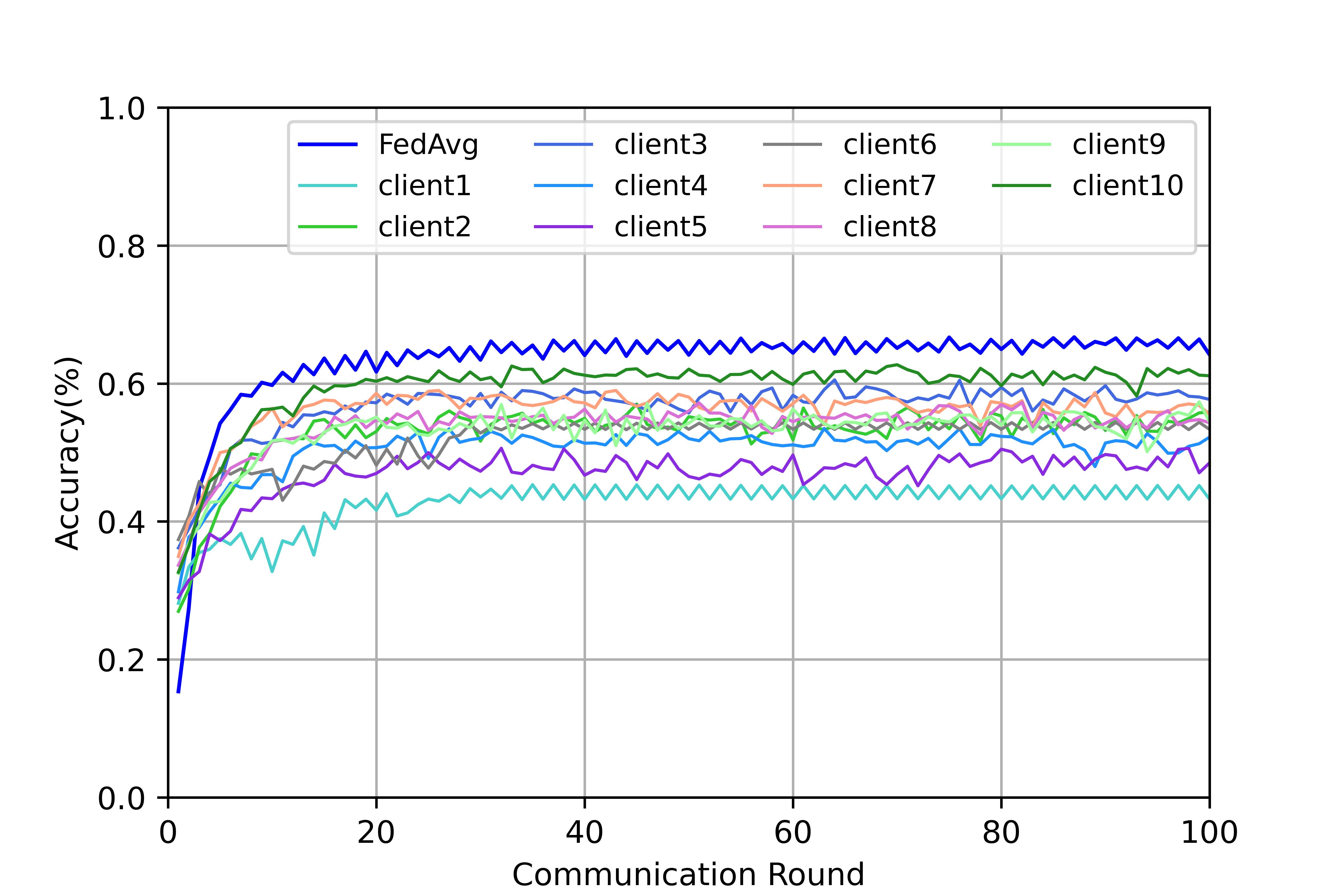}
        \centerline{(a) 10 clients FedAvg}
    \end{minipage}%
    \begin{minipage}[t]{0.33\linewidth}
        \centering
        \includegraphics[width=\textwidth]{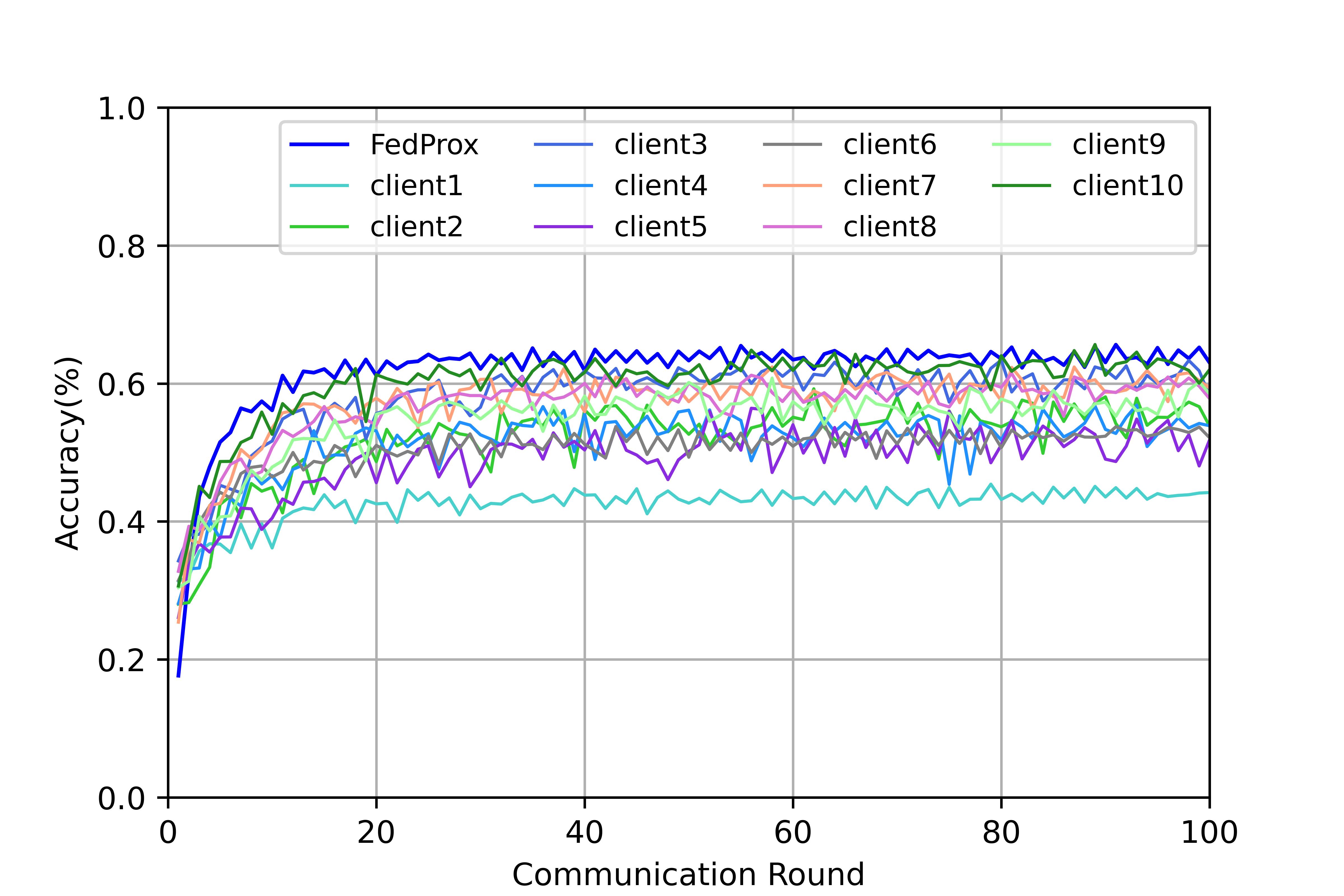}
        \centerline{(b) 10 clients FedProx}
    \end{minipage}
    \begin{minipage}[t]{0.33\linewidth}
        \centering
        \includegraphics[width=\textwidth]{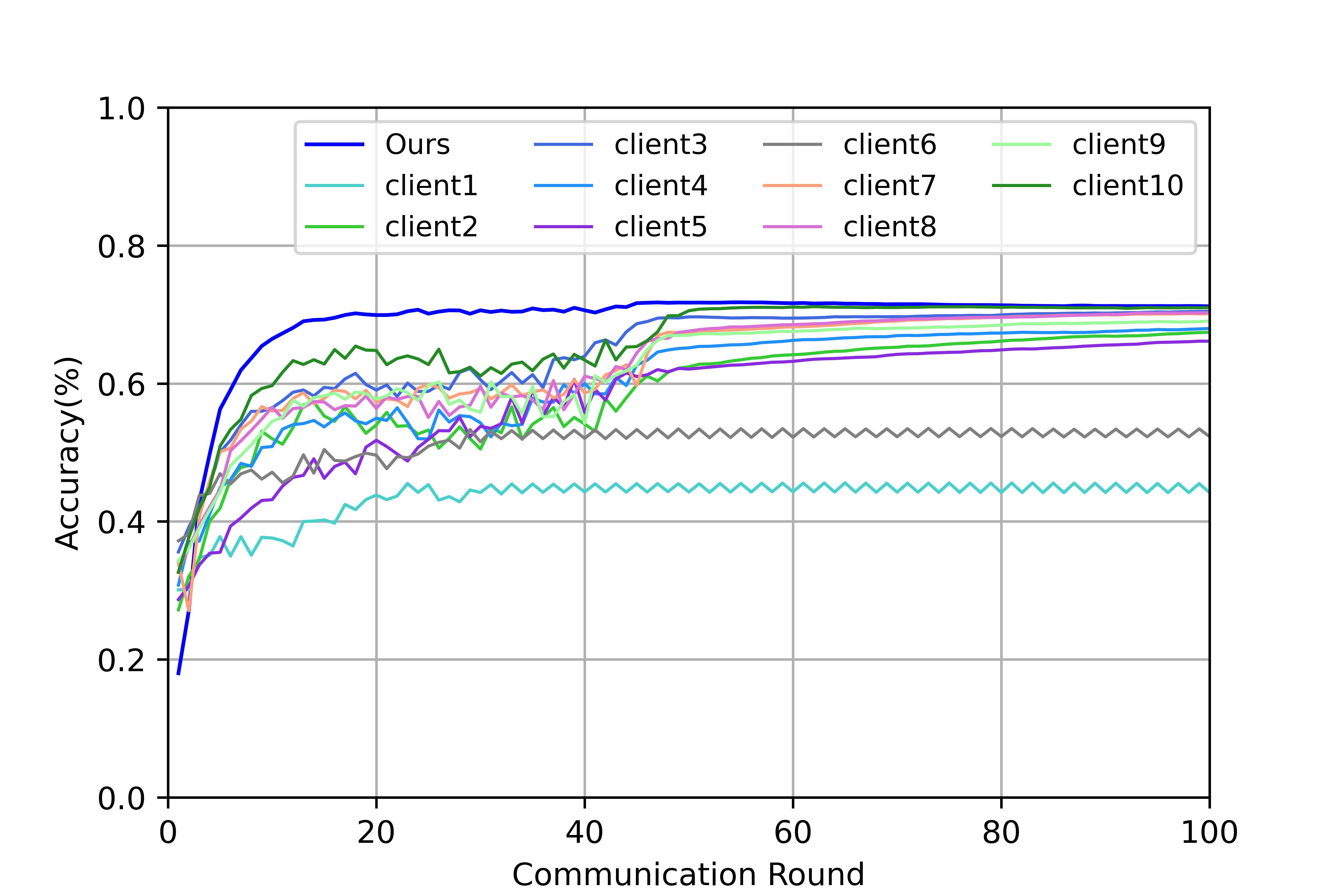}
        \centerline{(c) 10 clients  FedDRL}
    \end{minipage}
    \begin{minipage}[t]{0.33\linewidth}
        \centering
        \includegraphics[width=\textwidth]{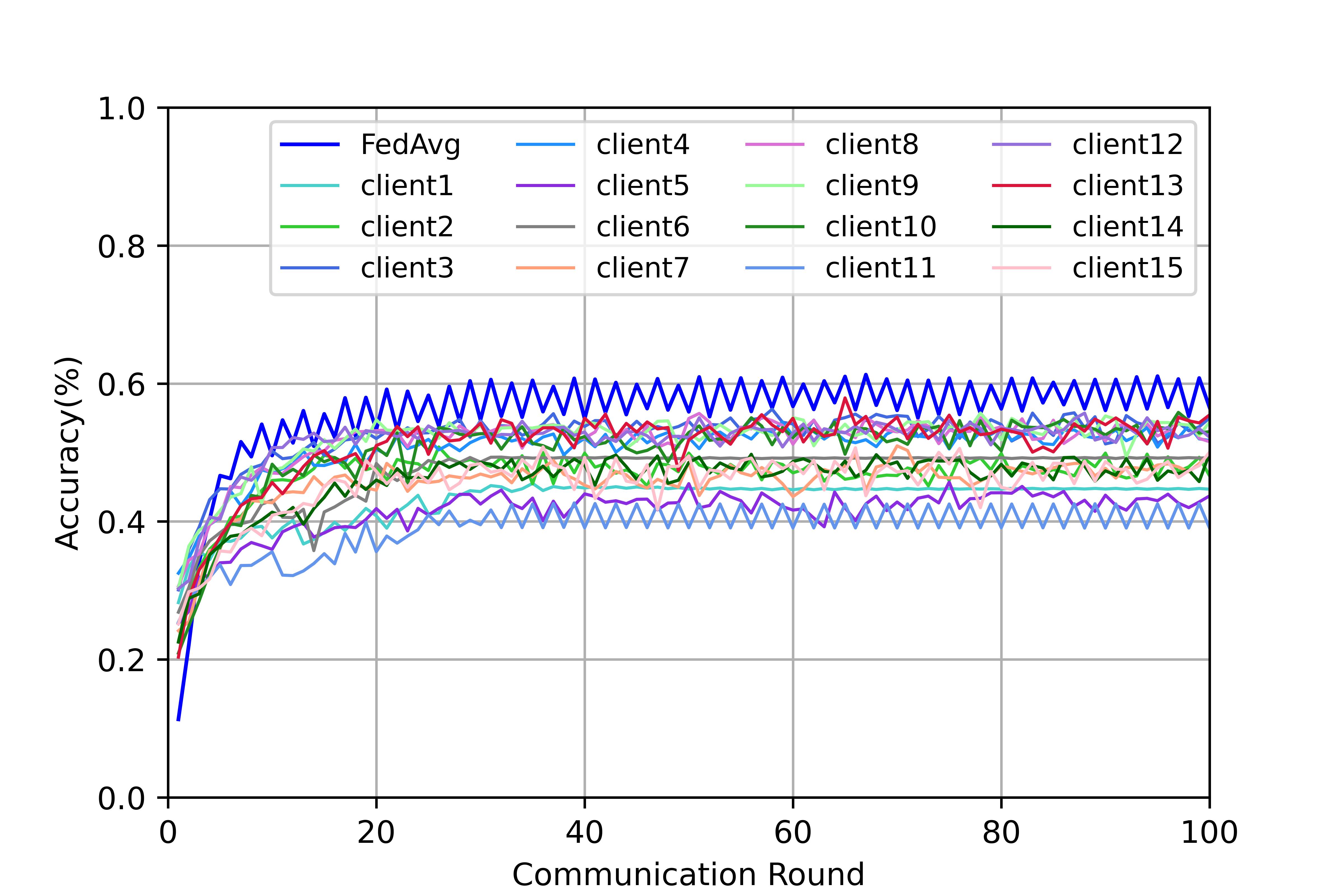}
        \centerline{(d) 15 clients FedAvg }
    \end{minipage}%
    \begin{minipage}[t]{0.33\linewidth}
        \centering
        \includegraphics[width=\textwidth]{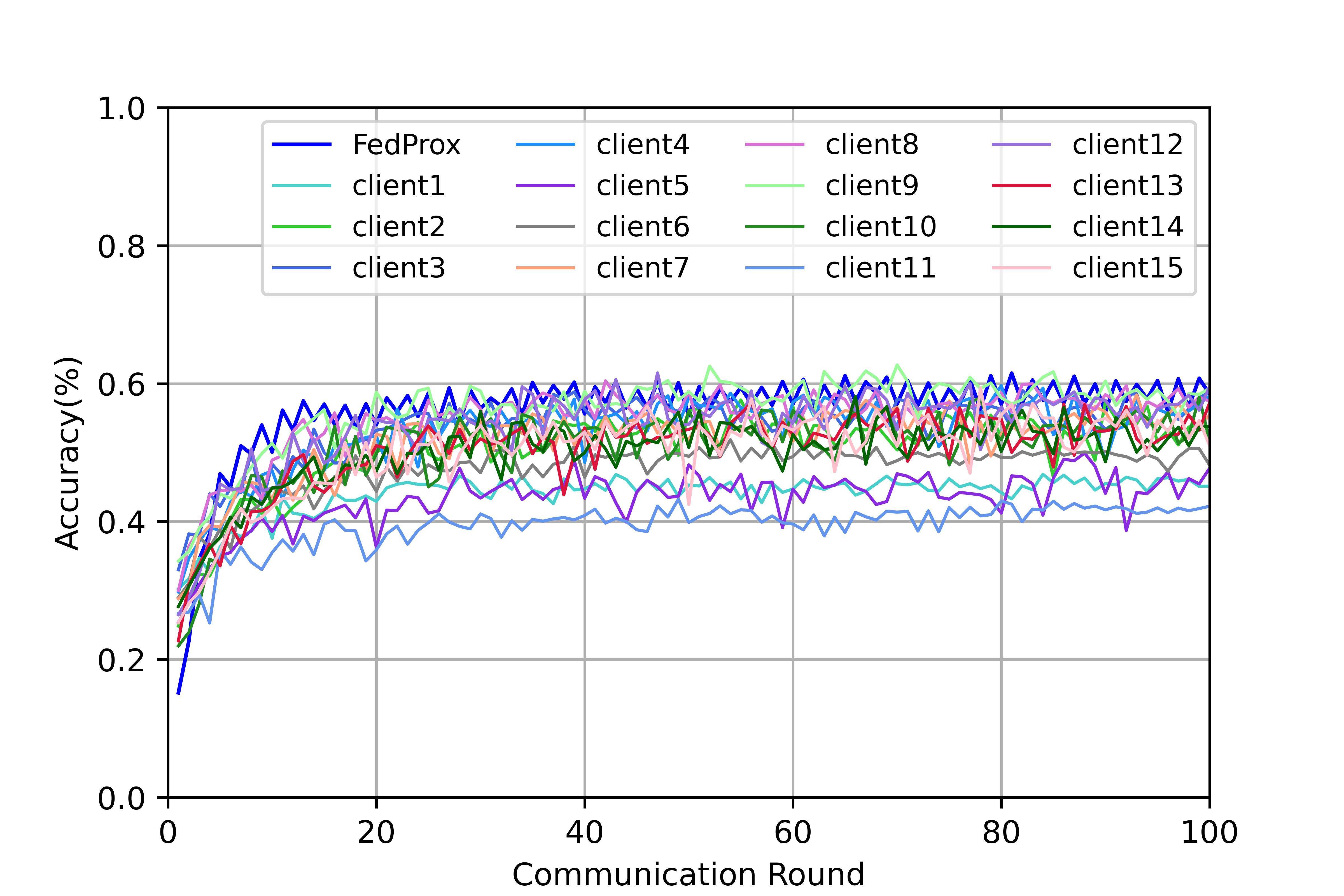}
        \centerline{(e) 15 clients FedProx }
    \end{minipage}
    \begin{minipage}[t]{0.33\linewidth}
        \centering
        \includegraphics[width=\textwidth]{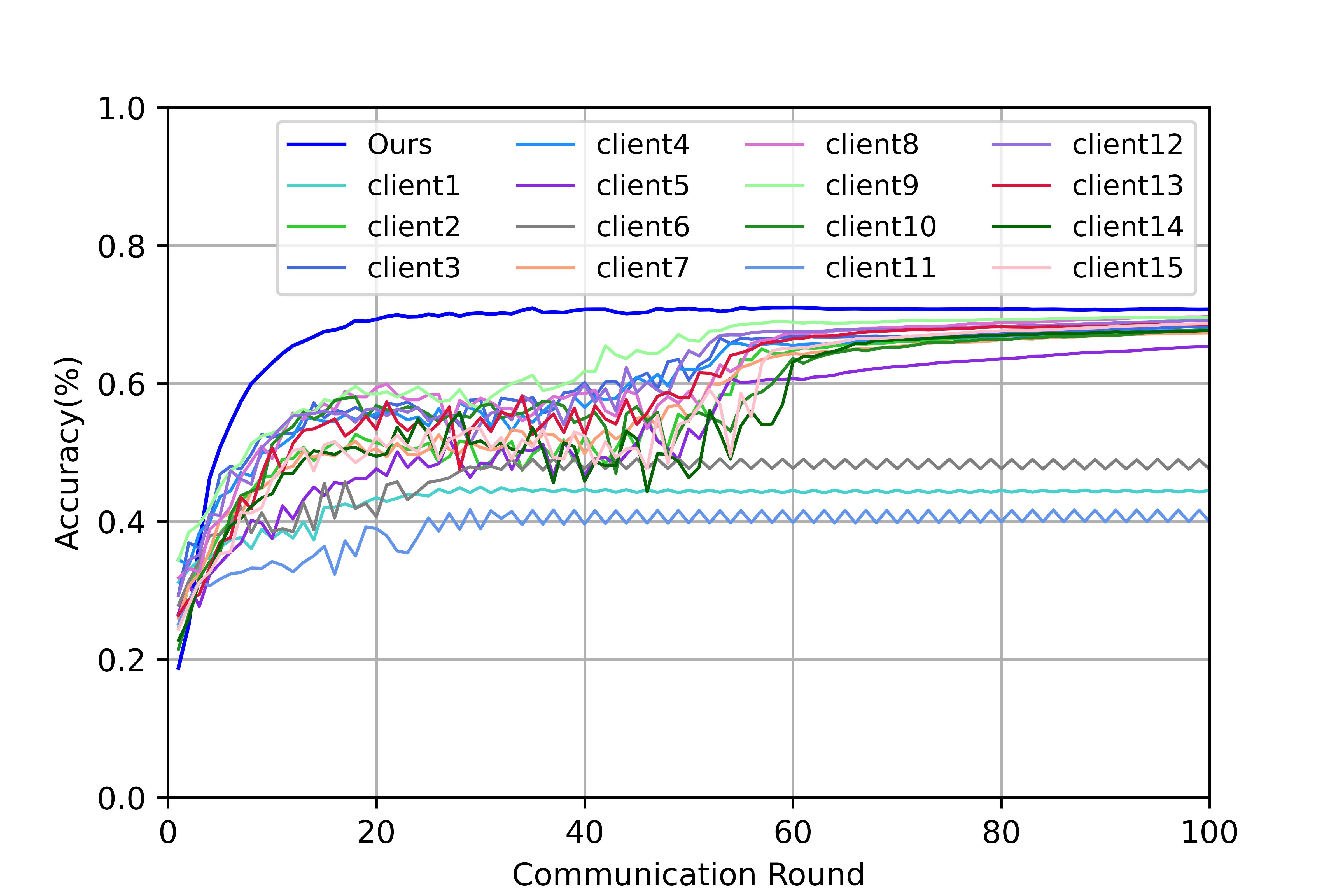}
        \centerline{(f) 15 clients FedDRL }
    \end{minipage}
    \caption{The accuracy of a global model for different numbers of client in Low-quality scenario.}
\label{fig-LowModel-Attack}
\end{figure}

Employing the CIFAR-10 dataset for illustrative purposes, we performed comparative analyses for setups with 10 and 15 clients, respectively; the findings are depicted in Figure \ref{fig-LowModel-Attack}. The experiments indicate that the accuracy of the FedAvg and FedProx methods deteriorates as the prevalence of low-quality models increases. This decline can be attributed to these algorithms' reliance on sample count for determining the fusion weight values of the models, where the inclusion of low-quality models adversely impacts the global model's accuracy. Conversely, FedDRL surpasses both methodologies in terms of global model convergence speed and accuracy. This is because FedDRL adaptively recalibrates the weights assigned to each client's model based on quality, thereby diminishing the adverse effects of low-quality models on the global model's accuracy and consequently hastening the global model's convergence rate.

\subsubsection{Hybrid experiment}

In this section, we establish a hybrid scenario incorporating two types of attacking clients (type 1 and type 3) alongside clients submitting low-quality models. We assess the effectiveness of the FedDRL algorithm within this mixed scenario and benchmark it against the FedAvg and FedProx approaches.

Employing the CIFAR-10 dataset, we set different numbers of clients (10,15) participating in global model fusion, respectively. Client 1 persistently uploads merely the initial model at each round. Client 6 emulates the submission of low-quality models for fusion, and Client 10 or 11 engages in attack behaviour during odd communication rounds but normally participates during even rounds. The remainder of the nodes contribute routinely to each cycle of the federated learning tasks. The experimental setup specifics are delineated in Table \ref{table-Hybrid-Config}.

\begin{table}
\centering
\caption{Experimental settings for hybrid scenarios}
\resizebox{\textwidth}{!}{
\begin{tabular}{ccccccc} %
\toprule %
\multirow{1}{*}{Number} & 
\multicolumn{1}{c}{Client ID} &
\multicolumn{1}{c}{Type} &
\multicolumn{1}{c}{Number of samples} &
\multicolumn{1}{c}{Model Accuracy} \\
\midrule %

\multirow{3}{*}{Clients=10} 
& Client1 &Attack Type 1   & 4222 &  $A \leq 10\%$  
\\
& Client6  &Low-quality Model  & 4938  &  $45\%\leq A \leq50\%$  \\
& Client10 &Attack Type 3  & 3560 & Attack round   $A \leq15\%$
\\
\multirow{3}{*}{Clients=15} 
& Client1 &Attack Type 1   & 3670 &  $A \leq 10\%$  
\\
& Client6  &Low-quality Model  & 3314  &  $45\%\leq A \leq50\%$  \\
& Client11 &Attack Type 3  & 4453 & Attack round   $A \leq15\%$
\\
\bottomrule %
\end{tabular}}
\label{table-Hybrid-Config}
\end{table}

\begin{figure}
      \begin{minipage}[t]{0.33\linewidth}
        \centering
        \includegraphics[width=\textwidth]{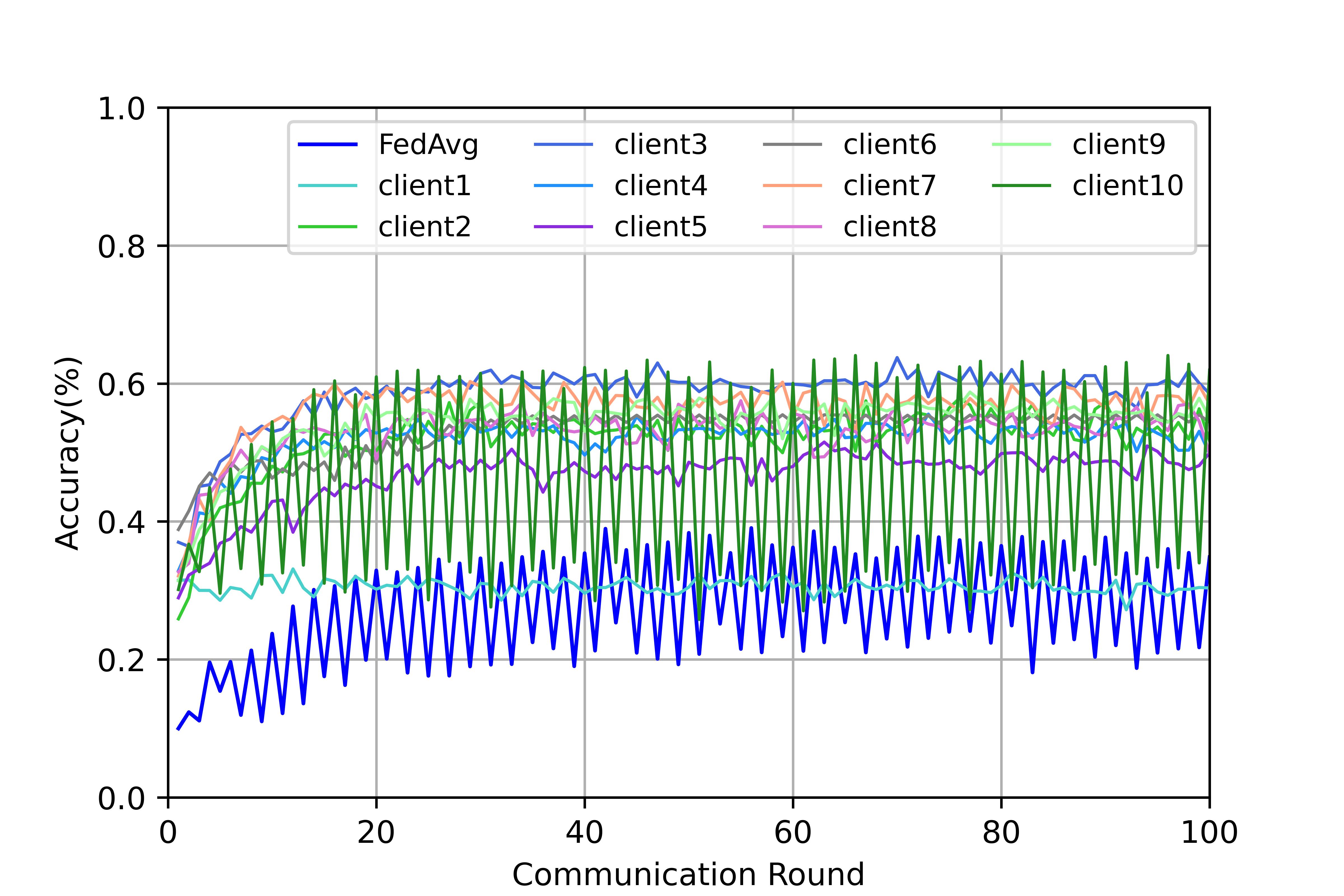}
        \centerline{(a) 10 clients FedAvg}
    \end{minipage}%
    \begin{minipage}[t]{0.33\linewidth}
        \centering
        \includegraphics[width=\textwidth]{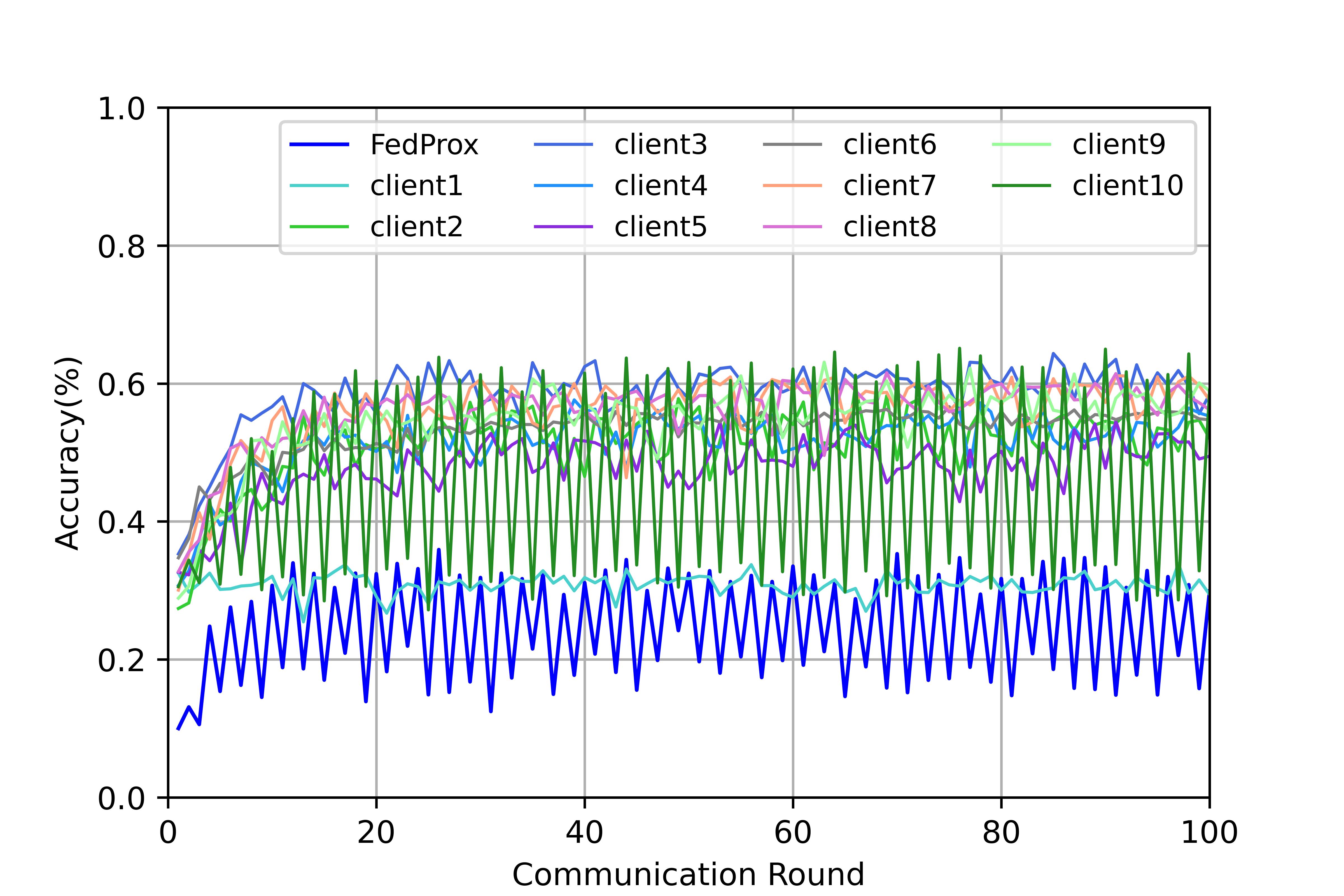}
        \centerline{(b) 10 clients FedProx}
    \end{minipage}
    \begin{minipage}[t]{0.33\linewidth}
        \centering
        \includegraphics[width=\textwidth]{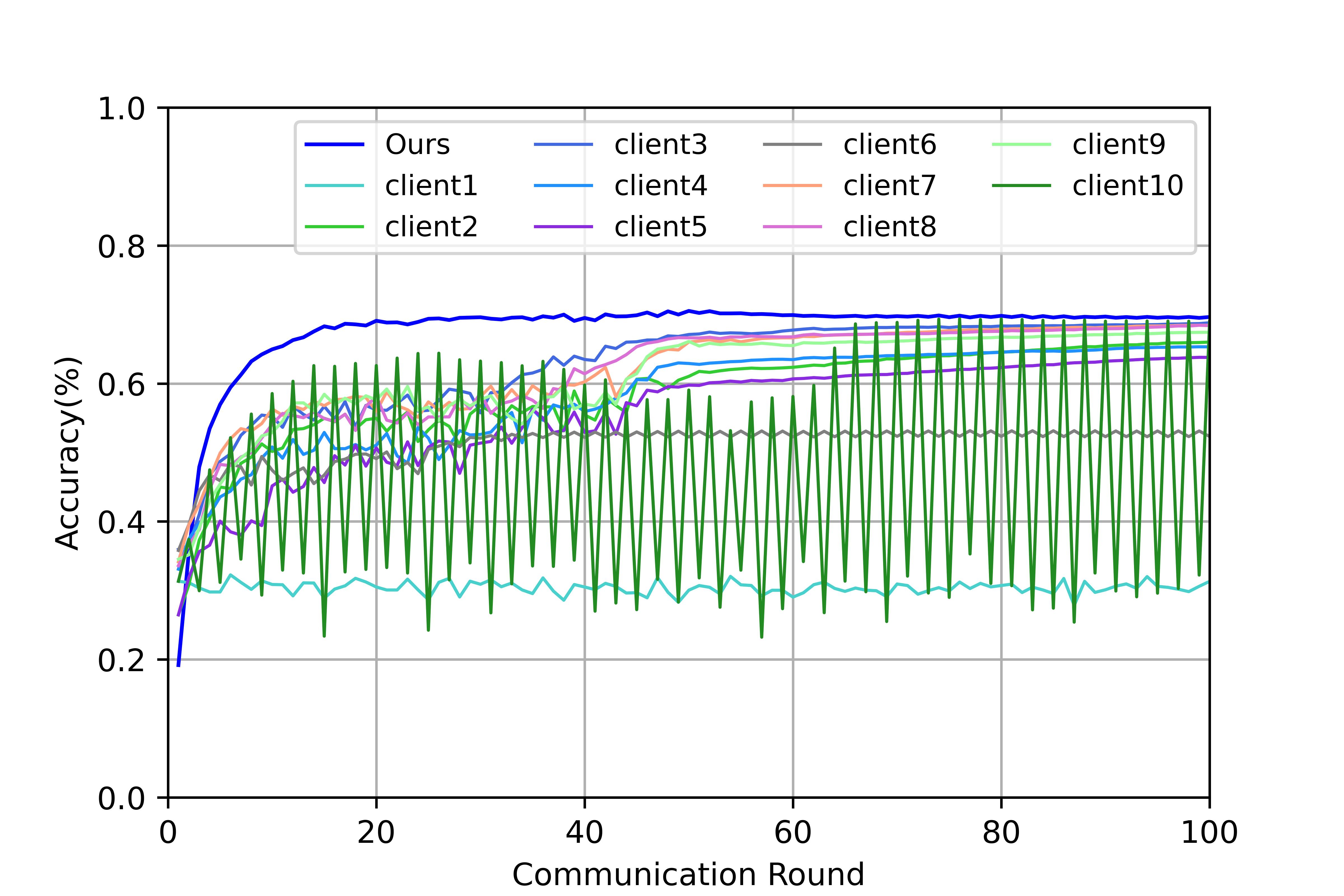}
        \centerline{(c) 10 clients FedDRL}
    \end{minipage}
      \begin{minipage}[t]{0.33\linewidth}
        \centering
        \includegraphics[width=\textwidth]{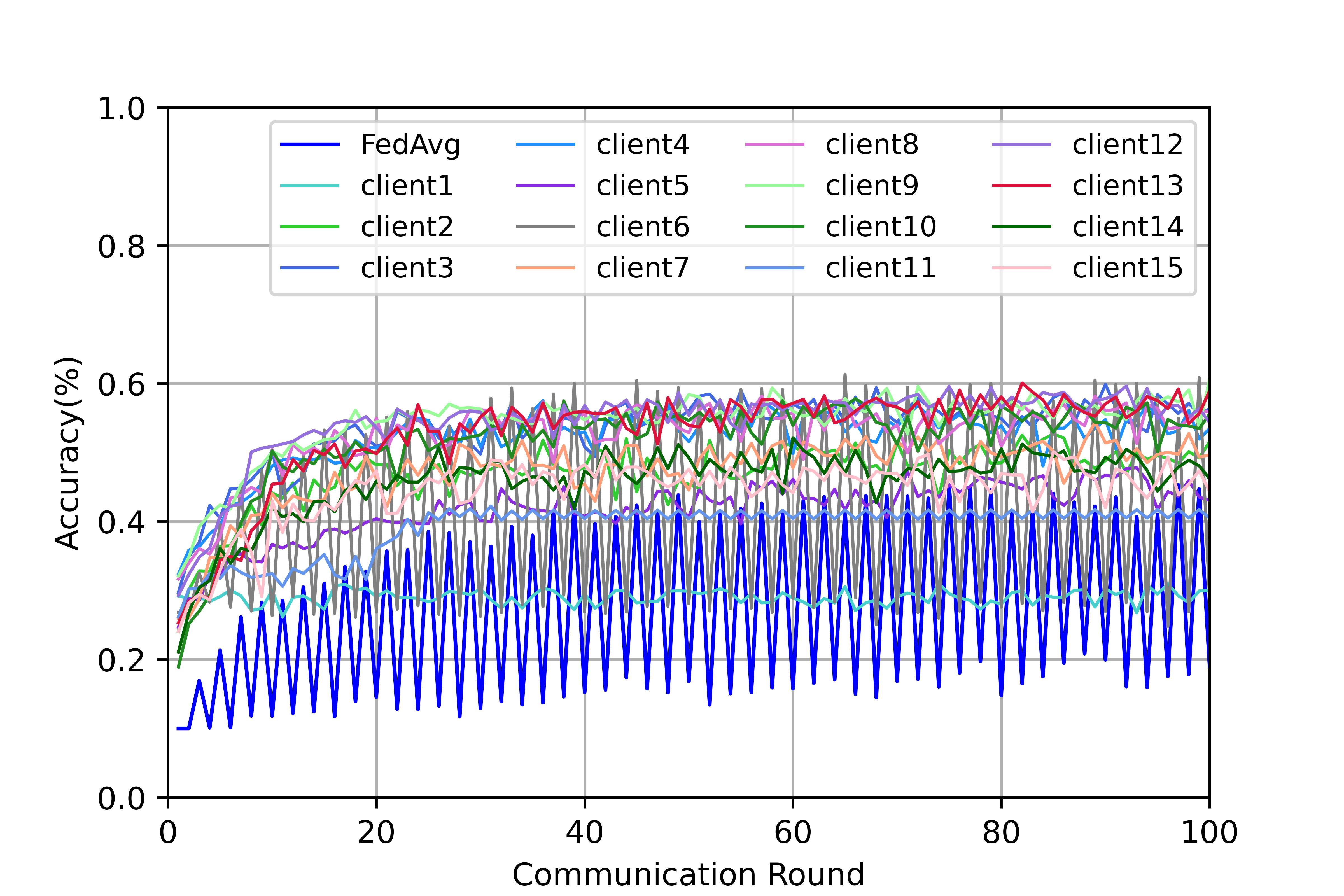}
        \centerline{(d) 15 clients FedAvg}
    \end{minipage}%
    \begin{minipage}[t]{0.33\linewidth}
        \centering
        \includegraphics[width=\textwidth]{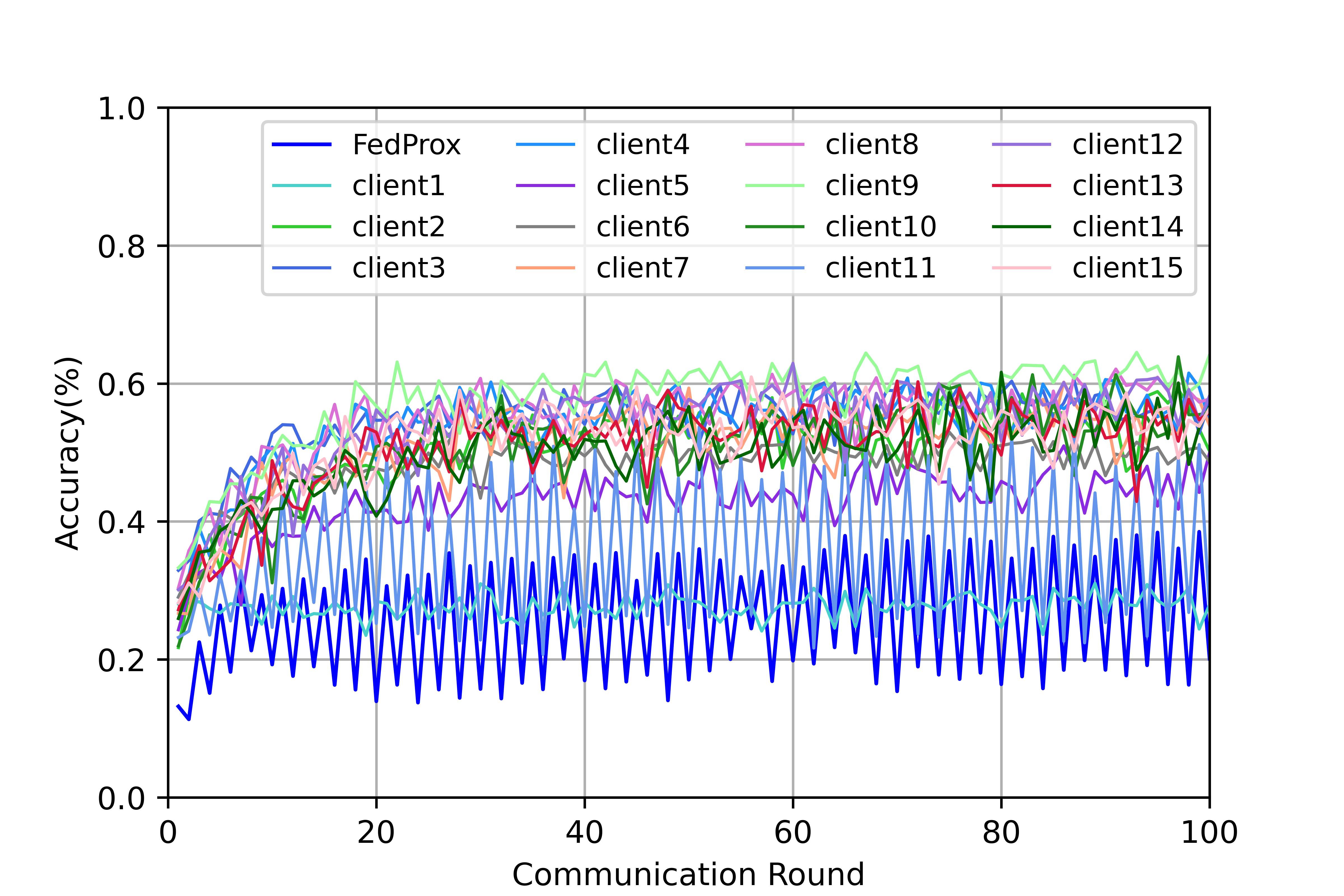}
        \centerline{(e) 15 clients FedProx}
    \end{minipage}
    \begin{minipage}[t]{0.33\linewidth}
        \centering
        \includegraphics[width=\textwidth]{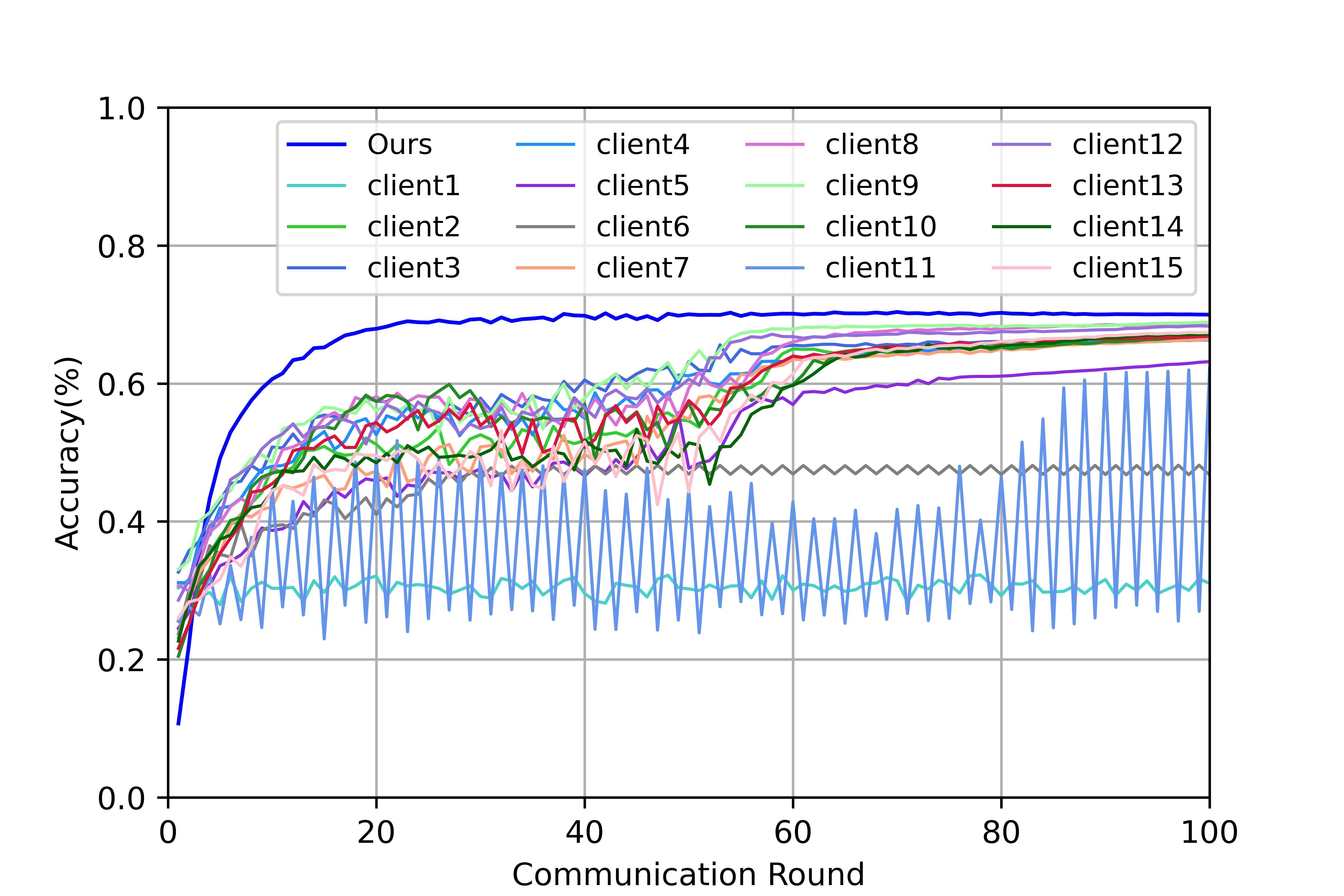}
        \centerline{(f) 15 clients FedDRL}
    \end{minipage}
\caption{Comparison of global model accuracy between different algorithms.}
\label{fig-Hybrid-acc}
\end{figure}

After completing 100 communication rounds, we present the global model accuracy for each algorithm in Table \ref{table-Hybrid-Accuracy}. The comparative global model accuracies and individual client model accuracies per communication round, as determined by these three algorithms, are depicted in Figure \ref{fig-Hybrid-acc}.
The experimental outcomes from the hybrid scenario reveal that the FedAvg and FedProx algorithms falter in properly conducting global model fusion due to the adversarial behavior of certain clients. Incorporating malicious models under traditional algorithmic frameworks significantly degrades the global model's accuracy.

\begin{table}
\centering
\caption{Accuracy of each algorithm for hybrid scenarios experiments}
\resizebox{\textwidth}{!}{
\begin{tabular}{ccccccc} %
\toprule %
\multirow{2}{*}{Method}
&\multicolumn{2}{c}{Fashion-MINST}&\multicolumn{2}{c}{CIFAR-10}&\multicolumn{2}{c}{CIFAR-100}\\
\cline{2-7}
& Clients=10 & Clients=15 & Clients=10 & Clients=15  & Clients=10 & Clients=15\\ 
\midrule %
FedAvg &  0.835 & 0.823 & 0.368 & 0.348 & 0.223 & 0.238 \\
\hline
FedProx  &  0.821 & 0.846 & 0.308 & 0.341 & 0.241 & 0.266 \\
\hline
Ours & \pmb{0.876} & \pmb{0.883}  & \pmb{0.701} & \pmb{0.698} & \pmb{0.426} & \pmb{0.418}\\
\bottomrule %
\end{tabular}}
\label{table-Hybrid-Accuracy}
\end{table}

The experimental outcomes show that FedAvg and FedProx's global model accuracies suffer from malicious attacks due to their weighted average-based fusion, which doesn't block harmful participants. Conversely, the FedDRL algorithm, through its two-stage approach, initially filters out malicious models from fusion and subsequently applies an adaptive weight strategy to diminish the impact of substandard models. Consequently, our algorithm maintains operational integrity even within this complex scenario.

\subsubsection{Agent Training Efficiency in the FedDRL Framework}

In this segment, our primary objective is to assess the training efficiency of agents within the FedDRL framework. To expedite the training process, we have implemented optimizations in two key areas. Initially, we adopted a distributed reinforcement learning methodology, enabling multi-agents to interact concurrently with the external environment. Concurrently, we introduced a memory cache module designed to prevent redundant sampling by multiple agents.

\textbf{Experimental Scenarios}: Our investigation encompasses varied attack scenarios across two distinct datasets: Fashion-MNIST and Cifar-10. In each scenario, we involve a total of 10 and 15 clients in the federated task, including 2 and 3 malicious clients accordingly.

\textbf{Comparison Experiments}: To ascertain the efficacy of the FedDRL framework, we initiated experiments featuring 1, 5, 10, and 20 agents. To guarantee the stability of the reward values acquired by the final agents, we designated the number of iterations for each experimental group to be 10,000, 15,000, 20,000, and 25,000, correspondingly.

\textbf{Experimental metrics}: Our evaluation involves counting the iterations necessary for reinforcement learning to reach stable rewards across different agent counts. We employ a sliding window approach to compute the average reward, depicting the progression of rewards attained by the agents. We define $r_{t}$ the agent's reward obtains in the $t$-th interaction and the sliding window as $W$. The formula for calculating the average reward is represented as equation \ref{eq-avg-reward}:
\begin{equation}\label{eq-avg-reward} 
\bar{R} =\frac{1}{W}  {\textstyle \sum_{t=1}^{W}}  r_{t}
\end{equation}
  
\textbf{Reward Parameter Setting}: Our reward function comprises two components: the global model accuracy reward and the reward for the number of credible nodes. For this experiment, these parameters are set to 100 and 10, respectively.

\begin{table}
\centering
\caption{The iterations of obtaining stable rewards for different numbers of agents}
{
\begin{tabular}{cccccc} %
\toprule %
\multirow{2}{*}{Dataset}
& \multirow{2}{*}{Attack Type}
&\multicolumn{4}{c}{The number of agents}
\\
\cline{3-6}
& & N=1 & N=5 & N=10 & N=20
\\ 
\midrule %

\multirow{2}{*}{Fashion-MNIST} 
&\multicolumn{1}{c}{Type-1}
&\multicolumn{1}{c}{25000}
&\multicolumn{1}{c}{20000}
&\multicolumn{1}{c}{16000}
&\multicolumn{1}{c}{\textbf{8000}}
\\
\cline{2-6}

\multirow{2}{*}{} 
&\multicolumn{1}{c}{Type-2}
&\multicolumn{1}{c}{25000}
&\multicolumn{1}{c}{21000}
&\multicolumn{1}{c}{15000}
&\multicolumn{1}{c}{\textbf{9000}}
\\
\cline{1-6}

\multirow{2}{*}{CIFAR-10} 
&\multicolumn{1}{c}{Type-1}
&\multicolumn{1}{c}{25000}
&\multicolumn{1}{c}{20000}
&\multicolumn{1}{c}{12000}
&\multicolumn{1}{c}{\textbf{10000}}
\\
\cline{2-6}
\multirow{2}{*}{} 
&\multicolumn{1}{c}{Type-2}
&\multicolumn{1}{c}{25000}
&\multicolumn{1}{c}{20000}
&\multicolumn{1}{c}{14000}
&\multicolumn{1}{c}{\textbf{9000}}
\\

\bottomrule %
\end{tabular}}
\label{table-Agent-Experiment}
\end{table}

\begin{figure}
    \begin{minipage}[t]{0.498\linewidth}
        \centering
        \includegraphics[width=\textwidth]{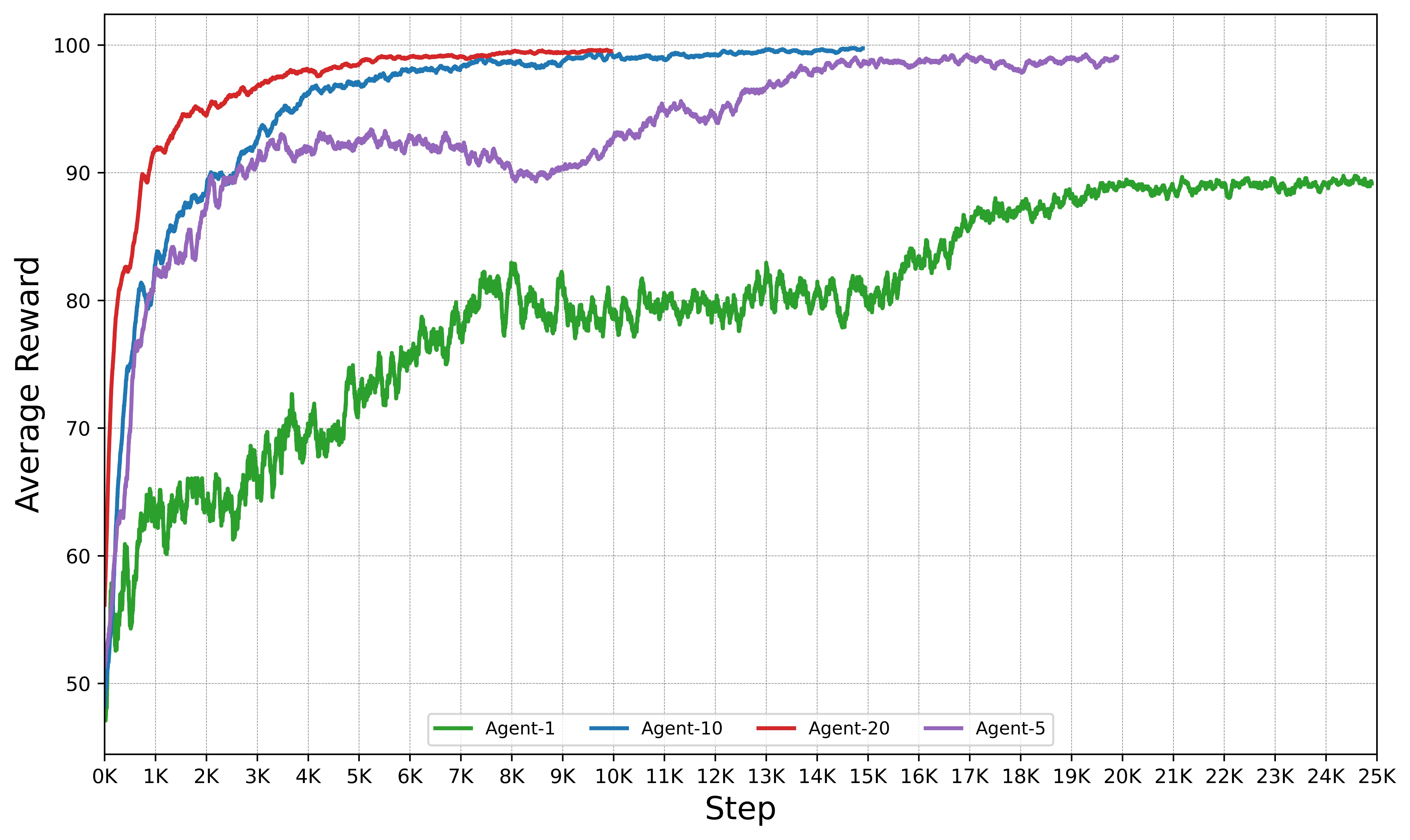}
        \centerline{(a) Fashion-MNIST (Attack Type 1)}
    \end{minipage}%
    \begin{minipage}[t]{0.498\linewidth}
        \centering
        \includegraphics[width=\textwidth]{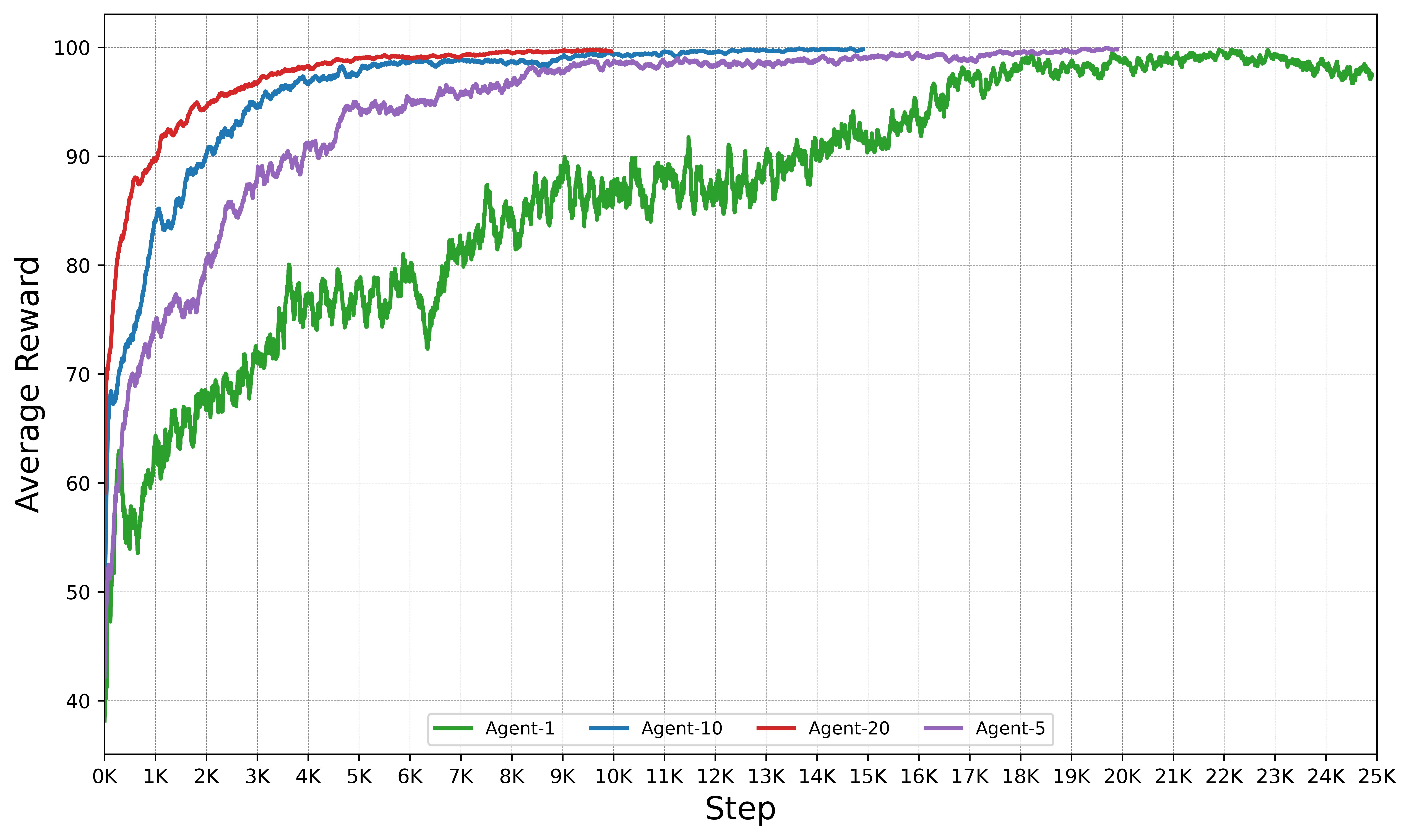}
        \centerline{(b) Fashion-MNIST (Attack Type 2)}
    \end{minipage}
    
    \begin{minipage}[t]{0.498\linewidth}
        \centering
        \includegraphics[width=\textwidth]{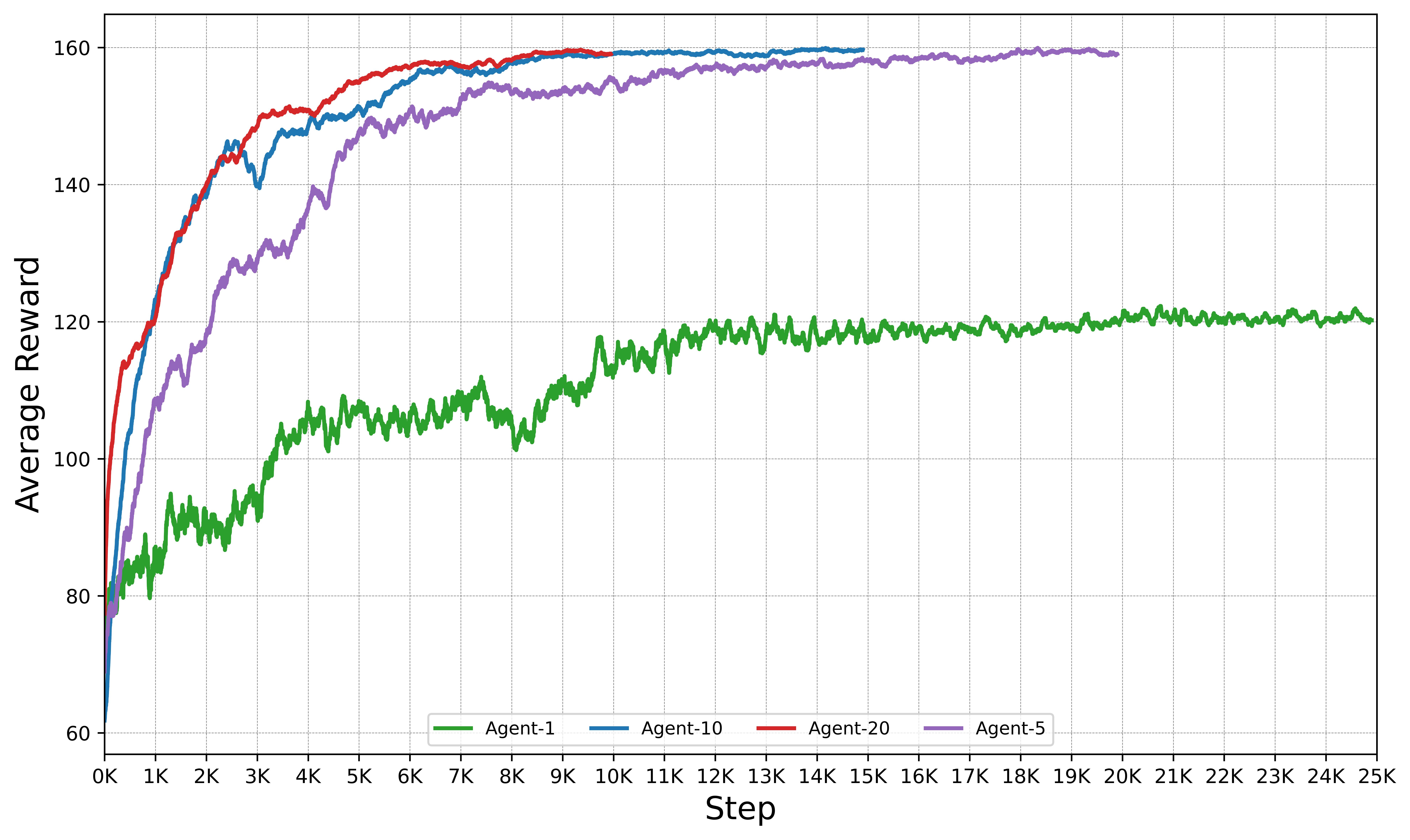}
        \centerline{(c) Cifar-10 (Attack Type 1)}
    \end{minipage}%
    \begin{minipage}[t]{0.498\linewidth}
        \centering
        \includegraphics[width=\textwidth]{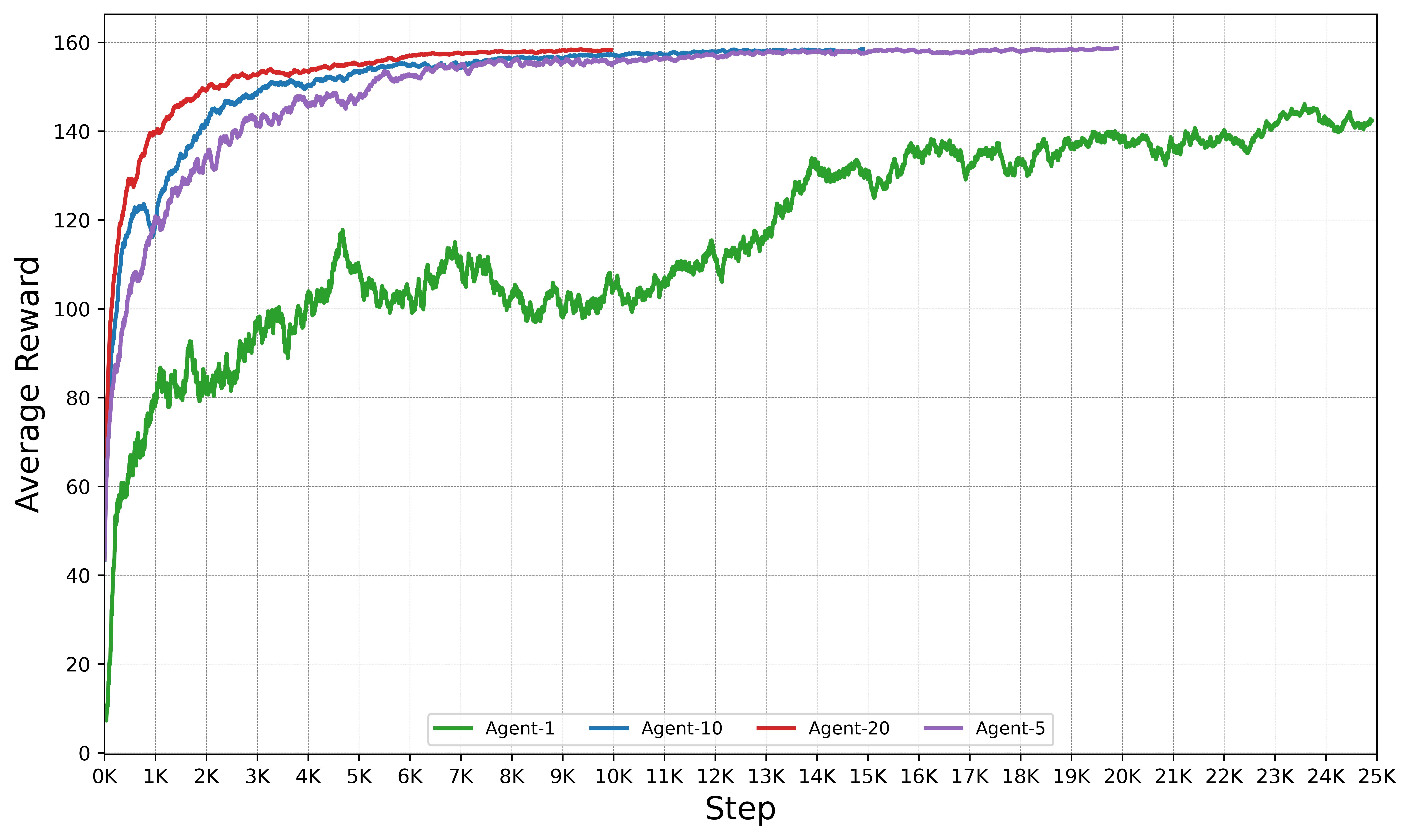}
        \centerline{(d) Cifar-10 (Attack Type 2)}
    \end{minipage}
\caption{Global model accuracy in 15 clients attack type 1}
\label{fig-train-agent}
\end{figure}

\textbf{Experimental Results}: In accordance with our experimental setup, we recorded the reward values for each iteration of the agents, as detailed in Figure \ref{fig-train-agent}. We have systematically arranged this information into Table \ref{table-Agent-Experiment} for enhanced clarity regarding the actual iterations across different experiments.

The data reveals a notable trend: The single agent does not get the optimal reward in some attack scenarios, because the single agent is easy to fall into the local optimal solution. Meanwile, an increase in agents correlates with reducing the iterations required to achieve a stable reward. However, this relationship is not strictly proportional because the multi-agent independently train their respective Actor and Critic networks. Each agent necessitates a distinct number of iterations to ensure the stability of its individual networks. Nevertheless, the simultaneous interaction of multiple agents with the environment markedly decreases the sampling time, demonstrating a clear trade-off between computational resources and time. This strategy underlines the significant computational resources required, highlighting a deliberate exchange of increased computational demand for reduced computational time.

\section{CONCLUSION}
To realize trustworthy federated learning, We propose a trusted reinforcement learning framework (FedDRL) based on staged reinforcement learning. The framework comprises two phases: selecting trusted clients and adaptive weight assignment. In the first phase, we design a reward strategy to train the agent, which allows the trained agent to exclude malicious client models from participating in model fusion based on the environment, and it also adaptively selects trustworthy clients for model fusion. In the second phase, we design a dynamic model weight calculation method, which can adaptively calculate the corresponding weights based on the model quality of each client. In addition, we propose a distributed reinforcement learning method to accelerate agent training. Finally, we design five model fusion scenarios to validate our approach, and the experiments show that our proposed algorithm can work reliably in various model fusion scenarios while maintaining global model accuracy.

Although a multi-agent distributed reinforcement learning approach can accelerate the agent training process, it sacrifices computational resources for the computational time. In our future work, we will continue to explore more lightweight and trustworthy federated learning methods. We will also investigate more efficient reinforcement learning methods for credible federated learning.


\end{document}